\documentclass[10pt,twocolumn,letterpaper]{article}

\usepackage[pagenumbers]{cvpr} 

\definecolor{cvprblue}{rgb}{0.21,0.49,0.74}
\usepackage[pagebackref,breaklinks,colorlinks,allcolors=cvprblue]{hyperref}
\usepackage{amssymb}
\usepackage{multirow}
\usepackage{pifont}
\usepackage[T1]{fontenc}
\def\paperID{4370} 
\def\confName{CVPR}
\def\confYear{2025}

\newcommand{\model}{SINGER}
\title{\model: Vivid Audio-driven Singing Video Generation with Multi-scale Spectral Diffusion Model}

\author{ Yan Li~ \quad
Ziya Zhou~\quad
Zhiqiang Wang~\quad
Wei Xue~\quad
Wenhan Luo~\quad
Yike Guo \\
{\tt \small \{ylitz, zzhoucp, zwangmk\}@connect.ust.hk \{weixue, whluo, yikeguo\}@ust.hk} \\
\url{https://yl4467.github.io/}
}

\begin{document}
\maketitle

\begin{abstract}
Recent advancements in generative models have significantly enhanced talking face video generation, yet singing video generation remains underexplored. 
The differences between human talking and singing limit the performance of existing talking face video generation models when applied to singing. 
The fundamental differences between talking and singing—specifically in audio characteristics and behavioral expressions—limit the effectiveness of existing models.
We observe that the differences between singing and talking audios manifest in terms of frequency and amplitude. 
To address this, we have designed a multi-scale spectral module to help the model learn singing patterns in the spectral domain.
Additionally, we develop a spectral-filtering module that aids the model in learning the human behaviors associated with singing audio.
These two modules are integrated into the diffusion model to enhance singing video generation performance, resulting in our proposed model, \textbf{\model}.
Furthermore, the lack of high-quality real-world singing face videos has hindered the development of the singing video generation community. To address this gap, we have collected an in-the-wild audio-visual singing dataset to facilitate research in this area. Our experiments demonstrate that \model~is capable of generating vivid singing videos and outperforms state-of-the-art methods in both objective and subjective evaluations.
\end{abstract}
    
\section{Introduction}
\label{sec:intro}
Audio-driven talking face generation has gained significant attention in recent years~\cite{prajwal2020lip,shen2023difftalk,chen2024echomimic,zhang2023sadtalker,xu2024hallo,cui2024hallo2,wang2021audio2head}, due to its diverse applications in areas such as digital human animation and movie production~\cite{wang2022one,ji2021audio,prajwal2020lip,yang2020one}. With the development of the diffusion-based generative models~\cite{dhariwal2021diffusion}, the diffusion-based talking face generation methods have gradually become mainstream, delivering impressive results in creating realistic talking head videos~\cite{wei2024aniportrait,zhang2024musetalk,xu2024hallo}. 
The primary objective of talking head generation is to create realistic videos that exhibit natural lip movements and head poses~\cite{zhang2023sadtalker}. However, the diffusion-based methods typically focus on fitting a common style across different identities with various speech styles to produce high-quality talking head videos~\cite{li2023generalization}, leading to synthesized videos where the talking lacks vividness. This limitation becomes particularly evident in the context of singing video generation, which demands richer expressions and more dynamic movements compared to speech. Thus the effectiveness of these diffusion-based methods is constrained when applied to singing tasks, necessitating the development of specialized models that can capture the complexity of singing.

Singing differs significantly from talking in two main aspects: the complexity of the singing audio and the variability of singing behavior expressions. These characteristics pose challenges for generating vivid singing videos. While there are some prior works focused on singing animation methods~\cite{wu2023singinghead,liu2024musicface,iwase2020song2face}, their performance has been limited, largely due to their inability to effectively capture and emphasize these unique singing patterns. In contrast to these earlier efforts, our approach aims to enhance singing video animation by leveraging the distinct properties of singing audio and behavior, thereby addressing the shortcomings of existing methods and facilitating the creation of more expressive and dynamic singing videos.

To capture the complex patterns inherent in singing audio, we note that it exhibits more intricate frequency and amplitude variations compared to talking audio~\cite{kim2021fre}. To extract these spectral patterns, we have developed a Multi-scale Spectral Module~(MSM) that employs wavelet transform~\cite{adamowski2011wavelet} to decompose the singing audio into sub-bands, each representing different frequency levels~\cite{pouyani2022lung,jin2024novel,dutt2023wavelet}. Then by assigning tunable weights to these sub-bands, we can highlight key frequency patterns critical for generating realistic singing videos. Additionally, we integrate visual information into the audio embeddings to strengthen the correlation between audio and visual features.
To further convey the vivid behavioral expressions characteristic of singing, we introduce a Self-adaptive Filter Module~(SFM). This module selectively enhances the features extracted from the audio, ensuring that the generated videos are closely aligned with the input audio. Together, these components work synergistically to improve the realism and expressiveness of the generated singing videos.

Moreover, the scarcity of public singing video datasets significantly hinders the advancement of singing animation technologies. While some works have introduced their own datasets, these often suffer from limitations such as small sample sizes~\cite{iwase2020song2face,liu2024musicface} or a lack of diversity in scenes and languages~\cite{wu2023singinghead}. There are currently no in-the-wild datasets available for this purpose. To address this challenge, we have curated a high-quality dataset of singing videos sourced from online platforms named the Singing Head Videos (SHV) dataset. This dataset comprises over $200$ subjects and features a total duration of approximately $20$ hours, providing a more comprehensive resource for advancing singing video generation research.

Our contributions are summarized as follows:
    \begin{itemize}[leftmargin=2em]
    \item We introduce a novel Multi-scale Spectral Module that utilizes wavelet transform to decompose singing audio into sub-bands, allowing us to capture complex frequency and amplitude patterns.
    \item We develop a Self-adaptive Filter Module designed to emphasize the extracted features correlated with audio, ensuring that the generated videos align closely with the input singing audio.
    \item We collect and curate a high-quality in-the-wild singing video dataset, which addresses the current lack of in-the-wild singing video datasets, providing a valuable resource for further research and development in singing video generation.
    \end{itemize}


\section{Related Work}
\label{sec:relatedwork}
\subsection{Talking/Singing Head Generation}
\textbf{Talking Head Generation}.
Audio-driven talking head generation focuses on animating faces based on provided speech audio. Early approaches primarily rely on GAN-based methods mapping audio signals into latent feature spaces~\cite{xu2024facechain,liang2022expressive,wang2023seeing,zhou2021pose} and employ a conditioned image generation framework to synthesize facial movements. To bridge the audio-visual gap, many studies integrate explicit structural information, such as facial landmarks~\cite{shen2023difftalk,zhong2023identity} and 3D face models~\cite{song2022everybody,lahiri2021lipsync3d,ji2021audio,zhang2023sadtalker}, to accurately reflect audio features in facial animations. However, these GAN-based methods often exhibit limited generalization capabilities, making them ineffective for arbitrary images and real-world audio scenarios.
In recent years, the advancements in diffusion models for image synthesis have inspired new approaches in talking head generation that leverage these diffusion-based techniques~\cite{shen2023difftalk,stypulkowski2024diffused,chen2024echomimic,wei2024aniportrait,xu2024hallo,cui2024hallo2}. These methods rely on large-scale training datasets to achieve improved performance. However, trainable diffusion models often struggle to maintain several desirable properties, including diverse generation capabilities. As a result, they tend to produce less vivid talking heads, lacking the expressive qualities needed for more engaging animations.

\textbf{Singing Head Generation}.
Singing Head Generation remains a largely underexplored area in the field of audio-driven animation. Early efforts, such as Song2Face~\cite{iwase2020song2face}, focus on creating animation models specifically for singing scenarios, but they primarily function well only with plain human singing voices and struggle with background music interference. Similarly, MusicFace~\cite{liu2024musicface} aims to synthesize expressive singing faces from mixed music signals but falls short of emphasizing the unique characteristics of singing, resulting in animations characterized by simple movements. Furthermore, Wu et al.~\cite{wu2023singinghead} introduce a unified framework for singing head animation that includes both 3D and 2D synthesis, yet it still relies on additional 3D facial motion data for effective animation. Given this limited research landscape, our work advances beyond previous studies by leveraging diffusion models to generate vivid singing videos driven solely by audio, highlighting the intricate dynamics of singing performances.


\subsection{Audio Time-Frequency Analysis}
Time-frequency analysis, which combines spectral representation with temporal evolution and localization~\cite{crocco2016audio}, is essential in audio processing.
Commonly used methods like wavelet coefficients~\cite{tzanetakis2001audio}, short-time Fourier transform (STFT)~\cite{sturmel2011signal, benesty2011speech, liu2016rolling}, and Wigner distribution function (WDF)~\cite{baydar2001comparative, huang2015sound, djebbari2013detection} have shown effectiveness in audio-related tasks compared to using time or frequency features alone. 
Among these, wavelet transform has been employed to address music-related tasks for several decades~\cite{tzanetakis2001audio, lambrou1998classification, loni2014formant}, including beat detection~\cite{tzanetakis2001audio} and music classification~\cite{lambrou1998classification}, owing to its superior capability in time-frequency representation.
Recent works have increasingly focused on modeling wavelet features using a multi-scale approach, leveraging its variable time resolution capabilities~\cite{pouyani2022lung,jin2024novel,dutt2023wavelet}. 
However, challenges still remain in selecting the appropriate wavelet basis function for specific downstream tasks.

Some studies have explored spectral features to enhance cross-modal understanding and generation tasks~\cite{pan2022wnet,jin2020exploring,phung2023wavelet}, yet these methods often overlook the critical correlations between audio signals and video frames in the spectral domain. To address this limitation while capitalizing on the advantages of discrete wavelet transform (DWT), we propose a tunable weighting scheme for the decomposed sub-bands, enhancing DWT's adaptability to various audio signals. Furthermore, we incorporate visual information into the audio sub-bands in the spectral domain, strengthening the correlation between audio and visual features.

\section{Method}
\label{sec:method}
\begin{figure*}[t]
    \centering
    \includegraphics[width=0.98\linewidth]{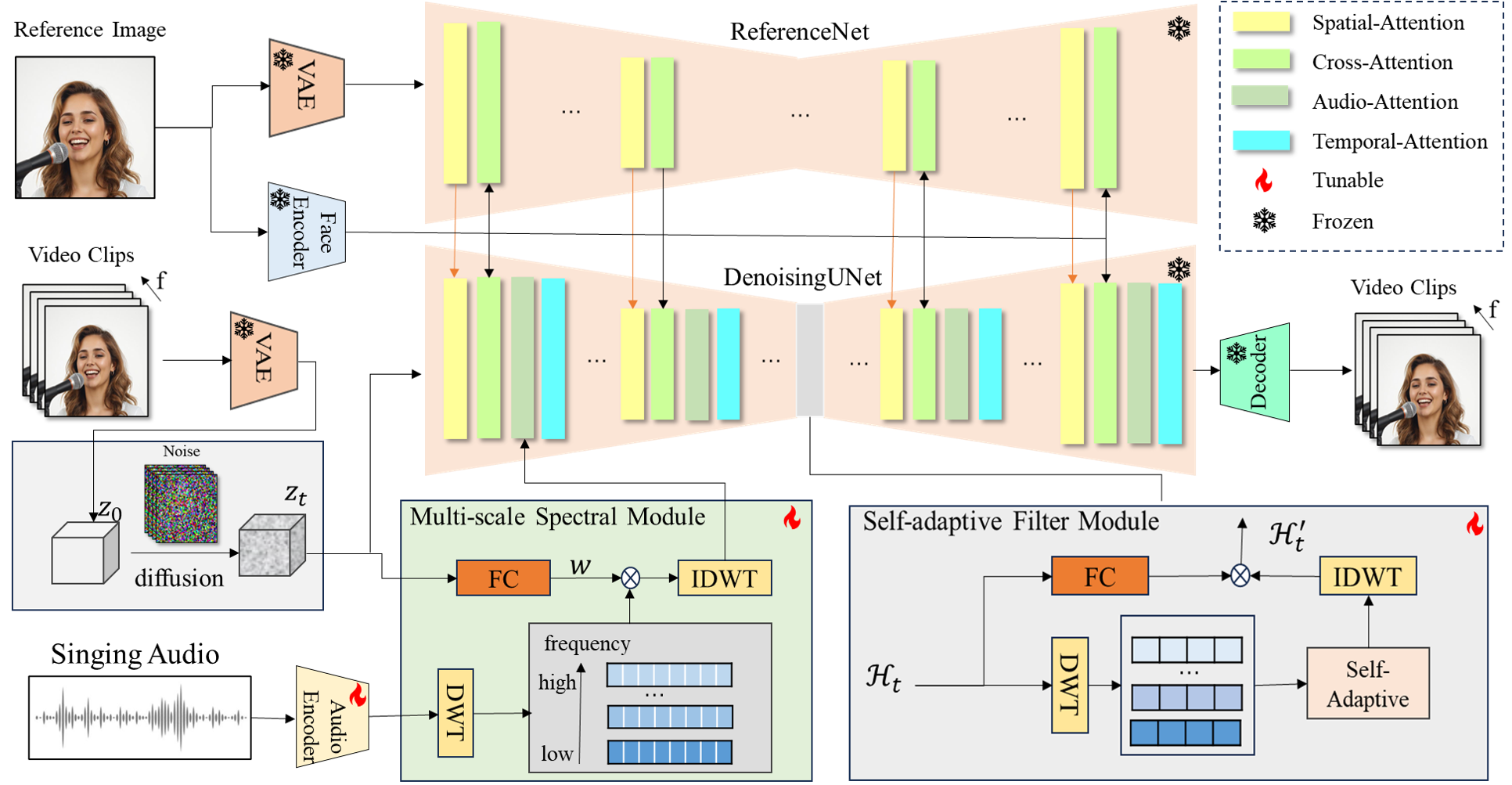}
    \caption{\textbf{The framework of \model.} (a) Training Process: Video clips are encoded and diffused to produce noised latent inputs for the Denoising UNet. The singing audio is processed through an audio encoder and the Multi-scale Spectral Module to capture multi-scale spectral patterns, generating an audio vector. This vector is integrated into the Denoising UNet via Audio-Attention. The reference image is encoded using both a VAE encoder and a Face encoder, with the resulting vectors integrated through Spatial-Attention and Cross-Attention into the Denoising UNet. Temporal-Attention handles dependencies along the temporal dimension. And the Self-adaptive Filter Module is put in the middle of the Denosing UNet to filter the patterns correlated with the multi-scale spectral patterns.
    The VAE decoder then reconstructs the output into a video clip. 
 (b) Inference Process: During inference, our \model~takes the driving singing audio and the reference image as inputs to generate vivid singing videos.}
    \label{fig:framework}
 \vspace{-8pt}   
\end{figure*}

To enhance audio-driven singing video generation, we propose \model, designed to generate vivid singing videos using the given singing audio and a reference image. 
The structure of \model~is illustrated in Figure~\ref{fig:framework}. Since singing involves two key aspects, audio patterns and behavioral expressions, we design dedicated modules to capture both effectively.
For learning audio patterns, we first apply a wavelet spectral transform to highlight key frequency and amplitude features of the input audio in the spectral domain~(see Section~\ref{sec:wavelettrans}). 
To capture these patterns across varying scales, we introduce a Multi-scale Spectral Module, enabling extraction of both global and fine-grained audio features~(see Section~\ref{sec:multiscale}). 
On the behavioral side, we design a self-adaptive Filter Module to dynamically identify and filter behavior patterns that are synchronized with the audio~(see Section~\ref{sec:selfadaptive}). 
This enables the generated facial expressions and movements to align naturally with the singing audio, resulting in lifelike and coherent singing videos. 

\subsection{Wavelet Spectral Transform}~\label{sec:wavelettrans}
The singing audio signal consists of multiple waves with varying frequencies and amplitudes, making it more complex than typical speech audio. Wavelet transforms offer the advantage of capturing information across time, location, and frequency simultaneously~\cite{adamowski2011wavelet}. Therefore, we apply wavelet transforms to map the audio vectors into the spectral domain, enabling the extraction of both spatial and frequency patterns from the singing audio. Among various wavelets, the Haar wavelet is widely used in real-world applications due to its simplicity and efficiency~\cite{phung2023wavelet}. It involves two key operations: discrete wavelet transform (DWT) and discrete inverse wavelet transform (IWT).

Given a singing audio \(S^l\) of length \(l\), it is first encoded by an audio encoder into a vector \(\mathcal{I} \in \mathbb{R}^{d_a \times l}\), where \(d_a\) is the embedding dimension. We then apply a 2D Haar wavelet transform to capture both low and high-frequency components. Let \(L\) and \(H\) represent the low-pass and high-pass filters, respectively:
\begin{equation}
    L = \frac{1}{\sqrt{2}}[1, 1],~ H = \frac{1}{\sqrt{2}}[-1, 1].
\end{equation}
Then four kernels of the Haar wavelet transform can be constructed using the low-pass filter \(L\) and high-pass filter \(H\). These kernels are denoted as \(\{LL^{\top}, LH^{\top}, HL^{\top}, HH^{\top}\}\). For simplicity, we refer to them as \(\{LL, LH, HL, HH\}\). By applying these kernels to the audio vector \(\mathcal{I}\), we obtain four distinct spectral representations, denoted as \(\{\mathcal{I}_{LL}, \mathcal{I}_{LH}, \mathcal{I}_{HL}, \mathcal{I}_{HH}\} \in \mathbb{R}^{\frac{d_a}{2} \times \frac{l}{2}}\) .

Different transformed spectral representations capture different levels of information. Specifically, \(\mathcal{I}_{LL}\) contains the overall texture and global features of the audio, offering a coarse yet holistic representation. In contrast, \(\{\mathcal{I}_{LH}, \mathcal{I}_{HL}, \mathcal{I}_{HH}\}\) encode local statistical features, capturing finer, high-frequency details across different orientations. This multi-resolution decomposition highlights different audio aspects to facilitate the learning of both global and local patterns critical for generating lifelike singing videos. Note that, since these filters are pairwise orthogonal, they can form invertible matrices, enabling accurate reconstruction of the original signals from the decomposed spectral features. 
The invertibility of the wavelet transform ensures precise forward decomposition and inverse transformation, maintaining consistency between the learned spectral representations and the original audio input. 

\subsection{Multi-scale Spectral Pattern Fusion}~\label{sec:multiscale}
After decomposing the audio vectors into different spectral representations, each representation captures features at varying levels of detail. To automatically determine which features are most relevant for singing video generation, we propose a fusion mechanism that incorporates the visual information from the video clips. By jointly analyzing the audio and visual features, this fusion process allows the model to dynamically identify and prioritize the spectral features that are most important for generating accurate and synchronized singing videos. This joint feature selection ensures that the generated video aligns seamlessly with the nuances of the input audio, enhancing the realism and coherence of facial expressions and movements.

During the training process, the video clips are encoded into a latent representation \(z_0 \in \mathbb{R}^{f \times c \times d_w \times d_h}\), where \(f\) is the number of frames, \(c\) is the number of channels, and \(d_w\) and \(d_h\) are the width and height of the encoded frames, respectively. 
By applying the diffusion process to this latent, we obtain a noisy latent \(z_t \in \mathbb{R}^{f \times c \times d_w \times d_h}\) at time step \(t\), which serves as the input to the Denoising UNet. 
We first use a tunable matrix $w$ with the same size as $z_t$ to derive the initial weights:
\begin{equation}
    w_z = w*z_t,
\end{equation}
where \(*\) denotes element-wise multiplication. Next, we divide \(w_z\) into four equal-sized chunks along the width dimension, yielding \(\{w_z^1, w_z^2, w_z^3, w_z^4\} \in \mathbb{R}^{f \times c \times (d_w / 4) \times d_h}\). We further design a two-layer fully connected model, denoted as \(FC\), to process the four chunked weights. Each of the chunks \(\{w_z^1, w_z^2, w_z^3, w_z^4\}\) is fed into the fully connected model to obtain the final weights:
\begin{equation}
    \hat{w}_z = \hat{w}_z^1, \hat{w}_z^2, \hat{w}_z^3, \hat{w}_z^4 = FC(\mathcal{C}(w_z^1, w_z^2, w_z^3, w_z^4)),
\end{equation}
where \(\mathcal{C}\) denotes the concatenation operation, and \(\hat{w}_z^{.} \in \mathbb{R}\) represents the obtained weights after processing through the fully connected model. The four resulting weights \(\{\hat{w}_z^1, \hat{w}_z^2, \hat{w}_z^3, \hat{w}_z^4\}\) are used to determine the degree of importance of the corresponding spectral representations:
\begin{equation}
   \hat{\mathcal{I}} = \{\hat{w}_z^1\mathcal{I}_{LL}, \hat{w}_z^2\mathcal{I}_{LH}, \hat{w}_z^3\mathcal{I}_{HL}, \hat{w}_z^4\mathcal{I}_{HH}\}.
\end{equation}
By applying these weights to the spectral features, we can prioritize certain aspects of the audio that are more relevant for video generation.

Tunable weights determine the importance of different spectral representations, allowing the model to emphasize patterns crucial for singing video generation.  
After obtaining the weighted multi-scale spectral representation \(\hat{\mathcal{I}}\), we apply the inverse wavelet transform to reconstruct an audio vector $\hat{S}^l$ that encapsulates these multi-scale spectral patterns. The reconstructed audio vector $\hat{S}^l$ is integrated into the Denoising UNet using the Audio-Attention mechanism. This integration allows the model to focus on relevant audio features while denoising the video frames, facilitating the learning of useful patterns that are essential for generating high-quality singing videos.

\subsection{Self-adaptive Feature Filter}~\label{sec:selfadaptive}
The Denoising UNet used in \model~employs an encoder-decoder structure. The first part, the encoder, is responsible for capturing contextual information and extracting features from the input data. The second part, known as the decoder, reconstructs the output from these extracted features~\cite{polat2023diagnostic}. To ensure that the reconstruction from the decoder emphasizes the important singing patterns identified in Section~\ref{sec:multiscale}, we design a Self-adaptive Filter Module. This module filters the features extracted by the encoder, dynamically selecting and enhancing the most relevant patterns for singing video generation, ensuring that the final output reflects the crucial elements of the singing performance while maintaining high fidelity and coherence. 

Denote the output of the encoder as \(\mathcal{H}_t \in \mathbb{R}^{f \times c_a \times hw \times hh}\) at time step \(t\), where \(f\) is the number of frames, \(c_a\) is the number of channels, and \(hw\) and \(hh\) are the width and height of the cross-attention mechanism, respectively. First, we apply a Haar wavelet transform to \(\mathcal{H}_t\), decomposing it into four sub-bands:
\begin{equation}
  \Tilde{\mathcal{H}}_t = \{\mathcal{H}_{t,i} \in \mathbb{R}^{f \times c_a \times \frac{hw}{2} \times \frac{hh}{2}} \}_{i \in \{LL, LH, HL, HH\}}.
\end{equation}
Next, we introduce the tunable weights \(w_h =\{w_h^{i}\in \mathbb{R}^{ f \times c_a \times \frac{hw}{2} \times \frac{hh}{2}}\}_{i \in \{LL, LH, HL, HH\}}\) to adaptively adjust the contributions of each sub-band:
\begin{equation}
  \hat{\mathcal{H}}_{t} = \{ w_h^i * \mathcal{H}_{t, i} \}_{i \in \{LL, LH, HL, HH\}}.
\end{equation}
This allows the model to emphasize specific frequency components that are most relevant to the task at hand, ensuring that the most important features are prioritized during the reconstruction process.
Next, we apply the inverse wavelet transform to the weighted sub-bands to obtain the weighted extracted features \(\hat{\mathcal{H}}^{'}_t \in \mathbb{R}^{f \times c_a \times hw \times hh}\). This reconstruction ensures that the most relevant patterns from the audio are preserved and emphasized. To further assess the important patterns within the extracted features, we calculate the attention score matrix \(w_a\) from \(\mathcal{H}_t\) as follows:
\begin{equation}
w_a = \text{sigmoid}(FC(\mathcal{H}_t)),
\end{equation}
where \(w_a \in \mathbb{R}^{f \times c_a \times hw \times hh}\). Here, the \(FC\) denotes the fully connected layer that processes the extracted features, and the sigmoid function ensures that the attention scores are scaled between 0 and 1. This attention score matrix \(w_a\) effectively highlights the significance of various features, guiding the model to focus on the most crucial aspects during the generation of the singing videos. We apply the attention scores to the obtained weighted extracted features \(\hat{\mathcal{H}}_{t}\) to obtain the final features:
\begin{equation}
\mathcal{H}^{'}_{t} = w_a * \hat{\mathcal{H}}_{t}^{'},
\end{equation}
where \( * \) denotes element-wise multiplication. This operation ensures that the decoder focuses on the most significant audio-visual features, enhancing the model's ability to generate synchronized and realistic singing videos.

By channeling these final features into the decoder, we enable it to prioritize critical patterns, ultimately improving the coherence and quality of the output. This process not only facilitates better alignment between audio and visual elements but also enhances the overall viewer experience by producing lifelike singing performances.

\section{Experiment} \label{sec:exp}

\begin{figure*}[h]
    \begin{minipage}{0.02\linewidth}
    \centering
        \rotatebox{90}{GT}
    \end{minipage}
    \begin{minipage}{0.97\linewidth}
    \begin{subfigure}{0.12\linewidth}
        \includegraphics[width=\linewidth]{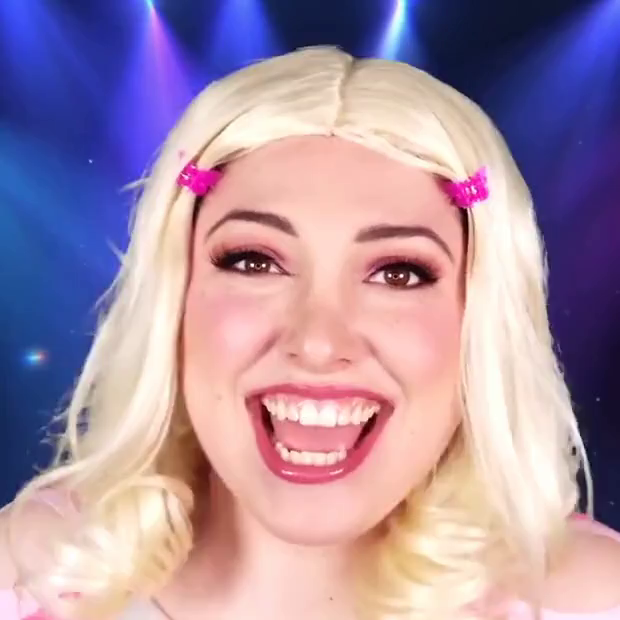}
    \end{subfigure}
    \hspace{-4pt}
        \begin{subfigure}{0.12\linewidth}
        \includegraphics[width=\linewidth]{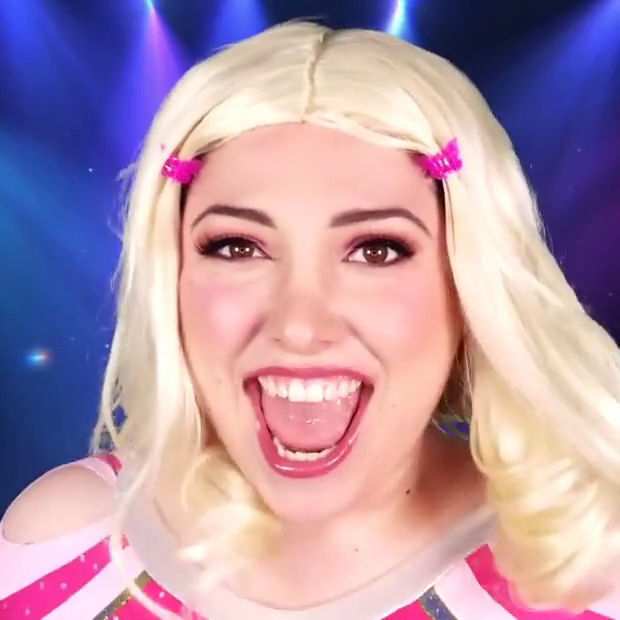}
    \end{subfigure}
     \hspace{-4pt}
        \begin{subfigure}{0.12\linewidth}
        \includegraphics[width=\linewidth]{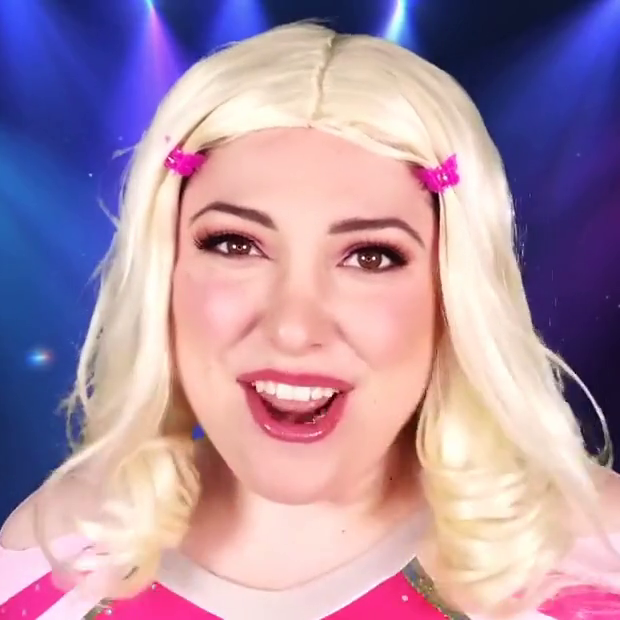}
    \end{subfigure}
     \hspace{-4pt}
        \begin{subfigure}{0.12\linewidth}
        \includegraphics[width=\linewidth]{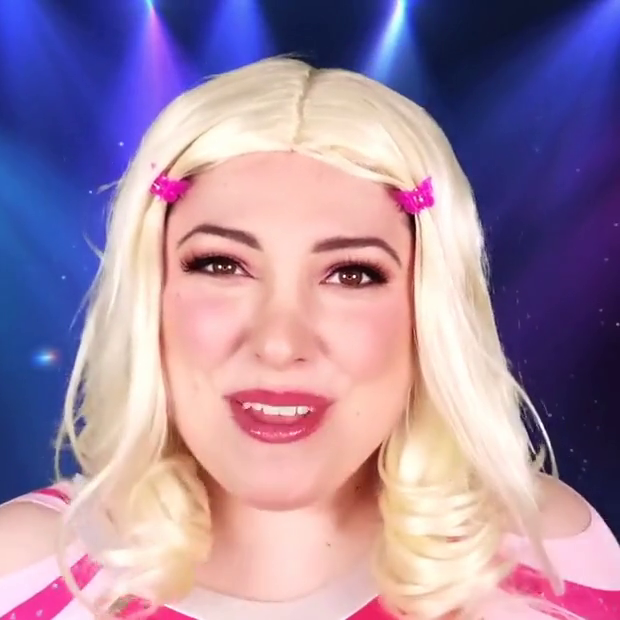}
    \end{subfigure}
     \hspace{-4pt}
        \begin{subfigure}{0.12\linewidth}
        \includegraphics[width=\linewidth]{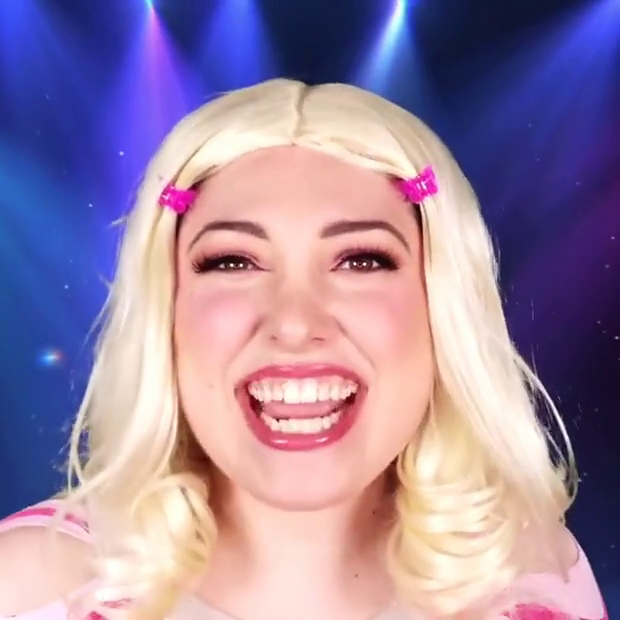}
    \end{subfigure}
     \hspace{-4pt}
        \begin{subfigure}{0.12\linewidth}
        \includegraphics[width=\linewidth]{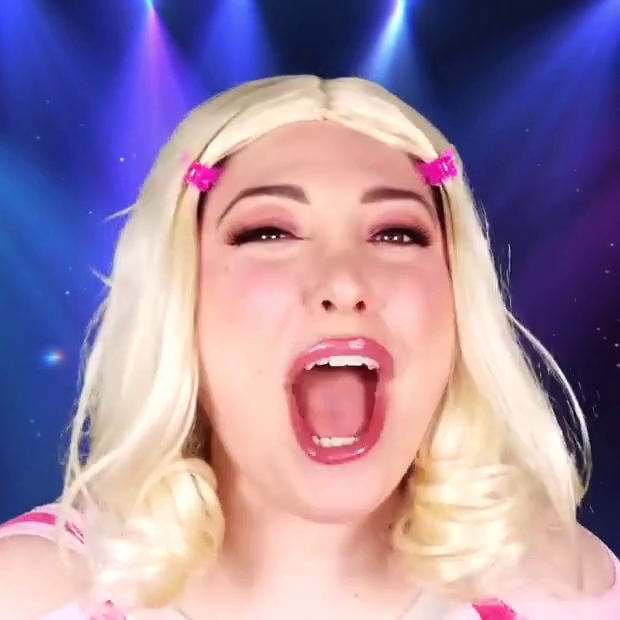}
    \end{subfigure}
    \hspace{-4pt}
        \begin{subfigure}{0.12\linewidth}
        \includegraphics[width=\linewidth]{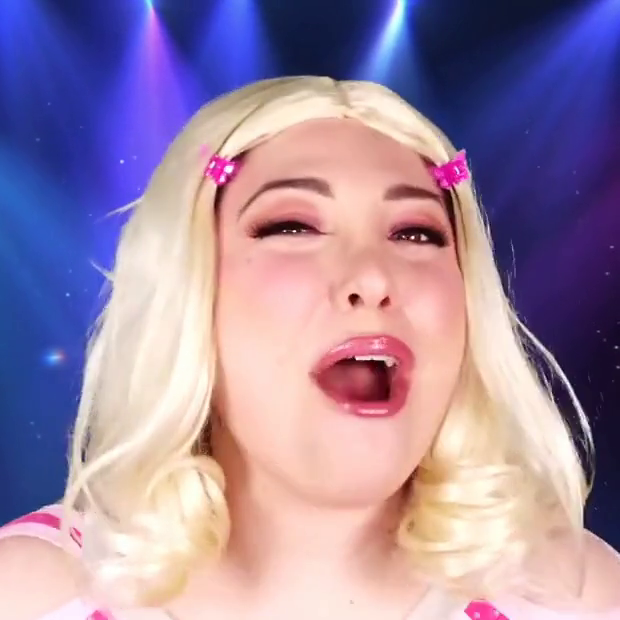}
    \end{subfigure}
    \hspace{-4pt}
        \begin{subfigure}{0.12\linewidth}
        \includegraphics[width=\linewidth]{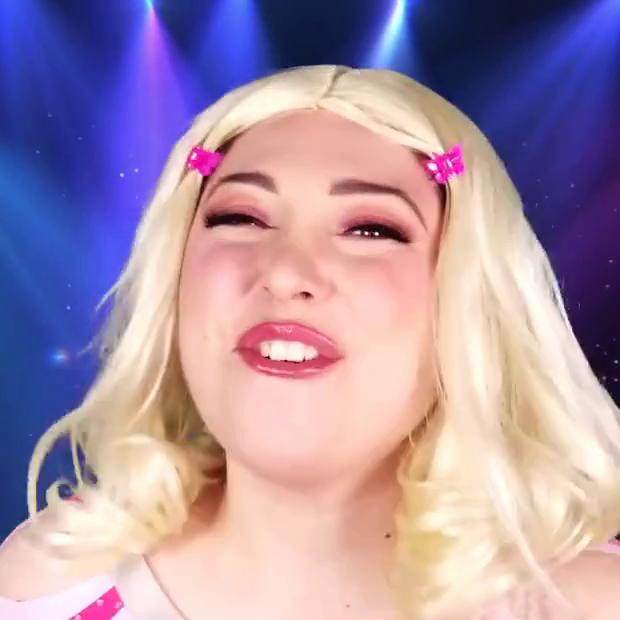}
    \end{subfigure}
    \end{minipage}

      \begin{minipage}{0.02\linewidth}
    \centering
        \rotatebox{90}{AniPortrait}
    \end{minipage}
    \begin{minipage}{0.97\linewidth}
    \begin{subfigure}{0.12\linewidth}
        \includegraphics[width=\linewidth]{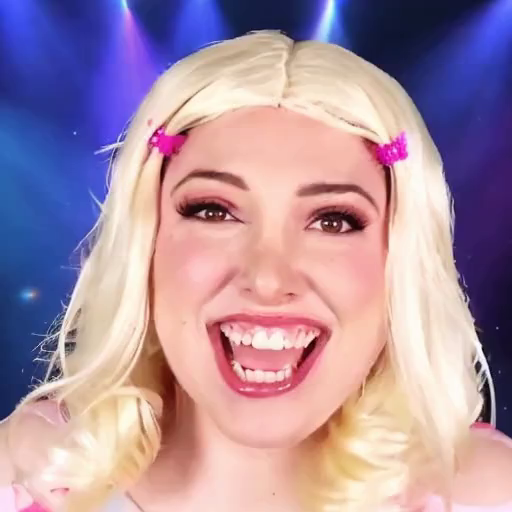}
    \end{subfigure}
    \hspace{-4pt}
        \begin{subfigure}{0.12\linewidth}
        \includegraphics[width=\linewidth]{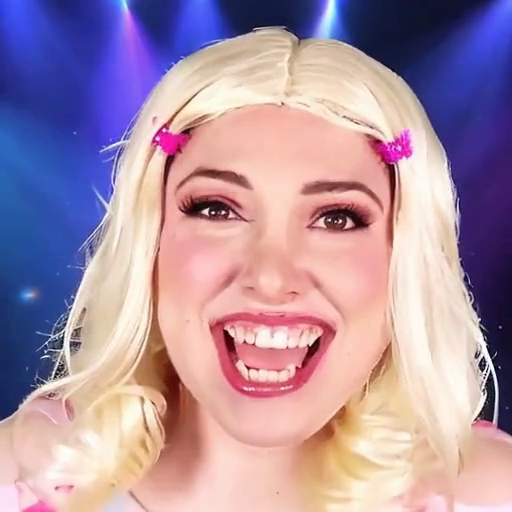}
    \end{subfigure}
     \hspace{-4pt}
        \begin{subfigure}{0.12\linewidth}
        \includegraphics[width=\linewidth]{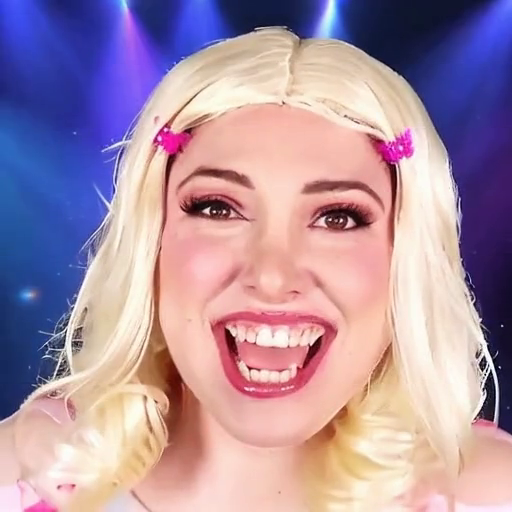}
    \end{subfigure}
     \hspace{-4pt}
        \begin{subfigure}{0.12\linewidth}
        \includegraphics[width=\linewidth]{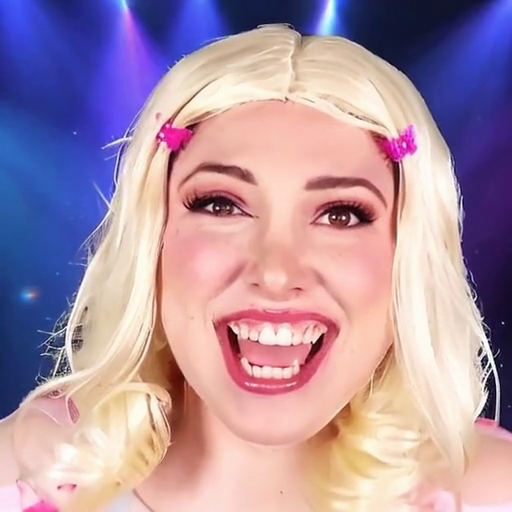}
    \end{subfigure}
     \hspace{-4pt}
        \begin{subfigure}{0.12\linewidth}
        \includegraphics[width=\linewidth]{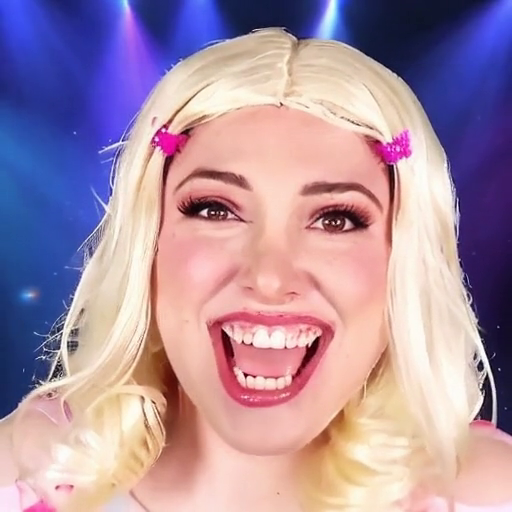}
    \end{subfigure}
     \hspace{-4pt}
        \begin{subfigure}{0.12\linewidth}
        \includegraphics[width=\linewidth]{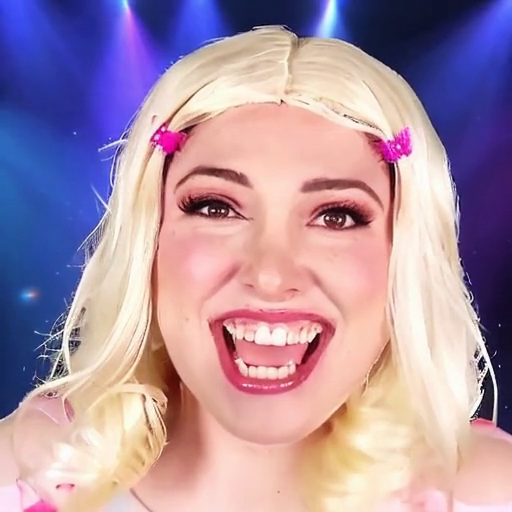}
    \end{subfigure}
    \hspace{-4pt}
        \begin{subfigure}{0.12\linewidth}
        \includegraphics[width=\linewidth]{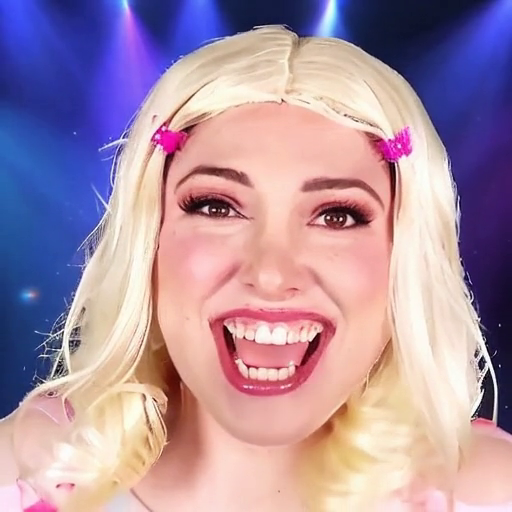}
    \end{subfigure}
    \hspace{-4pt}
        \begin{subfigure}{0.12\linewidth}
        \includegraphics[width=\linewidth]{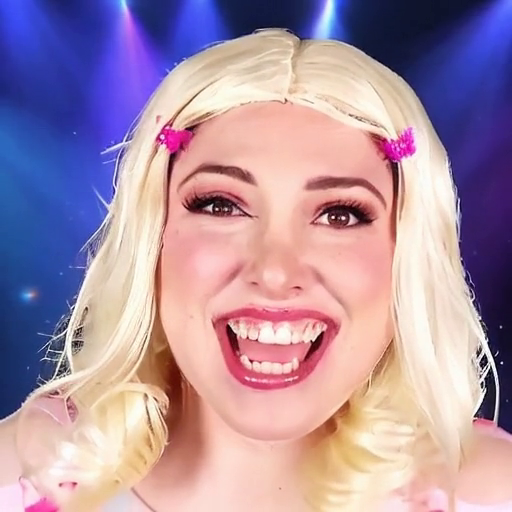}
    \end{subfigure}
    \end{minipage}
      \begin{minipage}{0.02\linewidth}
    \centering
        \rotatebox{90}{Echomimic}
    \end{minipage}
    \begin{minipage}{0.97\linewidth}
    \begin{subfigure}{0.12\linewidth}
        \includegraphics[width=\linewidth]{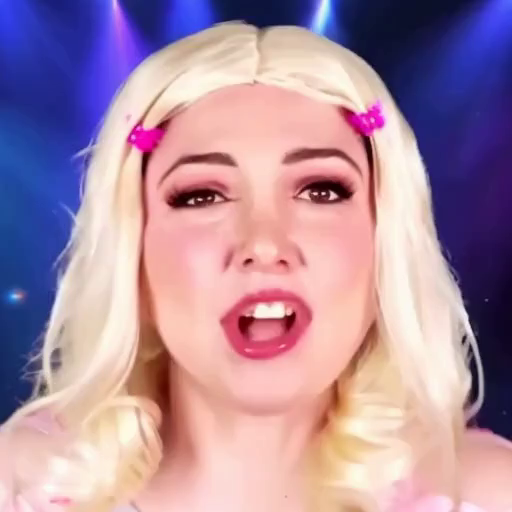}
    \end{subfigure}
    \hspace{-4pt}
        \begin{subfigure}{0.12\linewidth}
        \includegraphics[width=\linewidth]{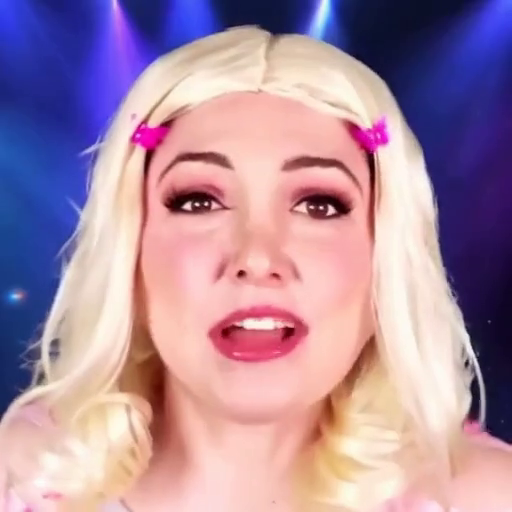}
    \end{subfigure}
     \hspace{-4pt}
        \begin{subfigure}{0.12\linewidth}
        \includegraphics[width=\linewidth]{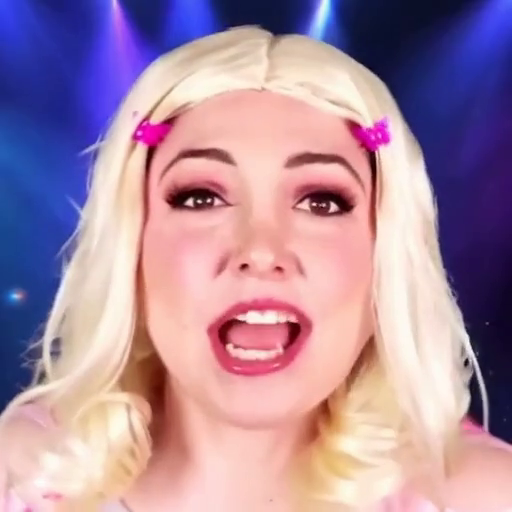}
    \end{subfigure}
     \hspace{-4pt}
        \begin{subfigure}{0.12\linewidth}
        \includegraphics[width=\linewidth]{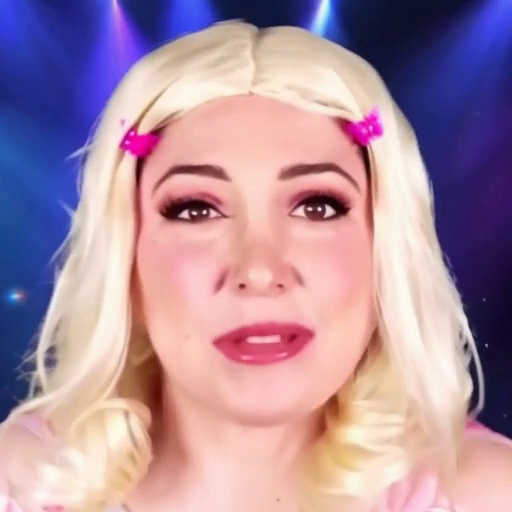}
    \end{subfigure}
     \hspace{-4pt}
        \begin{subfigure}{0.12\linewidth}
        \includegraphics[width=\linewidth]{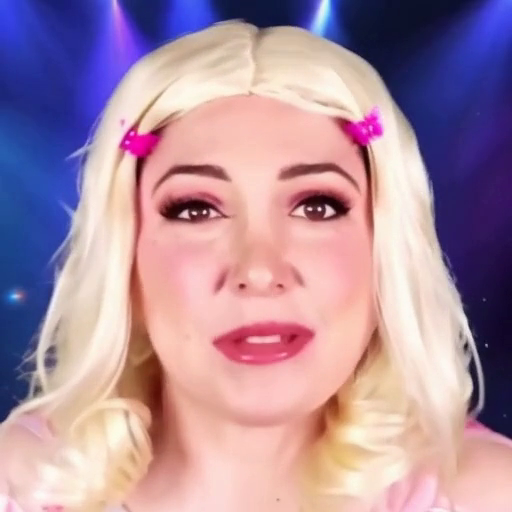}
    \end{subfigure}
     \hspace{-4pt}
        \begin{subfigure}{0.12\linewidth}
        \includegraphics[width=\linewidth]{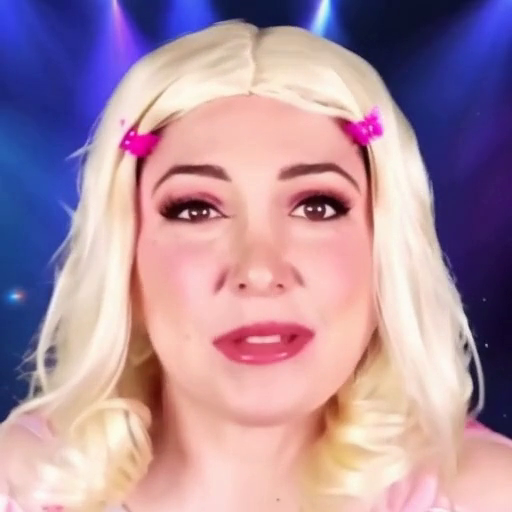}
    \end{subfigure}
    \hspace{-4pt}
        \begin{subfigure}{0.12\linewidth}
        \includegraphics[width=\linewidth]{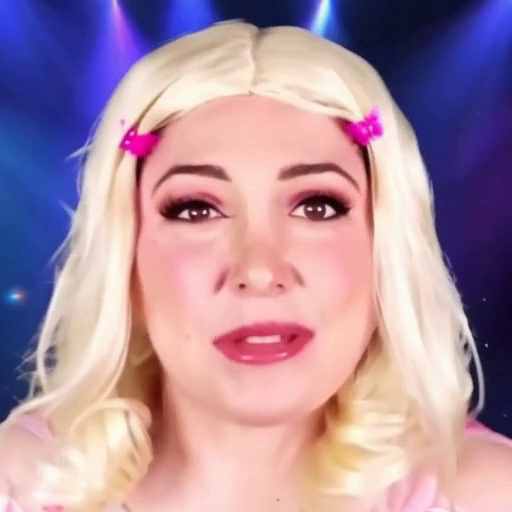}
    \end{subfigure}
    \hspace{-4pt}
        \begin{subfigure}{0.12\linewidth}
        \includegraphics[width=\linewidth]{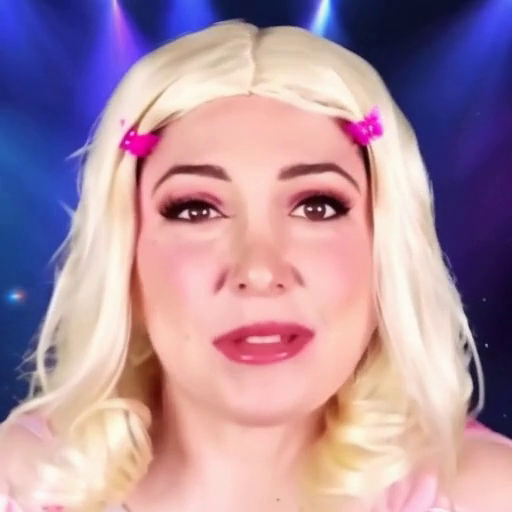}
    \end{subfigure}
    \end{minipage}
      \begin{minipage}{0.02\linewidth}
    \centering
        \rotatebox{90}{Hallo}
    \end{minipage}
    \begin{minipage}{0.97\linewidth}
    \begin{subfigure}{0.12\linewidth}
        \includegraphics[width=\linewidth]{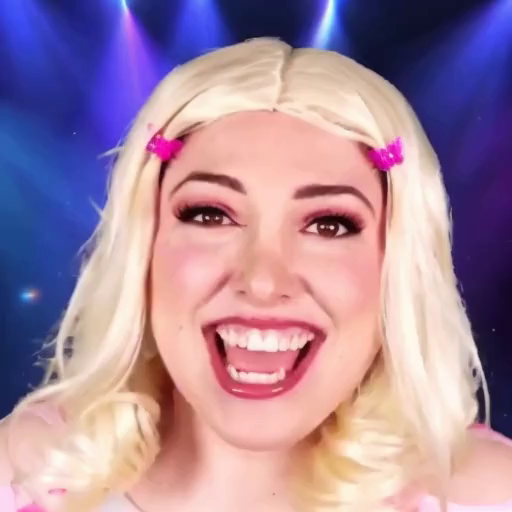}
    \end{subfigure}
    \hspace{-4pt}
        \begin{subfigure}{0.12\linewidth}
        \includegraphics[width=\linewidth]{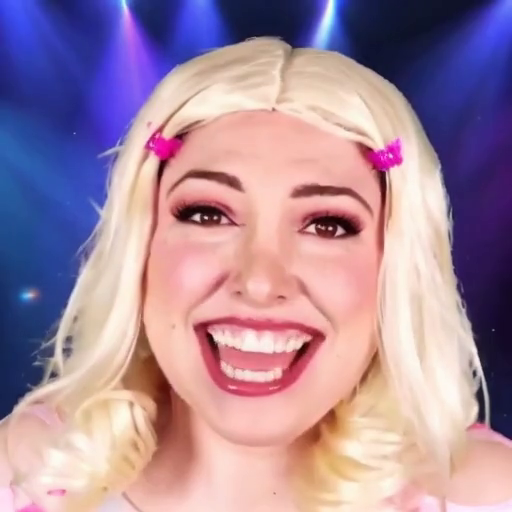}
    \end{subfigure}
     \hspace{-4pt}
        \begin{subfigure}{0.12\linewidth}
        \includegraphics[width=\linewidth]{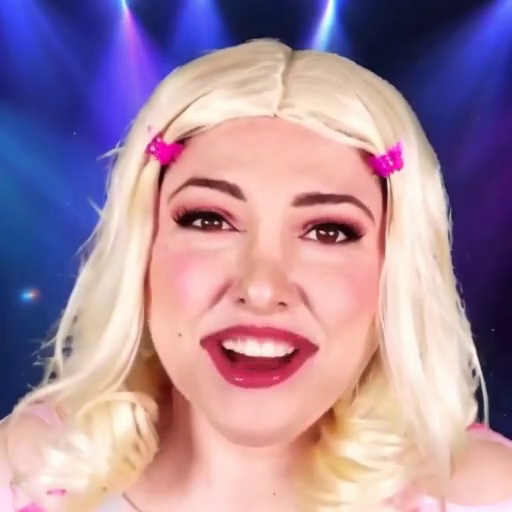}
    \end{subfigure}
     \hspace{-4pt}
        \begin{subfigure}{0.12\linewidth}
        \includegraphics[width=\linewidth]{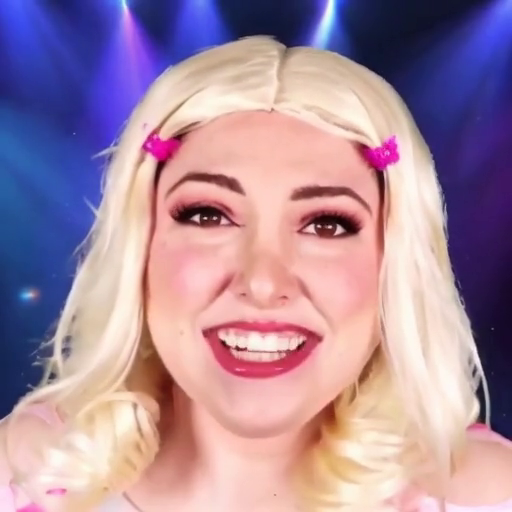}
    \end{subfigure}
     \hspace{-4pt}
        \begin{subfigure}{0.12\linewidth}
        \includegraphics[width=\linewidth]{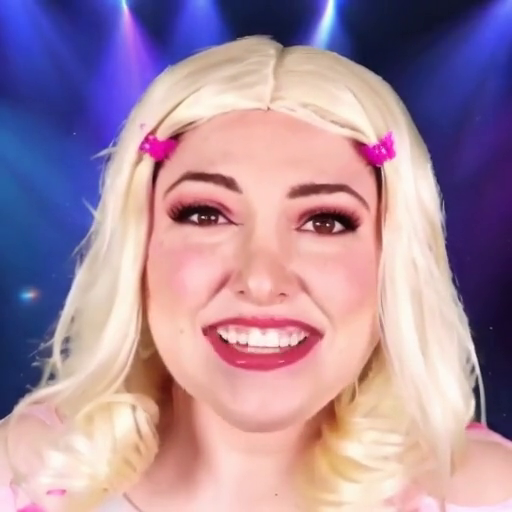}
    \end{subfigure}
     \hspace{-4pt}
        \begin{subfigure}{0.12\linewidth}
        \includegraphics[width=\linewidth]{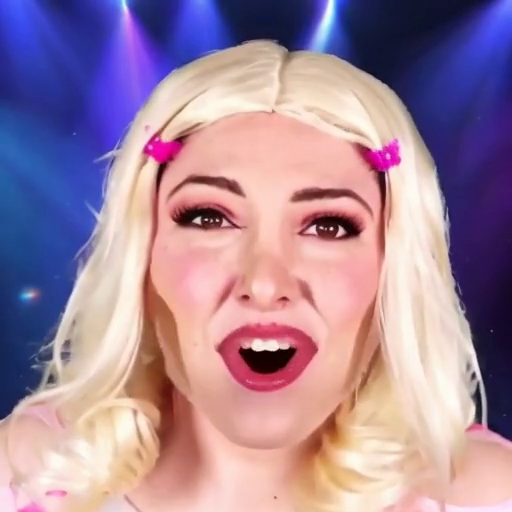}
    \end{subfigure}
    \hspace{-4pt}
        \begin{subfigure}{0.12\linewidth}
        \includegraphics[width=\linewidth]{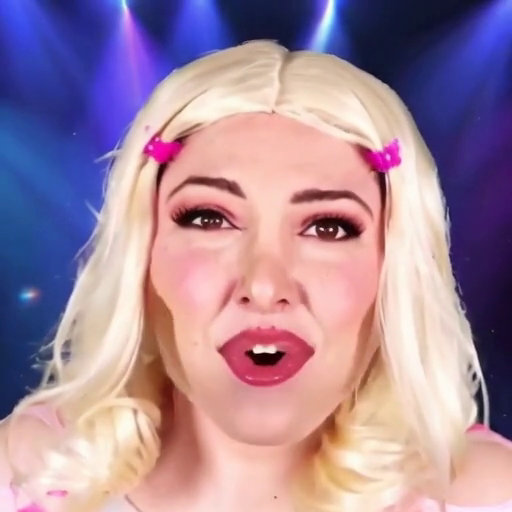}
    \end{subfigure}
    \hspace{-4pt}
        \begin{subfigure}{0.12\linewidth}
        \includegraphics[width=\linewidth]{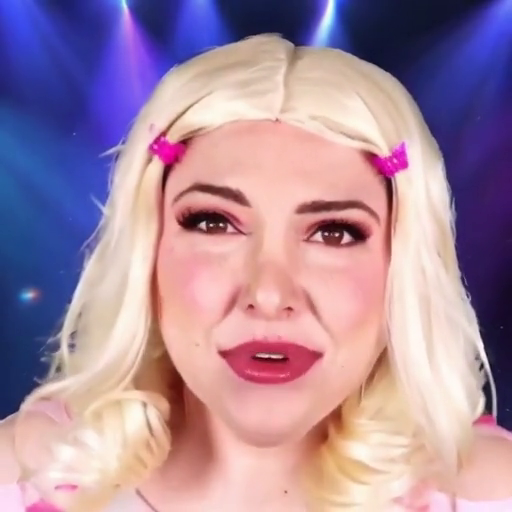}
    \end{subfigure}
    \end{minipage}
      \begin{minipage}{0.02\linewidth}
    \centering
        \rotatebox{90}{Hallo2}
    \end{minipage}
    \begin{minipage}{0.97\linewidth}
    \begin{subfigure}{0.12\linewidth}
        \includegraphics[width=\linewidth]{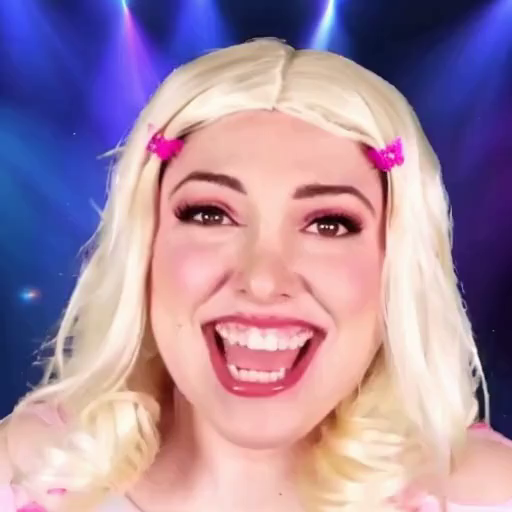}
    \end{subfigure}
    \hspace{-4pt}
        \begin{subfigure}{0.12\linewidth}
        \includegraphics[width=\linewidth]{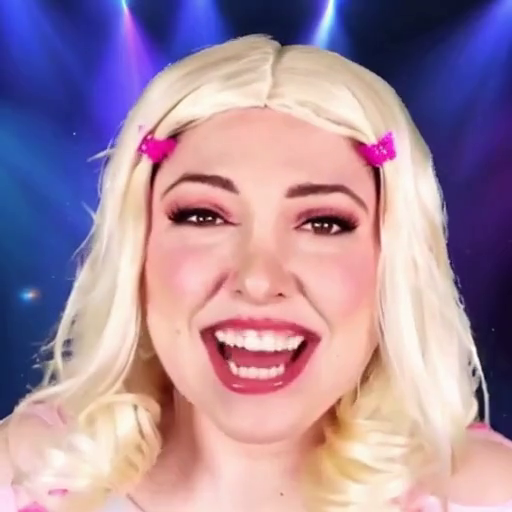}
    \end{subfigure}
     \hspace{-4pt}
        \begin{subfigure}{0.12\linewidth}
        \includegraphics[width=\linewidth]{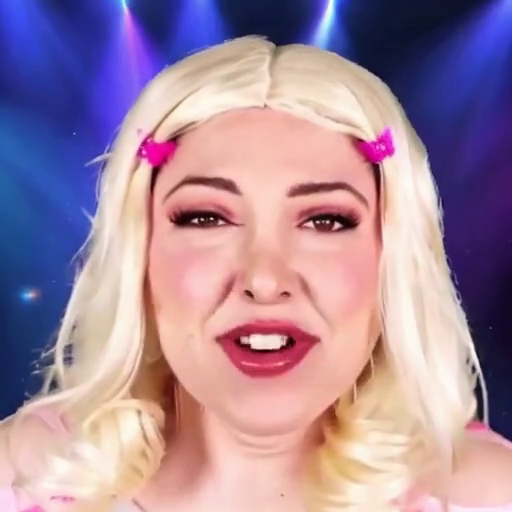}
    \end{subfigure}
     \hspace{-4pt}
        \begin{subfigure}{0.12\linewidth}
        \includegraphics[width=\linewidth]{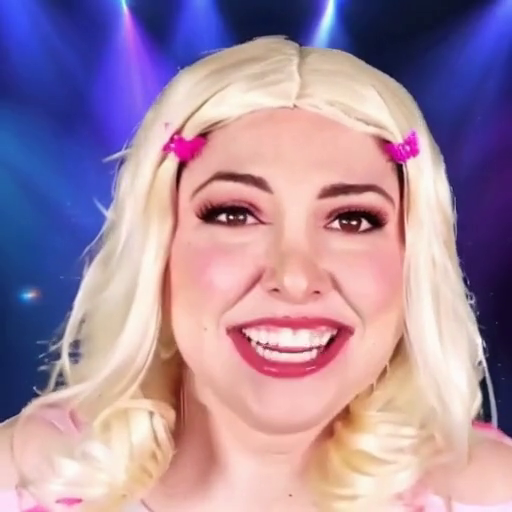}
    \end{subfigure}
     \hspace{-4pt}
        \begin{subfigure}{0.12\linewidth}
        \includegraphics[width=\linewidth]{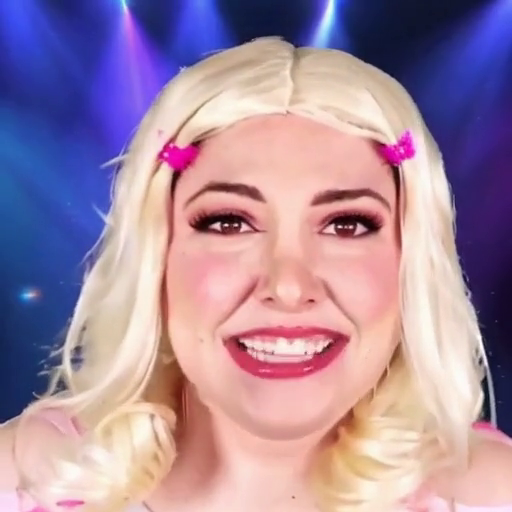}
    \end{subfigure}
     \hspace{-4pt}
        \begin{subfigure}{0.12\linewidth}
        \includegraphics[width=\linewidth]{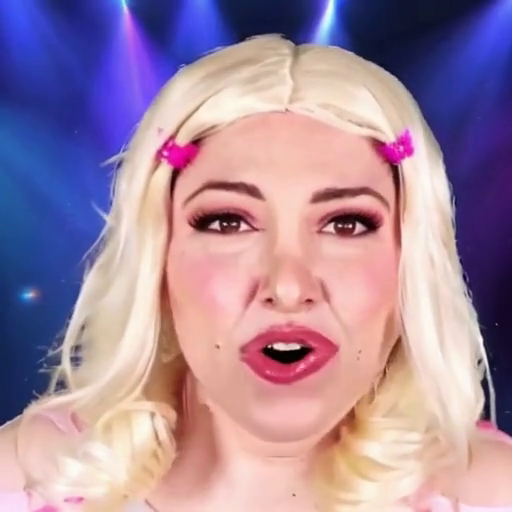}
    \end{subfigure}
    \hspace{-4pt}
        \begin{subfigure}{0.12\linewidth}
        \includegraphics[width=\linewidth]{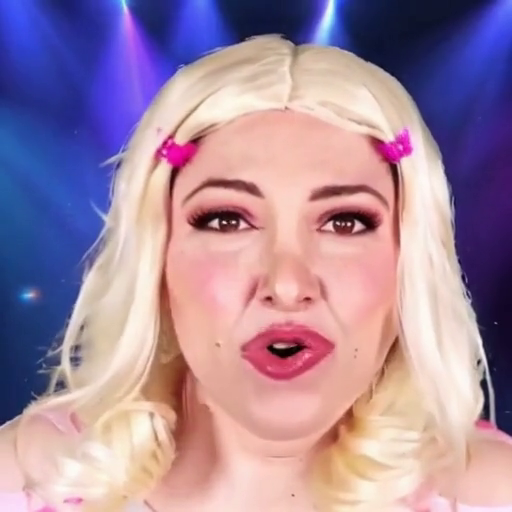}
    \end{subfigure}
    \hspace{-4pt}
        \begin{subfigure}{0.12\linewidth}
        \includegraphics[width=\linewidth]{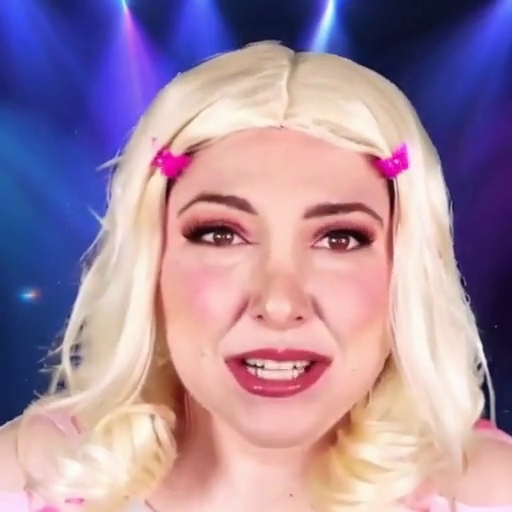}
    \end{subfigure}
    \end{minipage}
  
    \begin{minipage}{0.02\linewidth}
    \centering
        \rotatebox{90}{\model}
    \end{minipage}
    \begin{minipage}{0.97\linewidth}
    \begin{subfigure}{0.12\linewidth}
        \includegraphics[width=\linewidth]{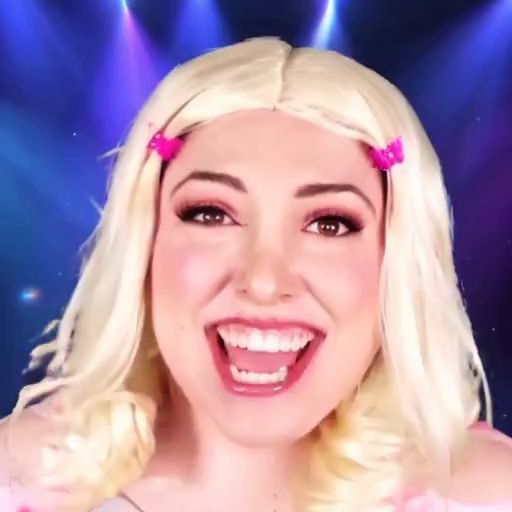}
    \end{subfigure}
    \hspace{-4pt}
        \begin{subfigure}{0.12\linewidth}
        \includegraphics[width=\linewidth]{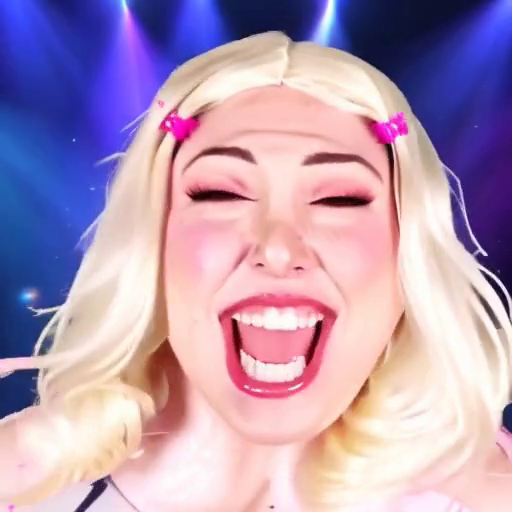}
    \end{subfigure}
     \hspace{-4pt}
        \begin{subfigure}{0.12\linewidth}
        \includegraphics[width=\linewidth]{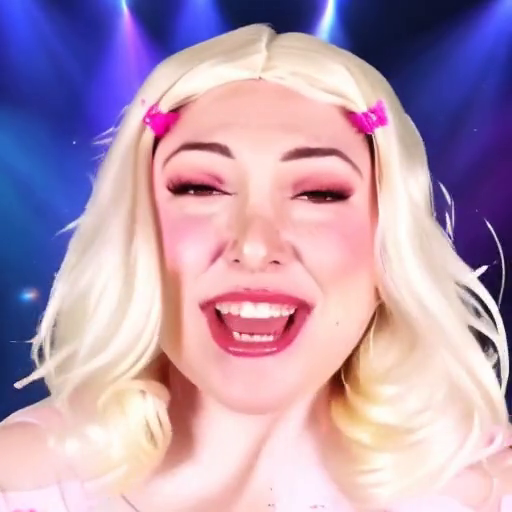}
    \end{subfigure}
     \hspace{-4pt}
        \begin{subfigure}{0.12\linewidth}
        \includegraphics[width=\linewidth]{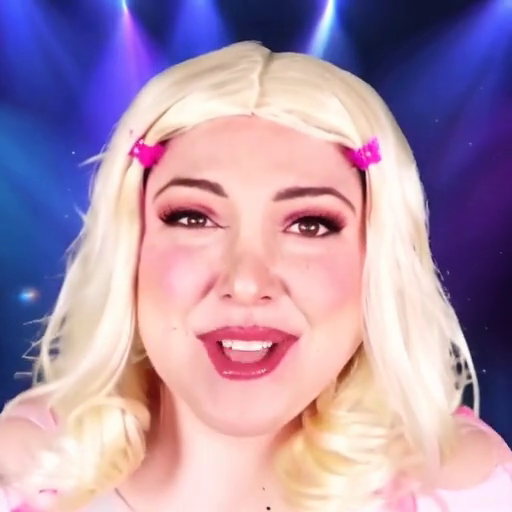}
    \end{subfigure}
     \hspace{-4pt}
        \begin{subfigure}{0.12\linewidth}
        \includegraphics[width=\linewidth]{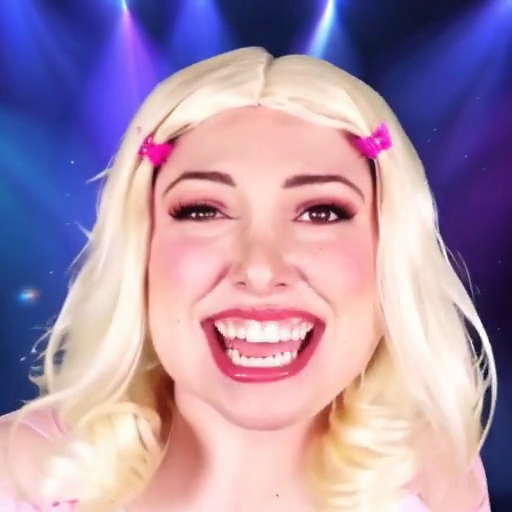}
    \end{subfigure}
     \hspace{-4pt}
        \begin{subfigure}{0.12\linewidth}
        \includegraphics[width=\linewidth]{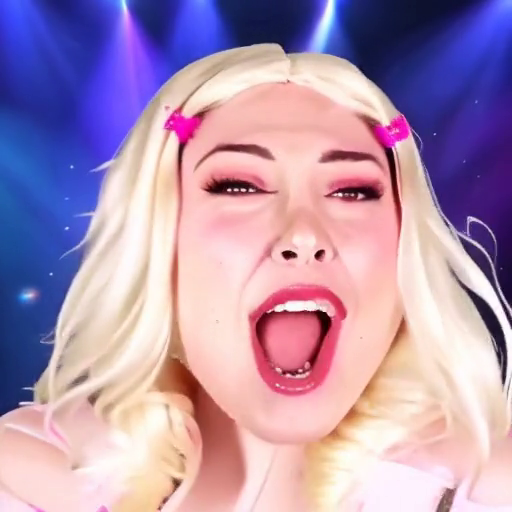}
    \end{subfigure}
    \hspace{-4pt}
        \begin{subfigure}{0.12\linewidth}
        \includegraphics[width=\linewidth]{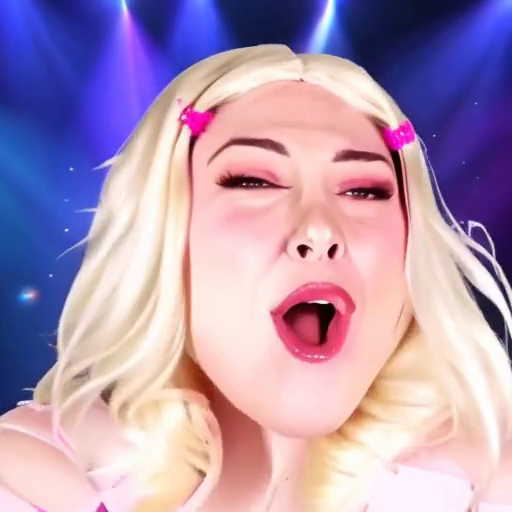}
    \end{subfigure}
    \hspace{-4pt}
        \begin{subfigure}{0.12\linewidth}
        \includegraphics[width=\linewidth]{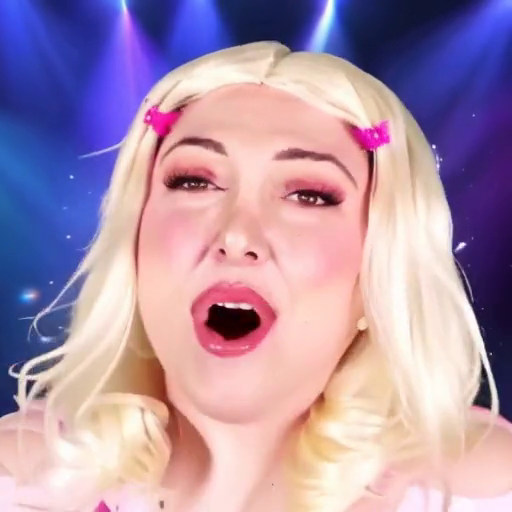}
    \end{subfigure}
    \end{minipage}
    \caption{Visualization of generated singing videos by different generative methods. ``GT" denotes the ground truth singing video.}
    \label{fig:main_comparison}
    \vspace{-8pt}
\end{figure*}
\begin{table*}[h]
    \centering
    \small
      \caption{Comparison between \model~and seven state-of-the-art methods in generating singing videos on the collected SHV dataset.}
    \resizebox{\linewidth}{!}{
    \begin{tabular}{c|cccc|ccc|cc}
    \toprule
      \multirow{2}{*}{Method}   & \multicolumn{4}{c|}{\textbf{Video Quality}} & \multicolumn{3}{c|}{\textbf{Lip Synchronization}} & \multicolumn{2}{c}{\textbf{Motion}}\\
      ~ & SSIM \((\uparrow) \)& PSNR  \((\uparrow) \) & CPBD  \((\uparrow) \) & FVD  \((\downarrow) \)& LMD  \((\downarrow) \)& LSE-D  \((\downarrow) \)& LSE-C  \((\uparrow) \) & Diversity  \((\uparrow) \) & BAS  \((\uparrow) \)\\
      \midrule
       Audio2Head  &  0.4896 & 28.281 & 0.4469 & 1089.7 & 75.791 & 9.1998 & 1.2458 & 10.846 & 0.1982\\
       SadTalker & 0.4134 & 29.872 & 0.5509 & 1030.3 & 59.625 & 9.1739 & 1.2454 & 14.253 & 0.1774 \\
       MuseTalk & 0.5762 & 30.243 & 0.5266 & 1323.2 & 64.669 & 10.143 & 1.0545 & 1.4802 & 0.2400 \\
       \midrule
       AniPortrait & 0.5364 &	29.872 & 0.5509 & 1030.3 & 76.442 & 10.226 &	0.8667 & 5.4195 & 0.2296\\
       Echomimic & 0.4035 & 28.865	&0.4896 & 1221.7 & 73.163 &	9.8249 &	1.2290 & 11.732 & 0.1483\\
       Hallo & 0.5722 & 29.984 & 0.5486 &897.65 & 64.346 &	9.1645 &	\textbf{1.7012} & 8.5367 & 0.1850\\
       Hallo2 & 0.5659 &	30.058 & \textbf{0.5558} & 1478.2 & 60.741	& 9.6107	& 1.5967 & 9.2739 & 0.2184\\
       \midrule
       \model~ & \textbf{0.6364} &	\textbf{30.686} &	0.5430 & \textbf{503.78} & \textbf{53.373}	& \textbf{9.1269}	& 1.6209 &\textbf{ 14.445}	& \textbf{0.2405}\\
       \midrule
       GT & - & - &0.5338 & 0.0000 & 0.0000 &8.5541 & 4.5286 &21.754 & 0.2484\\
    \bottomrule
    \end{tabular}
    }
    \label{tab:main_results}
    \vspace{-8pt}
\end{table*}

\begin{figure*}[!h]
    \centering
    \begin{minipage}{0.49\linewidth} 
    \centering
    \begin{minipage}{0.04\linewidth}
        \rotatebox{90}{GT}
    \end{minipage}
    \begin{minipage}{0.95\linewidth}
    \begin{subfigure}{0.23\linewidth}
        \includegraphics[width=\linewidth]{figures/gt/0010.png}
    \end{subfigure}
    \hspace{-4pt}
     \begin{subfigure}{0.23\linewidth}
        \includegraphics[width=\linewidth]{figures/gt/0022.png}
    \end{subfigure}
    \hspace{-4pt}
      \begin{subfigure}{0.23\linewidth}
        \includegraphics[width=\linewidth]{figures/gt/0031.png}
    \end{subfigure}
    \hspace{-4pt}
      \begin{subfigure}{0.23\linewidth}
        \includegraphics[width=\linewidth]{figures/gt/0036.png}
    \end{subfigure}
    \end{minipage}

    \begin{minipage}{0.04\linewidth}
    \centering
        \rotatebox{90}{Subject1}
    \end{minipage}
    \begin{minipage}{0.95\linewidth}
    \begin{subfigure}{0.23\linewidth}
        \includegraphics[width=\linewidth]{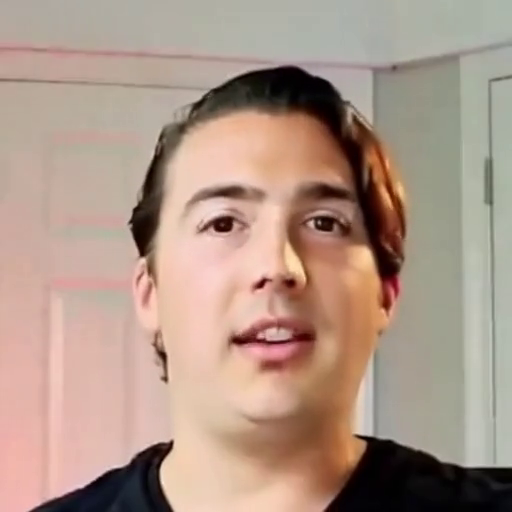}
    \end{subfigure}
    \hspace{-4pt}
     \begin{subfigure}{0.23\linewidth}
        \includegraphics[width=\linewidth]{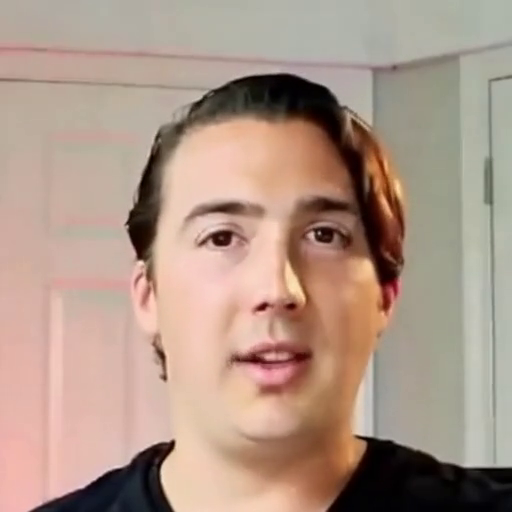}
    \end{subfigure}
    \hspace{-4pt}
      \begin{subfigure}{0.23\linewidth}
        \includegraphics[width=\linewidth]{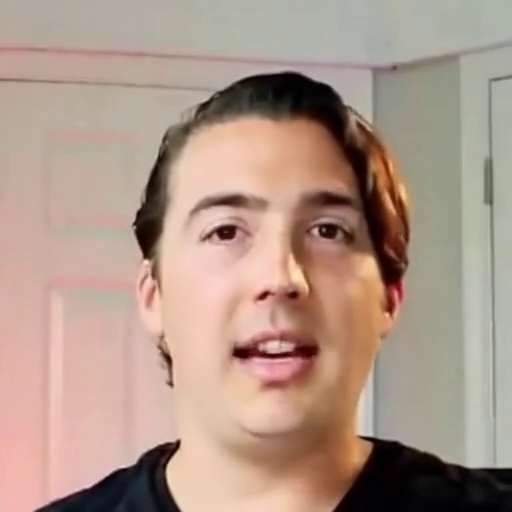}
    \end{subfigure}
    \hspace{-4pt}
      \begin{subfigure}{0.23\linewidth}
        \includegraphics[width=\linewidth]{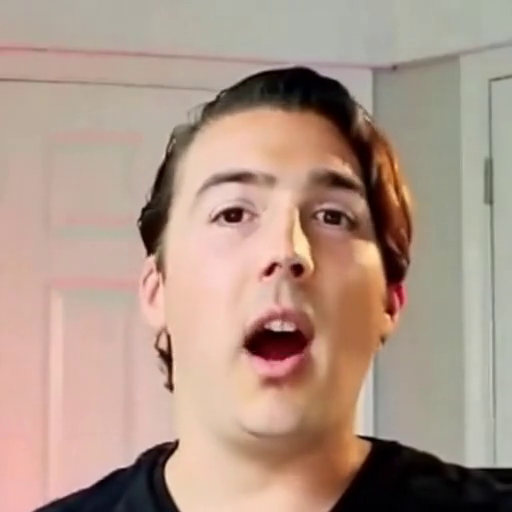}
    \end{subfigure}
    \end{minipage}
    
    \begin{minipage}{0.04\linewidth}
        \rotatebox{90}{Subject2}
    \end{minipage}
    \begin{minipage}{0.95\linewidth}
    \begin{subfigure}{0.23\linewidth}
        \includegraphics[width=\linewidth]{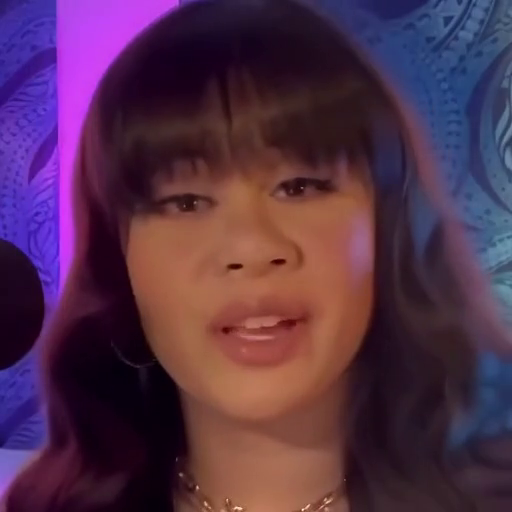}
    \end{subfigure}
    \hspace{-4pt}
     \begin{subfigure}{0.23\linewidth}
        \includegraphics[width=\linewidth]{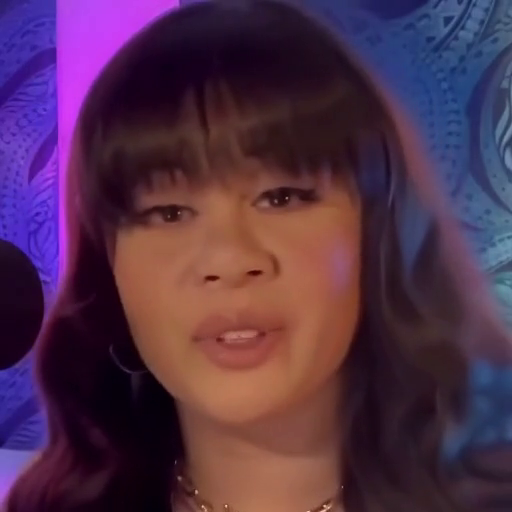}
    \end{subfigure}
    \hspace{-4pt}
      \begin{subfigure}{0.23\linewidth}
        \includegraphics[width=\linewidth]{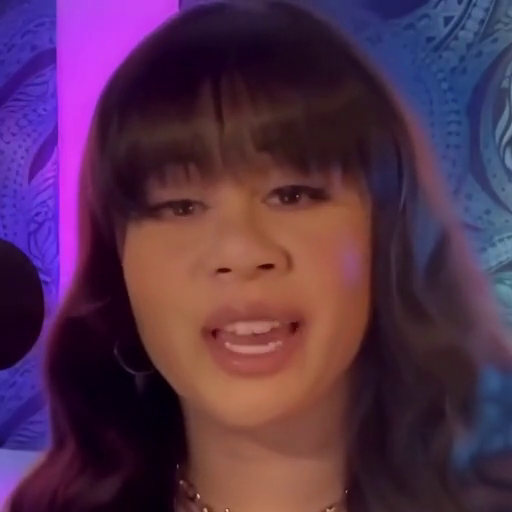}
    \end{subfigure}
    \hspace{-4pt}
      \begin{subfigure}{0.23\linewidth}
        \includegraphics[width=\linewidth]{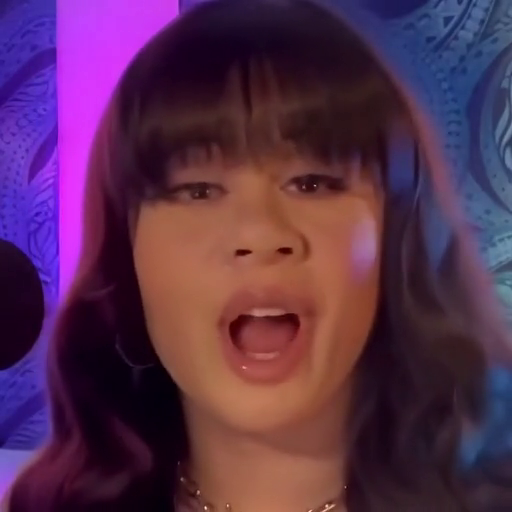}
    \end{subfigure}
    \end{minipage}
    \caption{Visualization of cross-subject singing videos generated by~\model, two subjects are included.}
    \label{fig:cross_subject}
    \end{minipage}
    \hspace{4pt}
    \begin{minipage}{0.49\textwidth}
    \centering
        \begin{minipage}{0.04\linewidth}
        \rotatebox{90}{Sketch}
    \end{minipage}
    \begin{minipage}{0.95\linewidth}
    \begin{subfigure}{0.23\linewidth}
        \includegraphics[width=\linewidth]{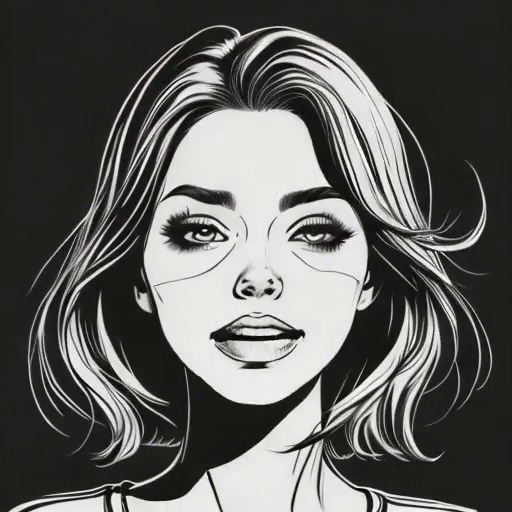}
    \end{subfigure}
    \hspace{-4pt}
     \begin{subfigure}{0.23\linewidth}
        \includegraphics[width=\linewidth]{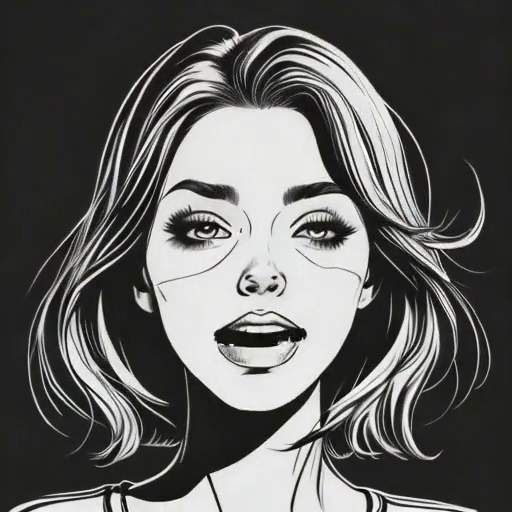}
    \end{subfigure}
    \hspace{-4pt}
      \begin{subfigure}{0.23\linewidth}
        \includegraphics[width=\linewidth]{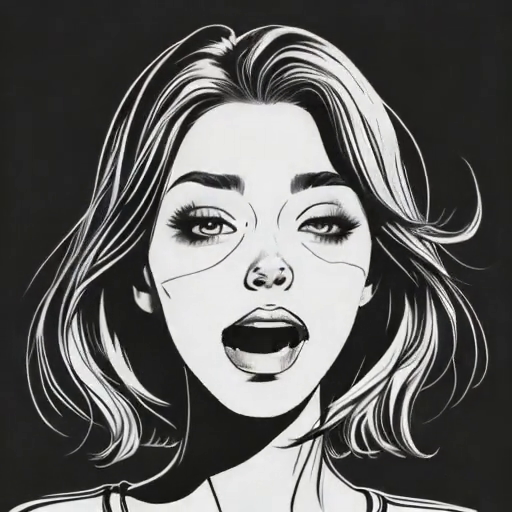}
    \end{subfigure}
    \hspace{-4pt}
      \begin{subfigure}{0.23\linewidth}
        \includegraphics[width=\linewidth]{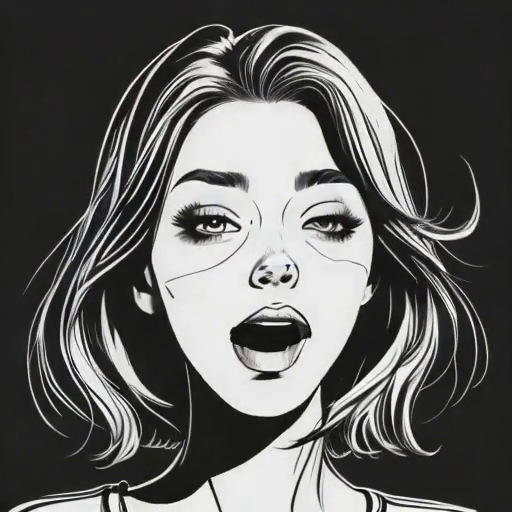}
    \end{subfigure}
    \end{minipage}

    \begin{minipage}{0.04\linewidth}
        \rotatebox{90}{Cartoon}
    \end{minipage}
    \begin{minipage}{0.95\linewidth}
    \begin{subfigure}{0.23\linewidth}
        \includegraphics[width=\linewidth]{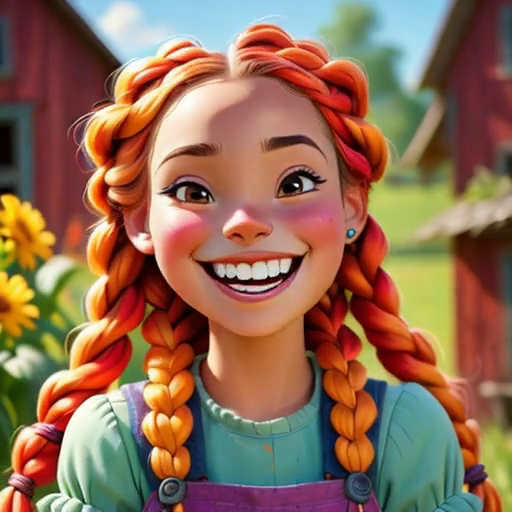}
    \end{subfigure}
    \hspace{-4pt}
     \begin{subfigure}{0.23\linewidth}
        \includegraphics[width=\linewidth]{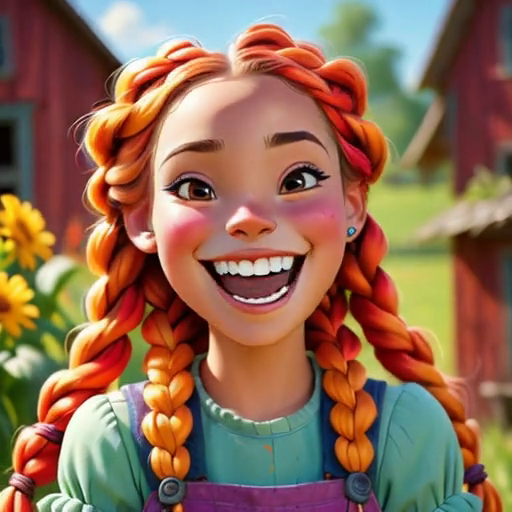}
    \end{subfigure}
    \hspace{-4pt}
      \begin{subfigure}{0.23\linewidth}
        \includegraphics[width=\linewidth]{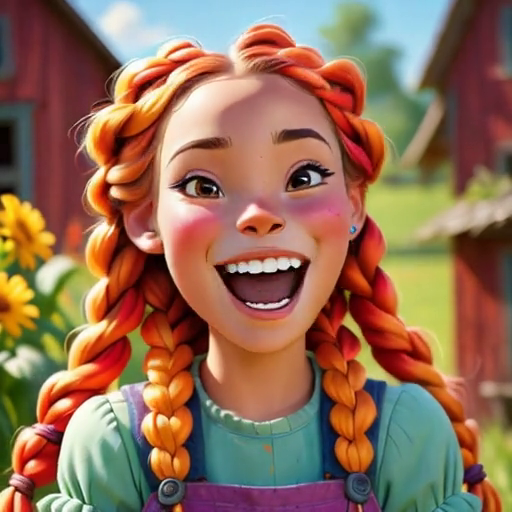}
    \end{subfigure}
    \hspace{-4pt}
      \begin{subfigure}{0.23\linewidth}
        \includegraphics[width=\linewidth]{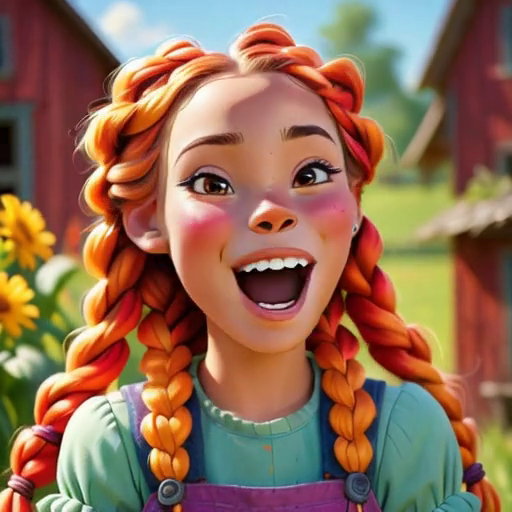}
    \end{subfigure}
    \end{minipage}
    
    \begin{minipage}{0.04\linewidth}
        \rotatebox{90}{Painting}
    \end{minipage}
    \begin{minipage}{0.95\linewidth}
    \begin{subfigure}{0.23\linewidth}
        \includegraphics[width=\linewidth]{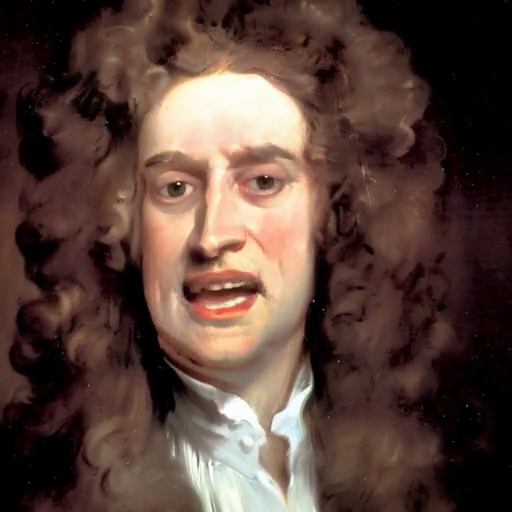}
    \end{subfigure}
    \hspace{-4pt}
     \begin{subfigure}{0.23\linewidth}
        \includegraphics[width=\linewidth]{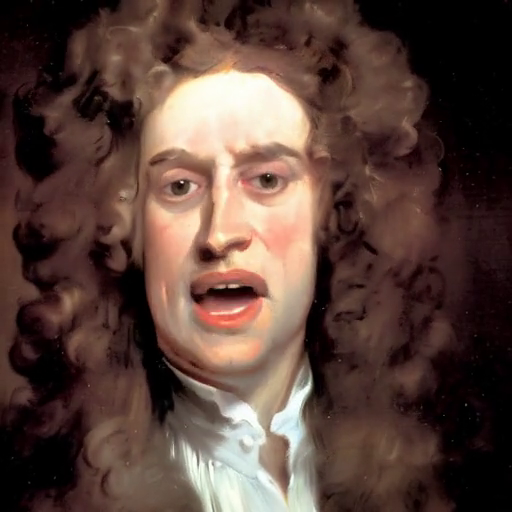}
    \end{subfigure}
    \hspace{-4pt}
      \begin{subfigure}{0.23\linewidth}
        \includegraphics[width=\linewidth]{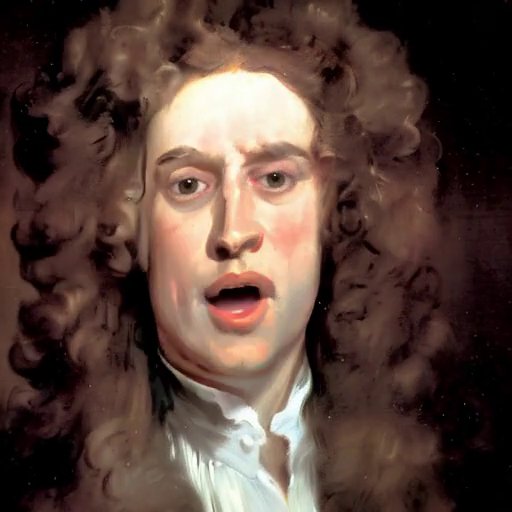}
    \end{subfigure}
    \hspace{-4pt}
      \begin{subfigure}{0.23\linewidth}
        \includegraphics[width=\linewidth]{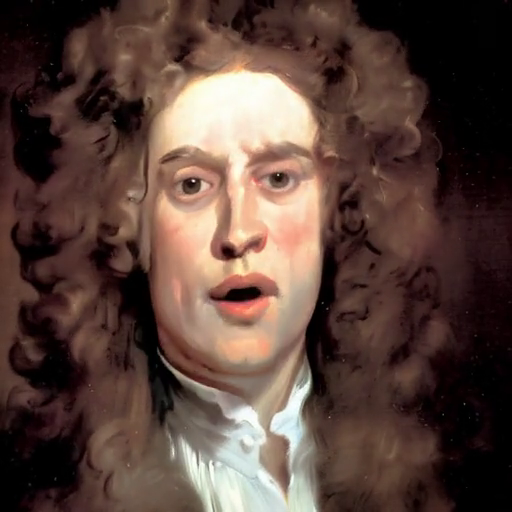}
    \end{subfigure}
    \end{minipage}
    \caption{Visualization of videos generated by \model~in different styles, including sketch, cartoon and painting. }
    \label{fig:multi-style}
    \end{minipage}
    \vspace{-8pt}
\end{figure*}
\subsection{Experiment Setting}
Note that we only give a brief introduction to the experimental setup here, detailed descriptions of experiment settings can refer to Appendix~\ref{app:detailed_exp}.

\subsubsection{Implementation}
We employ the pre-trained Reference UNet and Denoising UNet from~\cite{xu2024hallo}, keeping them frozen throughout the training process. Similarly, the VAE encoder, VAE decoder, and Face encoder are also frozen, allowing only the audio encoder, the proposed Multi-scale Spectral Module, and the Self-adaptive Filter Module to be updated during training.
The learning rate is set to \(10^{-5}\) using the Adam optimizer, and the training process consists of $17,000$ update steps.
All experiments are conducted on $4$ NVIDIA H800 GPUs. 
More details can refer to Appendix~\ref{app:imple_details}.

\subsubsection{Datasets}
In this paper, two datasets are used, one is the public in-the-lab dataset SingingHead dataset~\cite{wu2023singinghead} and another is our collected in-the-wild dataset SHV. More detailed introduction of the used datasets can refer to Appendix~\ref{app:dataset}. \textbf{SingingHead}: This dataset contains more than $27$ hours of synchronized singing video from $76$ subjects. 
\textbf{SHV}: Due to the lack of high-quality public in-the-wild singing video datasets, we collect wild singing head videos published on online platforms. We name the collected dataset as SHV, comprising $200$ videos, with a total duration of approximately $20$ hours\footnote{The collected dataset will be released once the paper is published}. We divide all videos into 2-second clips at a frame rate of $25$ fps and remove clips of poor quality. 
\subsubsection{Baseline Methods}
To demonstrate the singing video generation ability of our \model, we select several state-of-the-art baseline methods. For non-diffusion approaches, we choose Audio2Head~\cite{wang2021audio2head}, SadTalker~\cite{zhang2023sadtalker}, and MuseTalk~\cite{zhang2024musetalk}. Additionally, we include several diffusion methods: AniPortrait~\cite{wei2024aniportrait}, Echomimic~\cite{chen2024echomimic}, Hallo~\cite{xu2024hallo}, and Hallo2~\cite{cui2024hallo2}. More detailed description of the baseline methods can refer to Appendix~\ref{app:baseline}.

\subsubsection{Evaluation Metrics}
To evaluate the singing performance of different generation methods comprehensively, we choose multiple evaluation metrics. 
To evaluate the generated video quality, we employ FVD~\cite{unterthiner2019fvd}, CPBD~\cite{narvekar2011no}, PSNR and SSIM~\cite{wang2004image}. 
To evaluate lip synchronization and mouth shape, we utilize the Landmark Distance (LMD)~\cite{chen2019hierarchical}, focusing specifically on the area around the mouth~\cite{wang2022one}. Additionally, we assess the perceptual differences in mouth shape using metrics from~\cite{prajwal2020lip}, which include the distance score (LSE-D) and the confidence score (LSE-C).
For evaluating head motion and expression, we assess the diversity of the generated videos by calculating the standard deviation of the landmarks extracted from the frames. Additionally, we employ the Beat Align Score (BAS)~\cite{siyao2022bailando}, to evaluate the alignment between the audio and the generated head motions. 
\subsection{Quantitative Results}
We present the comparison results on SHV dataset in Table~\ref{tab:main_results} and SingingHead dataset in Table~\ref{tab:main_results_singing}, due to the limit of space, the full comparison results on SingingHead dataset are presented in Appendix~\ref{app:result_on_singing}. 
\begin{table}[h]
    \centering
      \caption{Comparison between \model~and seven state-of-the-art methods in generating singing videos on the SingingHead dataset~\cite{wu2023singinghead}.}
    \resizebox{\linewidth}{!}{
    \begin{tabular}{c|cccc}
    \toprule
      Method & Diversity  \((\uparrow) \) & BAS  \((\uparrow) \) & LMD  \((\downarrow) \)& FVD  \((\downarrow) \)\\
      \midrule
       Audio2Head  & \textbf{2.2523} & 0.2140 & 48.288 & 536.25  \\
       SadTalker &- & 0.2266 &  61.684 & 778.29  \\
       MuseTalk &0.7254 & 0.2513 &  42.121 &845.61  \\
       \midrule
       AniPortrait &1.3558 & 0.2511 & 61.931 & 415.73 \\
       Echomimic & 0.6733 & 0.0893 & 57.928 &  702.26\\
       Hallo & 1.2852 & 0.2540 & 42.832 & 539.62\\
       Hallo2 &  1.5500 & 0.2012  & 43.139 &  495.00\\
       \midrule
       \model~ & 1.9446 & \textbf{0.2630} &  \textbf{41.862} &  \textbf{365.41} \\
    \bottomrule
    \end{tabular}
    }
    \label{tab:main_results_singing}
    \vspace{-8pt}
\end{table}

The results indicate that our proposed \model~outperforms the baseline methods overall. Notably, our method generates more vivid head movements and demonstrates better synchronization with the singing audio. However, we acknowledge that when there are significant changes in head movement, some generated frames may appear slightly blurred. Additionally, it is important to note that the lip synchronization metrics (LSE-D and LSE-C) are primarily designed for evaluating talking video generation, which may lead to larger LSE-C scores for mouth movements associated with speaking rather than singing. This limitation suggests that while our model excels in many areas, further refinement of lip synchronization metrics tailored for singing videos could enhance evaluation accuracy.

To comprehensively evaluate the performance of our method and the baseline methods, we present the visualization results of the generative methods in Figure~\ref{fig:main_comparison}. The results show that our method generates more vivid singing videos while maintaining accurate lip synchronization. In contrast, the baseline methods struggle to produce singing videos, exhibiting less vivid lip synchronization and expressions. This highlights the effectiveness of our approach in capturing the nuances of singing performance, both in terms of visual quality and synchronization with the audio.

\subsection{Qualitative Results}
\begin{figure}[t]
    \centering
    \includegraphics[width=0.9\linewidth]{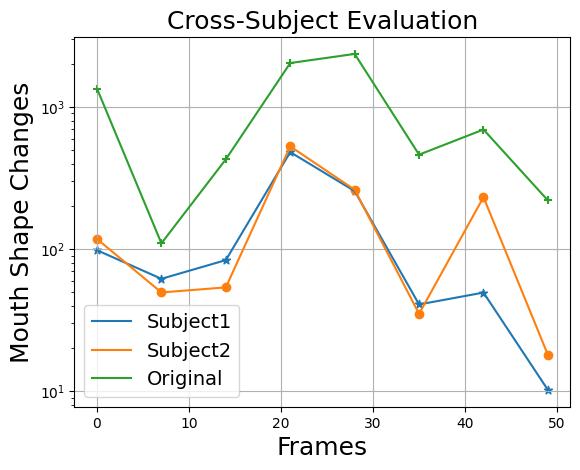}
    \vspace{-4pt}
    \caption{Visualization of mouth shape changes in generated videos across different subjects.}
    \label{fig:cross_sub_changes}
    \vspace{-8pt}
\end{figure}

\textbf{Cross-Subject Evaluation}.
We evaluate the cross-subject generation capability of our \model, which involves using the audio from one singing video while employing images of a different subject as the initial frame to generate new singing videos. The results of this cross-subject evaluation are shown in Figure~\ref{fig:cross_subject}. Our findings indicate that \model~is effective at retaining the patterns of the singing audio, successfully generating vivid singing videos featuring different subjects. We additionally provide a comparison of the changes in mouth shape for different samples generated by our \model~relative to the original reference images in Figure~\ref{fig:cross_sub_changes}, in which ``Original" indicates the reference video. This further demonstrates the flexibility and robustness of our approach in adapting to various visual inputs while accurately reflecting the audio characteristics.

\textbf{Multiple Style Evaluation}.
To assess the robustness of our \model~in handling various image styles, we select pictures in cartoon, painting, and sketch styles as reference images to evaluate their impact on singing performance. The results are presented in Figure~\ref{fig:multi-style}, indicating that \model~has strong robustness in generating singing videos across different artistic styles. Notably, the head movements and lip synchronization remain unaffected by the image art styles, highlighting the ability of our approach in adapting to diverse visual inputs. 

\subsection{Ablation Study}
To demonstrate the effectiveness of our proposed \model~in enhancing singing video generation through the Multi-scale Spectral Module (MSM) and the Self-adaptive Filter Module (SFM), we conduct an ablation study on these two components. The results of the ablation study on SHV dataset are presented in Table~\ref{tab:abla}. 
\begin{table}[t]
    \centering
    \small
      \caption{Ablation study results of the proposed two modules, conducted on the collected SHV dataset.. }
    \resizebox{\linewidth}{!}{
    \begin{tabular}{c|cccc}
    \toprule
     Method    &  Diversity (\(\uparrow\)) & BAS (\(\uparrow\)) & LMD (\(\downarrow\))& FVD (\(\downarrow\))\\
    \midrule
      \model$_{w/o~ MSM}$   &13.543 & 0.2289 & 54.217 & 881.19  \\
    \midrule
    \model$_{w/o~ SFM}$ & 16.087 & 0.2013 & 54.406 & 938.15\\
     \midrule
   \model$_{w/o~ both}$  & 17.511 & 0.2044 & 55.306 & 950.49\\
   \midrule
   \model & 14.445 & 0.2405 &53.373 & 503.78\\
    \bottomrule
    \end{tabular}
    }
    \label{tab:abla}
    \vspace{-12pt}
\end{table}
The findings indicate that removing the Multi-scale Spectral Module leads to less vivid video generation, as evidenced by decreased diversity and Beat Align Score (BAS). This is attributed to the model's inability to capture the complex spectral patterns of singing audio without the MSM. Furthermore, omitting the Self-adaptive Filter Module results in the retention of less important features, causing instability~(higher Diversity but lower BAS) in the generated outputs. These results underscore the importance of both modules in improving the quality and coherence of generated singing videos. 
\begin{figure}[t]
    \centering
     \begin{minipage}{0.04\linewidth}
        \rotatebox{90}{$w/o~MSM$}
    \end{minipage}
    \begin{minipage}{0.95\linewidth}
    \begin{subfigure}{0.23\linewidth}
        \includegraphics[width=\linewidth]{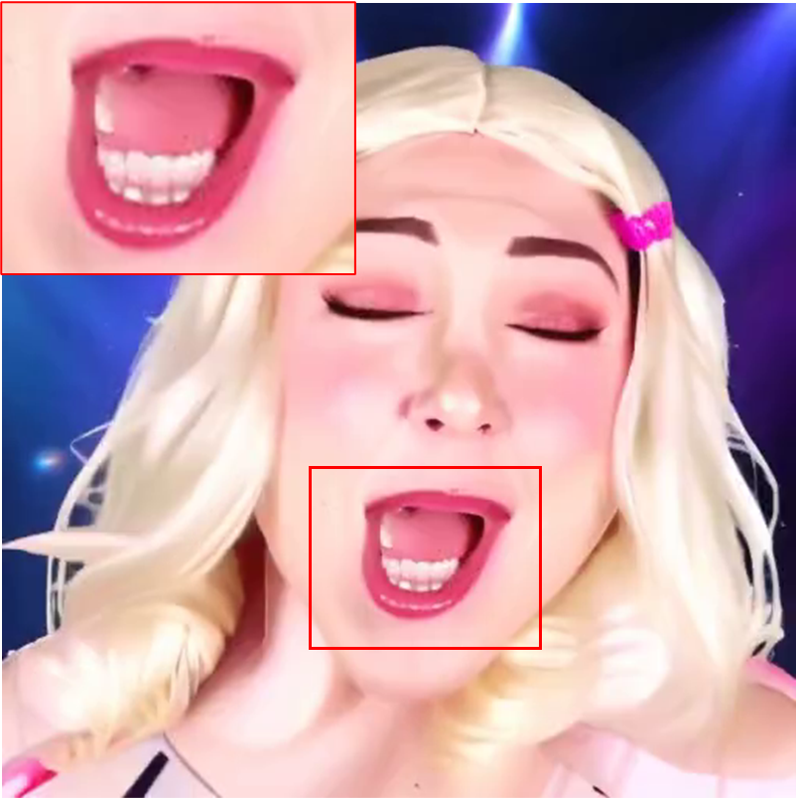}
    \end{subfigure}
    \begin{subfigure}{0.23\linewidth}
        \includegraphics[width=\linewidth]{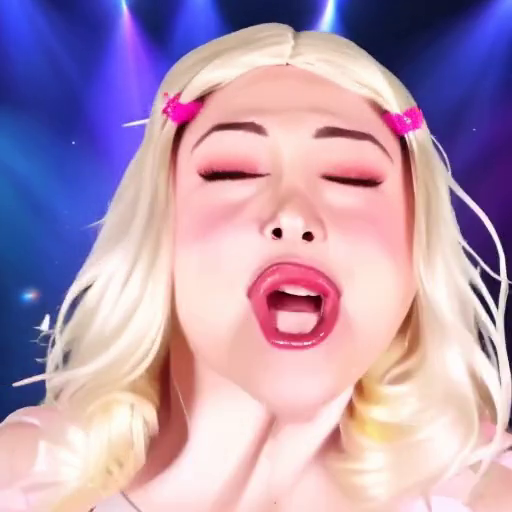}
    \end{subfigure}
    \begin{subfigure}{0.23\linewidth}
        \includegraphics[width=\linewidth]{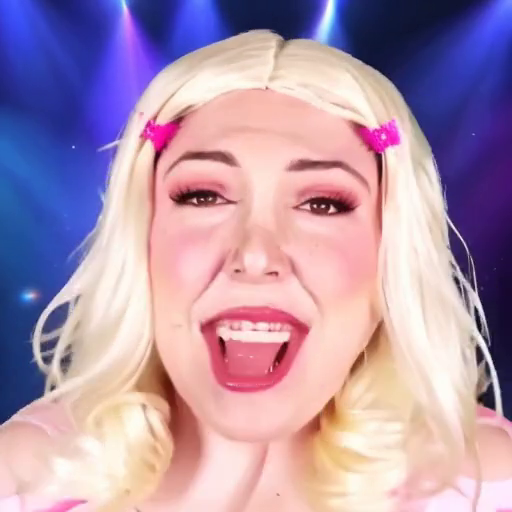}
    \end{subfigure}
    \begin{subfigure}{0.23\linewidth}
        \includegraphics[width=\linewidth]{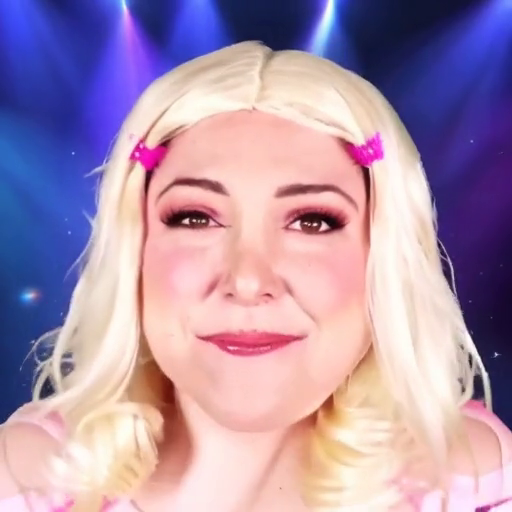}
    \end{subfigure}
    \end{minipage}

     \begin{minipage}{0.04\linewidth}
        \rotatebox{90}{$w/o~SFM$}
    \end{minipage}
    \begin{minipage}{0.95\linewidth}
    \begin{subfigure}{0.23\linewidth}
        \includegraphics[width=\linewidth]{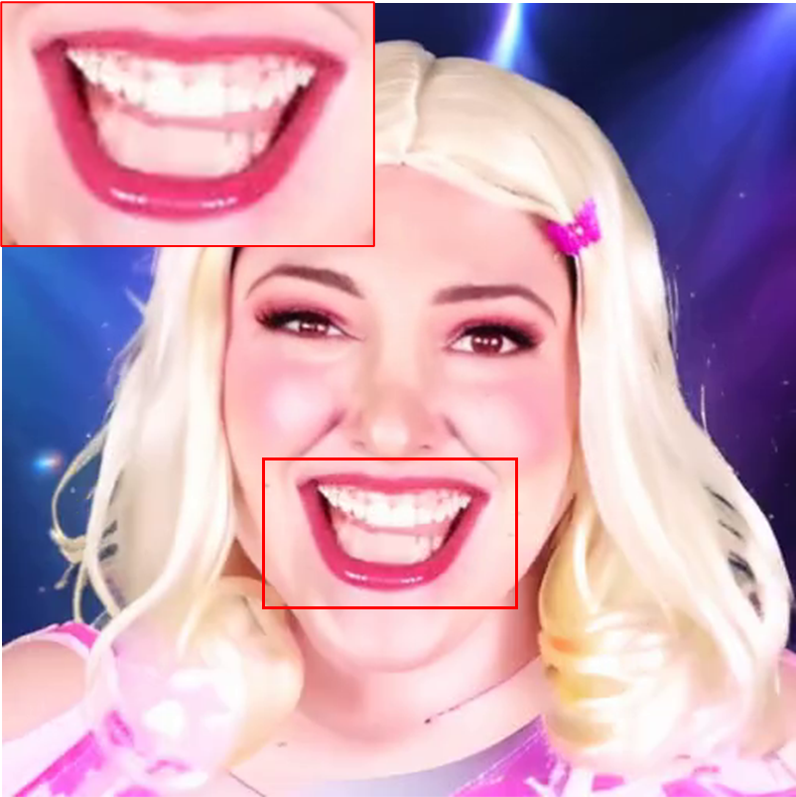}
    \end{subfigure}
    \begin{subfigure}{0.23\linewidth}
        \includegraphics[width=\linewidth]{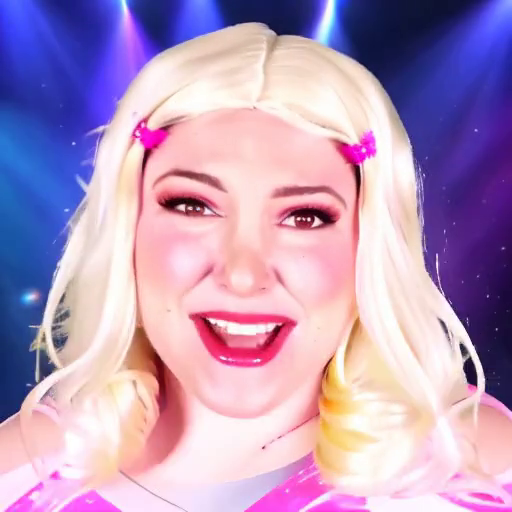}
    \end{subfigure}
    \begin{subfigure}{0.23\linewidth}
        \includegraphics[width=\linewidth]{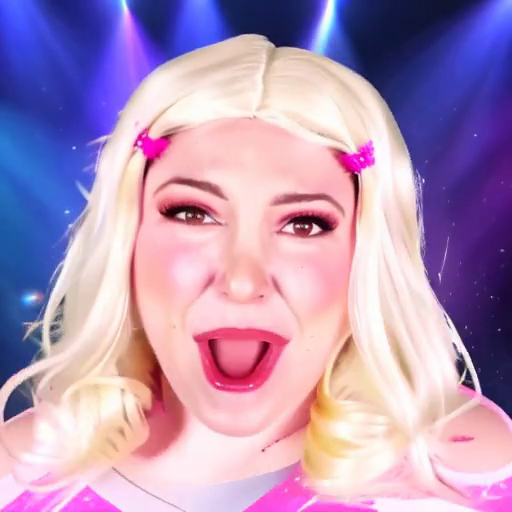}
    \end{subfigure}
    \begin{subfigure}{0.23\linewidth}
        \includegraphics[width=\linewidth]{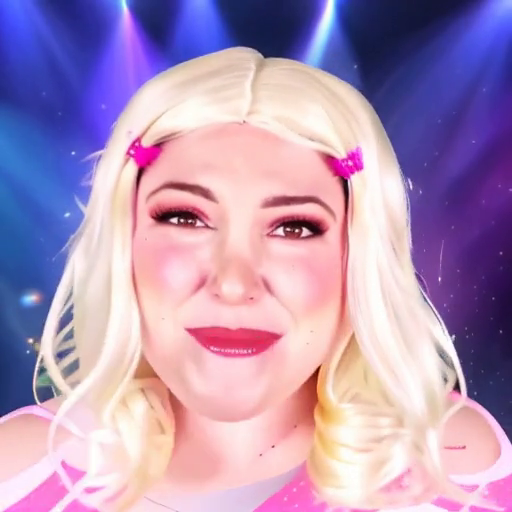}
    \end{subfigure}
    \end{minipage}

     \begin{minipage}{0.04\linewidth}
        \rotatebox{90}{$w/o~both$}
    \end{minipage}
    \begin{minipage}{0.95\linewidth}
    \begin{subfigure}{0.23\linewidth}
        \includegraphics[width=\linewidth]{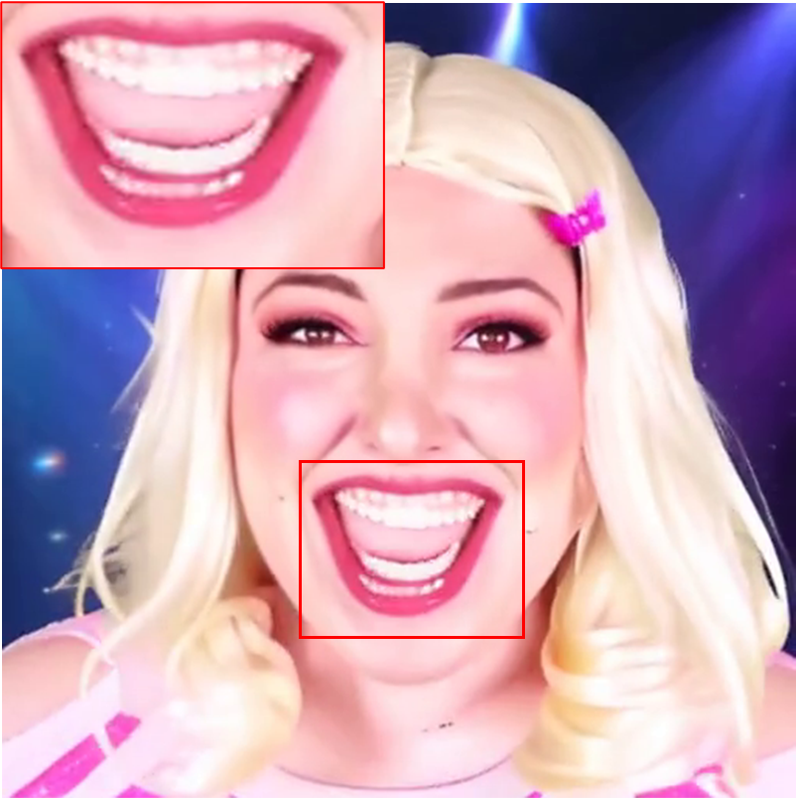}
    \end{subfigure}
    \begin{subfigure}{0.23\linewidth}
        \includegraphics[width=\linewidth]{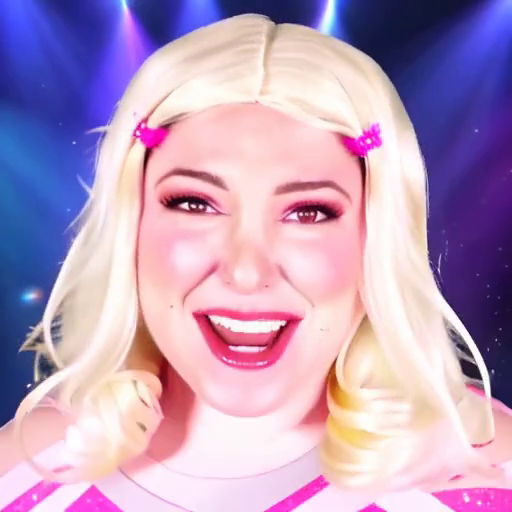}
    \end{subfigure}
    \begin{subfigure}{0.23\linewidth}
        \includegraphics[width=\linewidth]{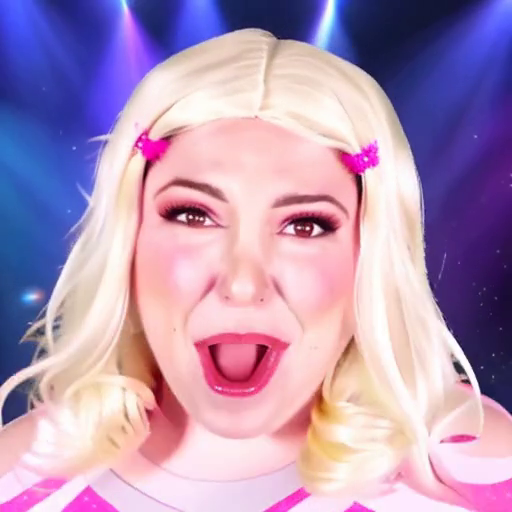}
    \end{subfigure}
    \begin{subfigure}{0.23\linewidth}
        \includegraphics[width=\linewidth]{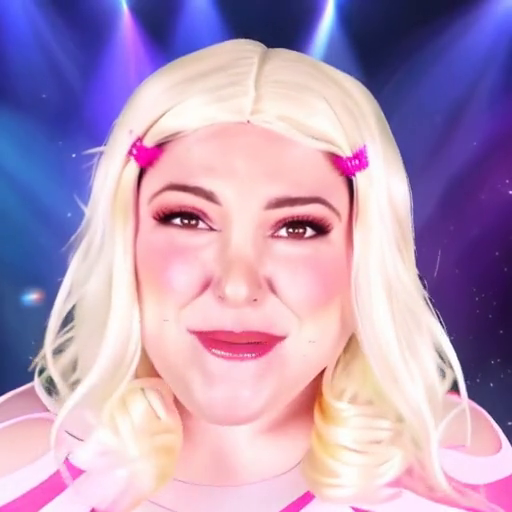}
    \end{subfigure}
    \end{minipage}
    
    \begin{minipage}{0.04\linewidth}
        \rotatebox{90}{\model}
    \end{minipage}
    \begin{minipage}{0.95\linewidth}
    \begin{subfigure}{0.23\linewidth}
        \includegraphics[width=\linewidth]{figures/our/2bCItaHq3JQ-008-2_0007.png}
    \end{subfigure}
    \hspace{-4pt}
    \begin{subfigure}{0.23\linewidth}
        \includegraphics[width=\linewidth]{figures/our/2bCItaHq3JQ-008-2_0010.png}
    \end{subfigure}
    \hspace{-4pt}
    \begin{subfigure}{0.23\linewidth}
        \includegraphics[width=\linewidth]{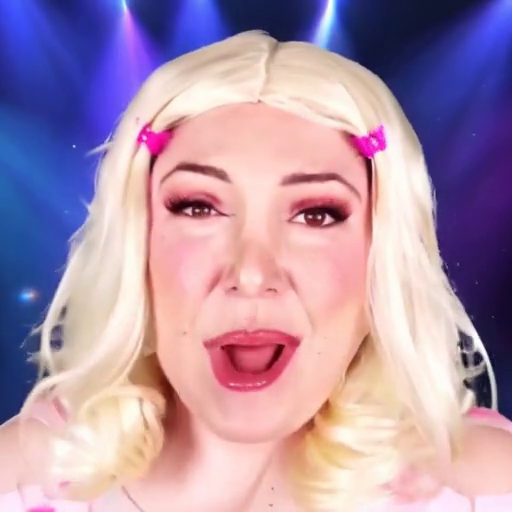}
    \end{subfigure}
    \hspace{-4pt}
    \begin{subfigure}{0.23\linewidth}
        \includegraphics[width=\linewidth]{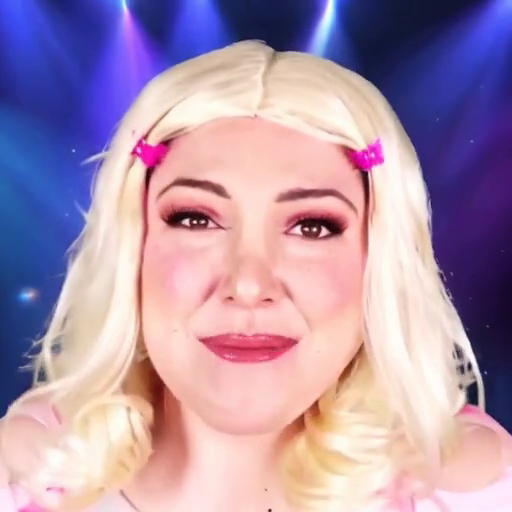}
    \end{subfigure}
    \end{minipage}
    \caption{The visualization samples when removing different modules of \model.}
    \label{fig:abla}
    \vspace{-16pt}
\end{figure}

To further evaluate the effectiveness of the proposed modules, we present samples of the generated videos without these components in Figure~\ref{fig:abla}. It shows that the absence of MSM and SFM results in inaccurate lip motions and less vivid video generation, which aligns with our previous conclusions, highlighting the critical role that both modules.

\section{Conclusion}
\label{sec:conclusion}
In conclusion, our proposed \model~addresses the challenges inherent in singing video generation by leveraging two innovative modules: the Multi-scale Spectral Module and the Self-adaptive Filter Module. Through extensive experiments and ablation studies, we have demonstrated that these modules significantly enhance the model's ability to capture the complex patterns of singing audio and emphasize relevant features for vivid expression. Our approach outperforms existing methods in generating realistic and synchronized singing videos, showcasing the effectiveness of audio-driven animation in capturing the nuances of singing behavior. Furthermore, the collection of our diverse SHV dataset addresses the current scarcity of high-quality singing video datasets, providing a valuable resource for future research. Overall, this work not only advances the state of singing head generation but also sets a foundation for further exploration of audio-driven animation techniques in various artistic contexts.

{
    \small
    \bibliographystyle{ieeenat_fullname}
    \bibliography{main}
}
\clearpage
\setcounter{page}{1}
\maketitlesupplementary

\section{Experiment}
Our source code and the generated video samples are available at: 
https://anonymous.4open.science/r/singer-E2F2
\subsection{Detailed Settings} ~\label{app:detailed_exp}
\subsubsection{Implementation Details} \label{app:imple_details}
We employ the pre-trained Reference UNet and Denoising UNet from~\cite{xu2024hallo}, keeping them frozen throughout the training process. Similarly, the VAE encoder, VAE decoder, and Face encoder are also frozen, allowing only the audio encoder, the proposed Multi-scale Spectral Module, and the Self-adaptive Filter Module to be updated during training.
For each training iteration, 16 video frames are randomly selected as input for the Denoising UNet, with the first frame serving as the reference image. These frames are resized to \(512 \times 512\) pixels. The learning rate is set to \(1 \times 10^{-5}\) using the Adam optimizer, and the training process consists of 17000 steps.
All experiments are conducted on 4 NVIDIA H800 GPUs.

\subsubsection{SHV Datasets} \label{app:dataset}
Due to the lack of high-quality public singing video datasets, we collected wild singing head videos published on online platforms. We name the dataset as SHV, it comprises 200 videos, with a total duration of approximately 20 hours\footnote{The collected dataset will be released once the paper is published}. We preprocess the video data by dividing all videos into 2-second clips at a frame rate of 25 fps. For each clip, we crop the face region with a scale ratio of 0.8, and we also remove clips that are of poor quality. These preprocessing steps ensure that the dataset is well-suited for training and evaluating our model, enhancing the generation of realistic singing videos.
\begin{figure}[h]
    \centering
    \begin{subfigure}{0.23\linewidth}
        \includegraphics[width=\linewidth]{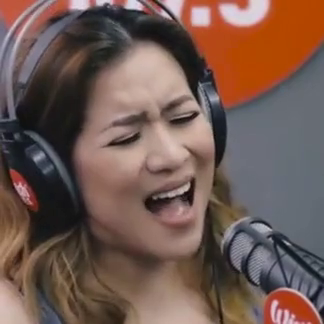}
    \end{subfigure}
        \begin{subfigure}{0.23\linewidth}
        \includegraphics[width=\linewidth]{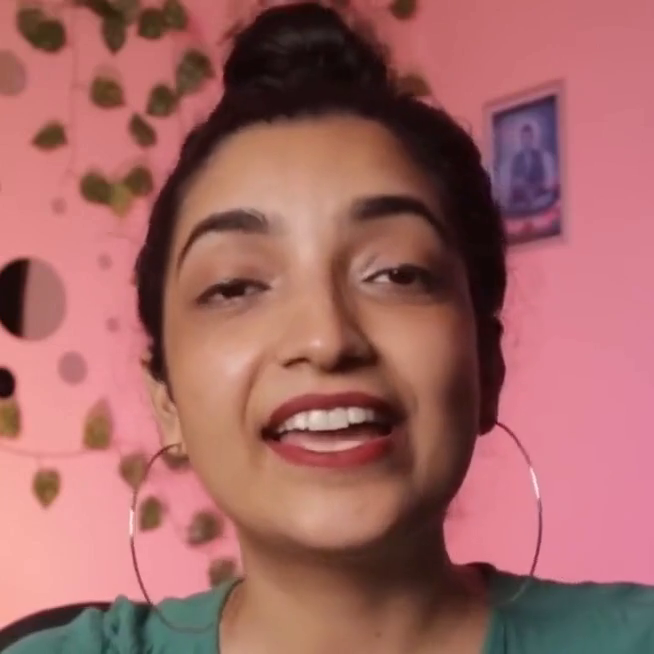}
    \end{subfigure}
        \begin{subfigure}{0.23\linewidth}
        \includegraphics[width=\linewidth]{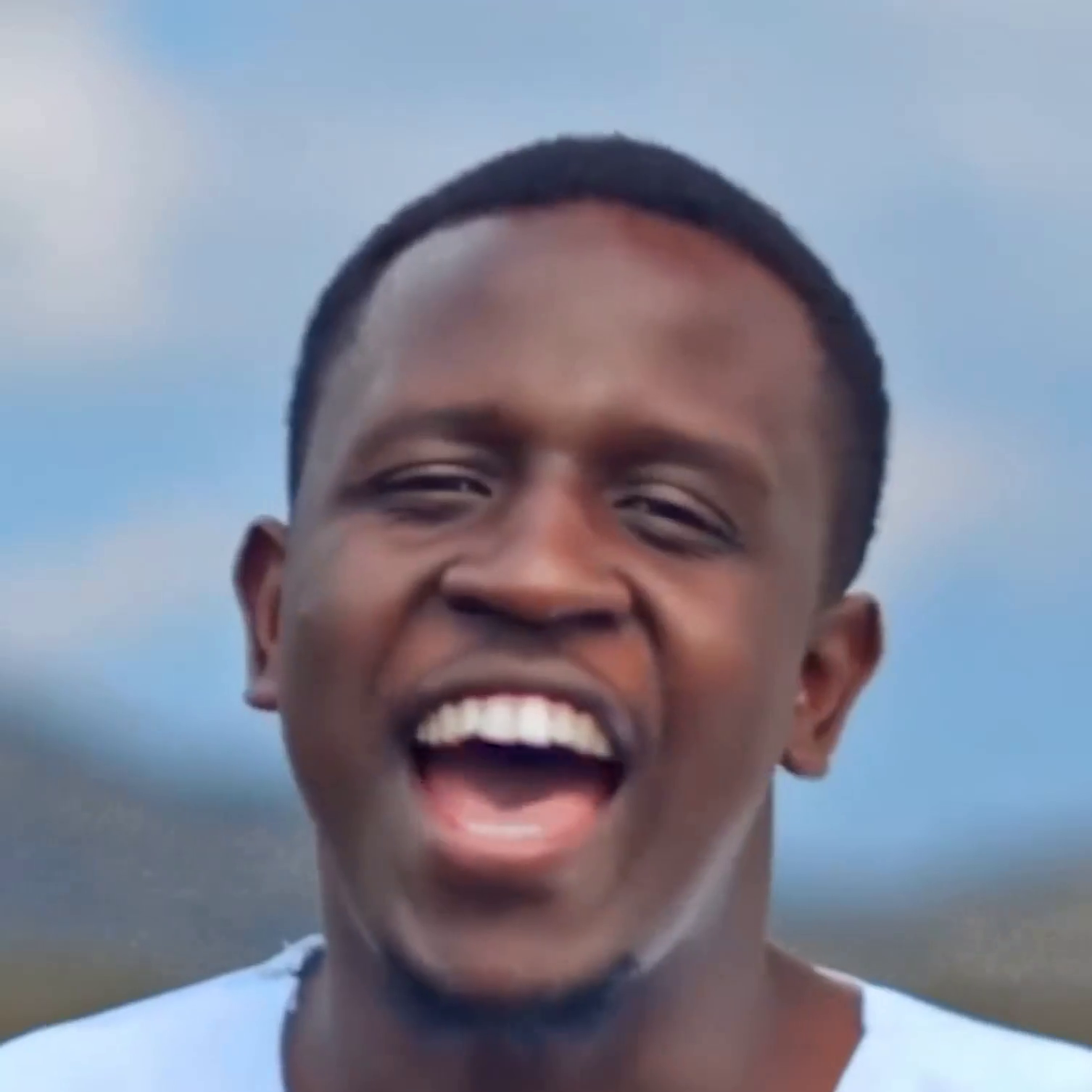}
    \end{subfigure}
        \begin{subfigure}{0.23\linewidth}
        \includegraphics[width=\linewidth]{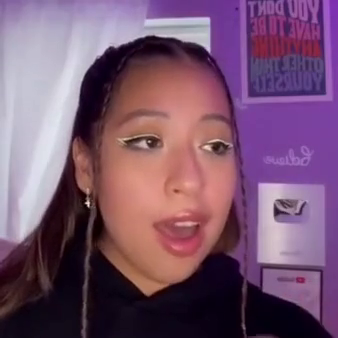}
    \end{subfigure}
    \caption{The samples of the preprocessed videos in SHV dataset. }
    \label{fig:samples_rep}
\end{figure}
We present some examples of the preprocessed videos in Figure~\ref{fig:samples_rep}. For our experiments, we selected approximately 700 processed clips and divided them into a training set and a test set in a 4:1 ratio. 

\begin{table}[h]
    \centering
    \resizebox{\linewidth}{!}{
    \begin{tabular}{c|ccccc}
    \toprule
        Dataset & Subj. & Dura. & BGM & Background &  2D \\
        \midrule
        RAVDESS~\cite{livingstone2018ryerson} & 23 & 2.6h & \ding{56} & \ding{56} & \ding{52}\\
        Song2Face~\cite{iwase2020song2face} & 7 & 2h & \ding{52} & 
        \ding{56} &\ding{56}\\
        Musicface~\cite{liu2024musicface} & 6 & 40h & \ding{52} & \ding{56} & \ding{56}  \\
        SingingHead~\cite{wu2023singinghead} & 76 & 27h & \ding{52}  & \ding{56}  & \ding{52} \\
    \midrule
        SHV (Our) & 200 & 20 & \ding{52} & \ding{52}  & \ding{52}  \\
    \bottomrule
    \end{tabular}
    }
\caption{The comparison of singing datasets. “Subj.”, “BGM”,
“2D”, Background represent Subjects, Background music, 2D
videos, Background image, respectively.}
 \label{tab:datasets_comp}
\end{table}

\begin{table*}[h]
    \centering
      \caption{Comparison between \model~and seven state-of-the-art methods in generating singing videos on the SingingHead dataset~\cite{wu2023singinghead}.}
    \resizebox{\linewidth}{!}{
    \begin{tabular}{c|cccc|ccc|cc}
    \toprule
      \multirow{2}{*}{Method}   & \multicolumn{4}{c|}{\textbf{Video Quality}} & \multicolumn{3}{c|}{\textbf{Lip Synchronization}} & \multicolumn{2}{c}{\textbf{Motion}}\\
      ~ & SSIM \((\uparrow) \)& PSNR  \((\uparrow) \) & CPBD  \((\uparrow) \) & FVD  \((\downarrow) \)& LMD  \((\downarrow) \)& LSE-D  \((\downarrow) \)& LSE-C  \((\uparrow) \) & Diversity  \((\uparrow) \) & BAS  \((\uparrow) \)\\
      \midrule
       Audio2Head  &  0.5967 & 30.651 & 0.4153 & 536.25 & 48.288 & 7.1820 & 1.2238 & \textbf{2.2523} & 0.2140 \\
       SadTalker & 0.5183 & 30.053 & 0.4627 & 778.29 & 61.684 & 8.2740 & 0.3301 & - & 0.2266 \\
       MuseTalk & 0.5613 & 27.972 & 0.4997 & 845.61 & 42.121 & 7.6332 & 1.9953 & 0.7254 & 0.2513 \\
       \midrule
       AniPortrait & 0.6306 & 31.312 & 0.5175 & 415.73 & 61.931 & 7.5120 & 0.3677 & 1.3558 & 0.2511\\
       Echomimic & 0.4564 & 28.289 & 0.4639 & 702.26 & 57.928 & 8.1475 & 0.2990 & 0.6733 & 0.0893 \\
       Hallo & 0.5658 & 29.242 & 0.4839 & 539.62 & 42.832 & 7.3554 & \textbf{2.0927} & 1.2852 & 0.2540 \\
       Hallo2 &  0.6490 & 31.163 & 0.4704 & 495.00 & 43.139 & \textbf{7.1534} & 1.9801 & 1.5500 & 0.2012 \\
       \midrule
       \model~ & \textbf{0.6608} & \textbf{31.701} & \textbf{0.5210} & \textbf{365.41} & \textbf{41.862} & 7.8552 & 0.4096 & 1.9446 & \textbf{0.2630} \\
       \midrule
       GT &  - & - & 0.5532 & 0.0000 & 0.0000 & 6.5759 & 1.4830 & 11.746 & 0.2738\\
    \bottomrule
    \end{tabular}
    }
    \label{tab:main_results_singing_ref}
\end{table*}
\subsubsection{Baseline Methods} \label{app:baseline}
To demonstrate the singing video generation ability of our \model, we selected several state-of-the-art baseline methods. For non-diffusion approaches, we chose Audio2Head~\cite{wang2021audio2head}, SadTalker~\cite{zhang2023sadtalker}, and MuseTalk~\cite{zhang2024musetalk}:
\begin{itemize}
    \item Audio2Head~\cite{wang2021audio2head}: Audio2Head generates realistic talking-head videos from a single reference image by predicting natural head motions that align with speech and maintaining appearance consistency during large movements. It uses an RNN-based head pose predictor and a keypoint-based dense motion field to capture full image motions from audio input. Finally, an image generation network renders the video based on these motion fields and the reference image.
    \item SadTalker~\cite{zhang2023sadtalker}: SadTalker generates realistic talking heads by mapping audio input to 3D motion coefficients for head pose and expression. It includes ExpNet for learning expressions and PoseVAE for diverse head motions, combining these to drive a 3D-aware face render that synthesizes natural, coherent video.
    \item MuseTalk~\cite{zhang2024musetalk}: MuseTalk is a high-fidelity talking face video generator that creates lip-sync targets in a Variational Autoencoder’s latent space for efficient inference. MuseTalk projects the lower half of the face and the full face image into a low-dimensional latent space, using a multi-scale U-Net to combine audio and visual features. 
\end{itemize}

Additionally, we included several diffusion methods: AniPortrait~\cite{wei2024aniportrait}, Echomimic~\cite{chen2024echomimic}, Hallo~\cite{xu2024hallo}, and Hallo2~\cite{cui2024hallo2}:
\begin{itemize}
    \item AniPortrait~\cite{wei2024aniportrait}:AniPortrait is a framework for creating high-quality animation from audio and a reference portrait image. It operates in two stages: first, extracting 3D representations from audio and projecting them into 2D facial landmarks; then, using a diffusion model with a motion module to transform the landmarks into photorealistic, temporally consistent animation.
    \item Echomimic~\cite{chen2024echomimic}: EchoMimic is trained on both audio and facial landmarks, enabling it to generate portrait videos from audio alone, facial landmarks alone, or a combination of both. This versatility is achieved through a novel training strategy.
    \item Hallo~\cite{xu2024hallo}: Hallo replaces traditional parametric models with an end-to-end diffusion-based framework, introducing a hierarchical audio-driven visual synthesis module for precise alignment of lip, expression, and pose motion with audio. Integrating diffusion models, a UNet denoiser, temporal alignment, and a reference network, it enables adaptive control over expressions and poses for personalized outputs.
    \item Hallo2~\cite{cui2024hallo2}: Hallo2 uses vector quantization of latent codes and temporal alignment to ensure coherence over time. It uses a high-quality decoder enables 4K visual synthesis, while adjustable semantic text labels for portrait expressions provide enhanced control and diversify generated content beyond traditional audio cues.
\end{itemize}

Due to the lack of in-the-wild public singing datasets, the evaluation is performed on the test set of our collected dataset SHV and the public in-the-lab dataset SingingHead~\cite{wu2023singinghead}, ensuring a relevant and fair comparison of singing performance across all selected methods.

\subsubsection{Evaluation Metrics}

To evaluate the singing performance of different generation methods comprehensively, we choose multiple evaluation metrics. 
To evaluate the generated video quality, we employ FVD~\cite{unterthiner2019fvd}, CPBD~\cite{narvekar2011no}, PSNR and SSIM~\cite{wang2004image}, FVD evaluates the generative methods from both spatial and temporal aspects, CPBD assesses the sharpness of generated frames, SSIM measures the differences between the properties of the pixels and PSNR checks the absolute error between the pixels. 
To evaluate lip synchronization and mouth shape, we utilize the Landmark Distance (LMD)~\cite{chen2019hierarchical}, focusing specifically on the area around the mouth~\cite{wang2022one}. Additionally, we assess the perceptual differences in mouth shape using metrics from~\cite{prajwal2020lip}, which include the distance score (LSE-D) and the confidence score (LSE-C).
For evaluating head motion, we assess the diversity of the generated movements by calculating the standard deviation of the landmarks extracted from the frames. Additionally, we employ the Beat Align Score (BAS)~\cite{siyao2022bailando}, to evaluate the alignment between the audio and the generated head motions. 

\subsection{Results on SingingHead Dataset}~\label{app:result_on_singing}

\subsubsection{Main Results} 
The full experiment results on the SingingHead dataset are presented in Table~\ref{tab:main_results_singing_ref}. It can be seen that our method achieves better performance in video quality and motion diversity compared to the baseline methods. Note that the SingingHead dataset mainly consists of Chinese songs, and the evaluation metrics LSE-C, LSE-D, and BAS are based on pretrained models that were trained on English songs. Therefore, the evaluation results on these metrics may not be fully reliable for the SingingHead dataset.

\subsubsection{Visualization} 
\begin{figure*}[h]
    \begin{minipage}{0.02\linewidth}
    \centering
        \rotatebox{90}{GT}
    \end{minipage}
    \begin{minipage}{0.97\linewidth}
    \begin{subfigure}{0.12\linewidth}
        \includegraphics[width=\linewidth]{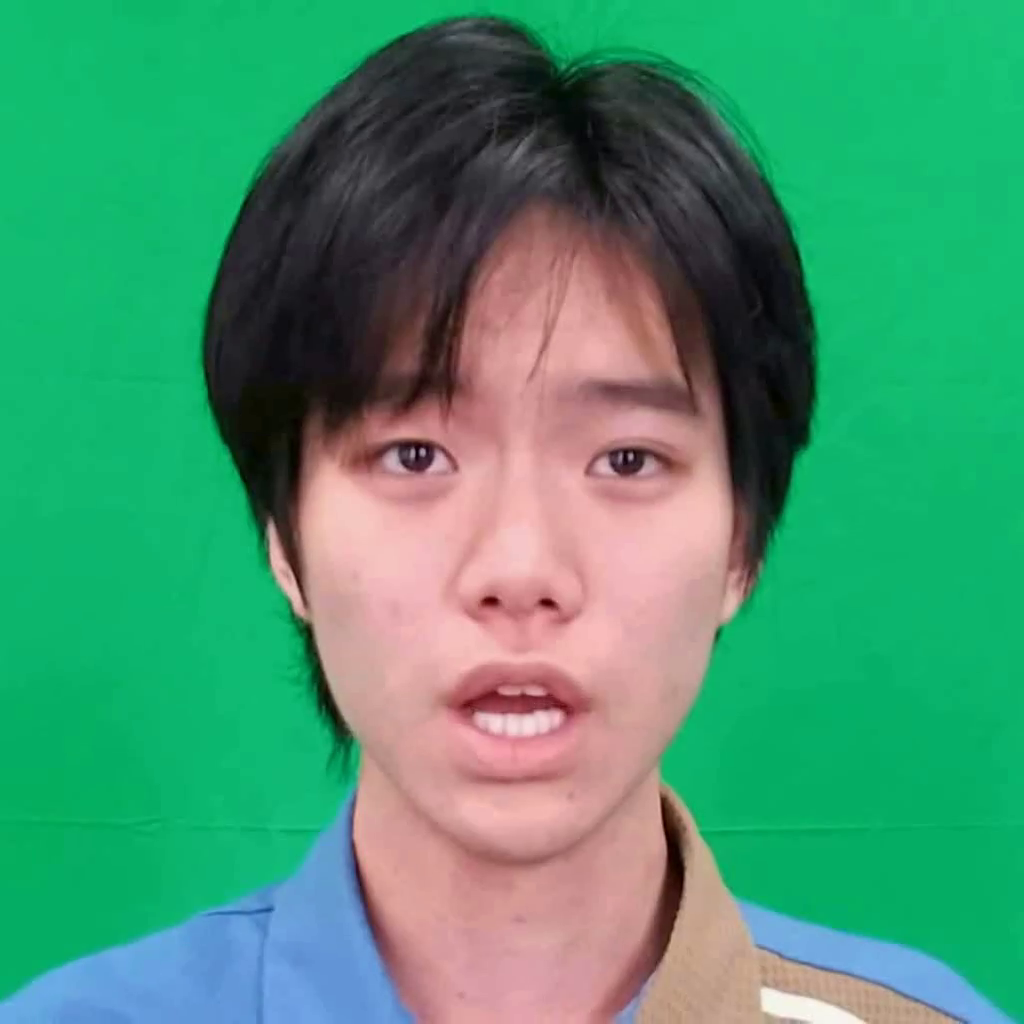}
    \end{subfigure}
    \hspace{-4pt}
        \begin{subfigure}{0.12\linewidth}
        \includegraphics[width=\linewidth]{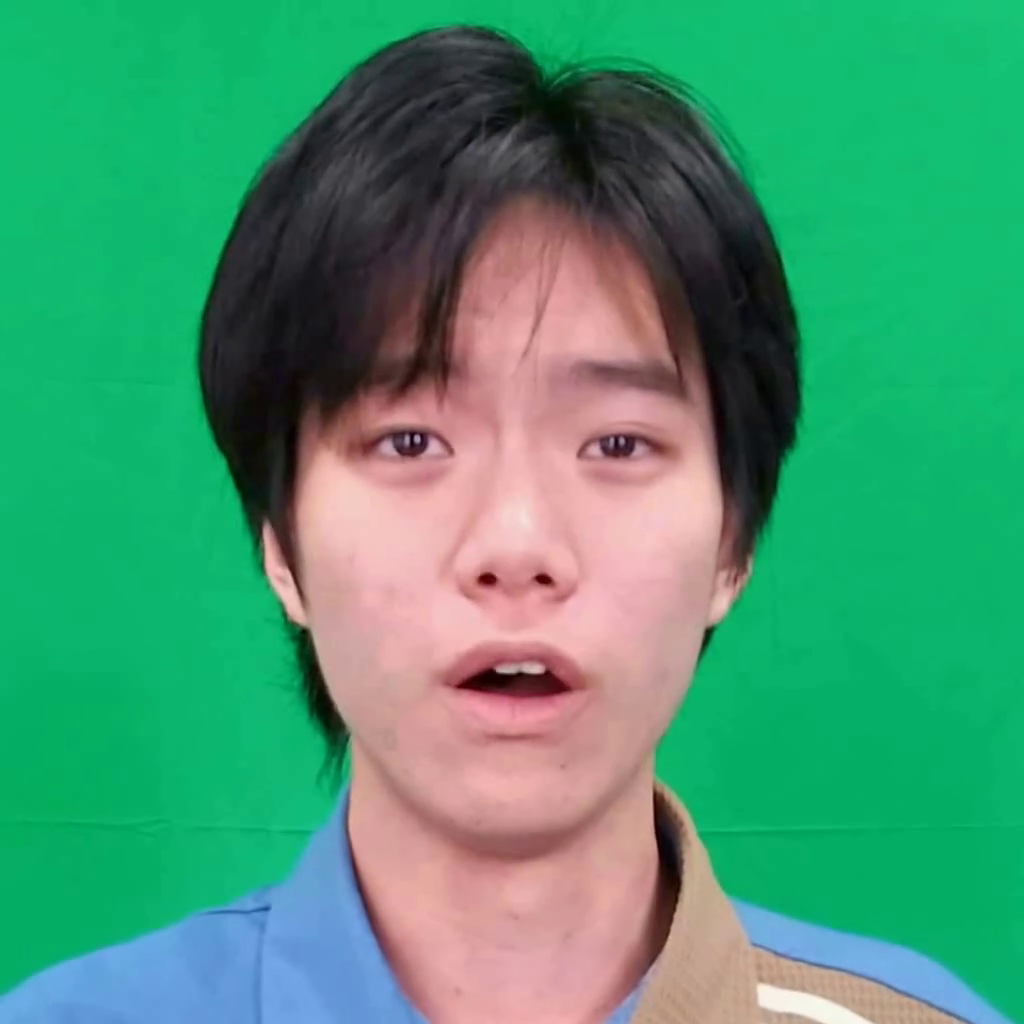}
    \end{subfigure}
     \hspace{-4pt}
        \begin{subfigure}{0.12\linewidth}
        \includegraphics[width=\linewidth]{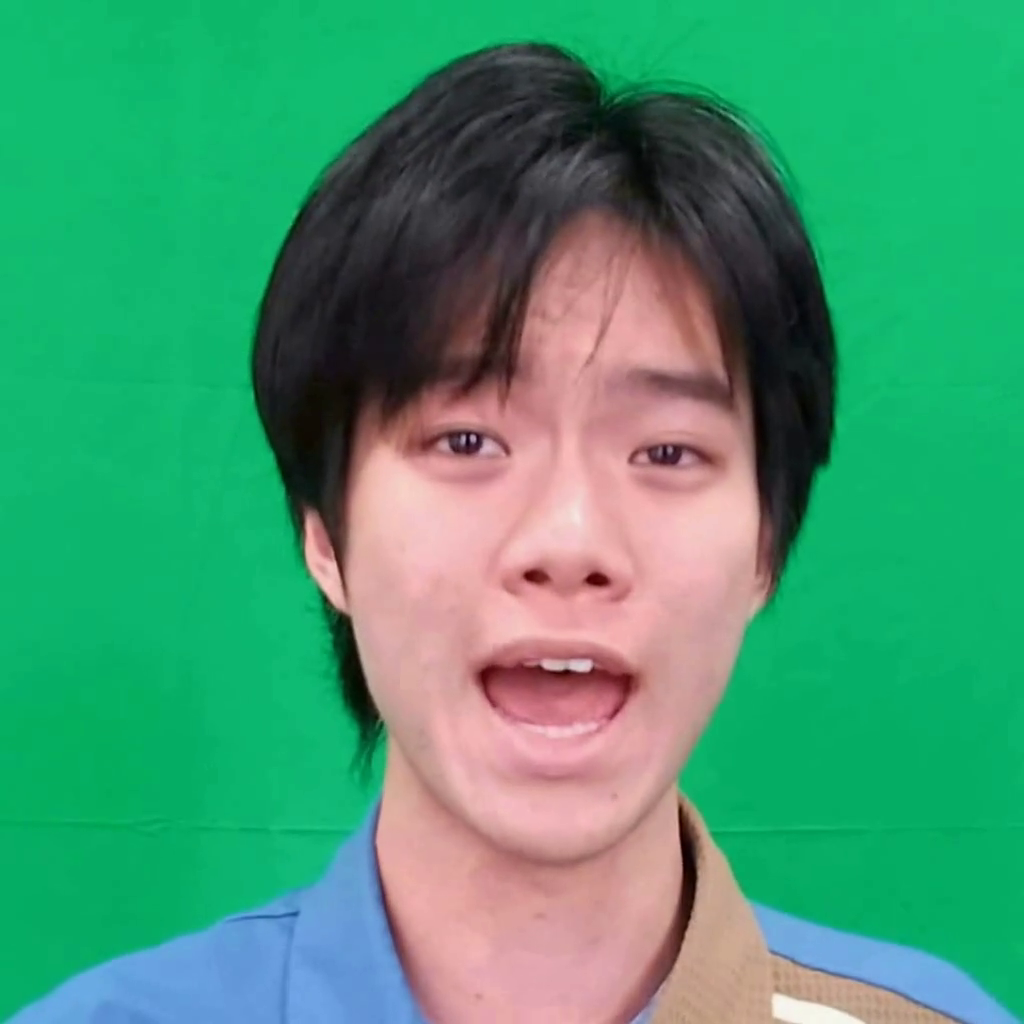}
    \end{subfigure}
     \hspace{-4pt}
        \begin{subfigure}{0.12\linewidth}
        \includegraphics[width=\linewidth]{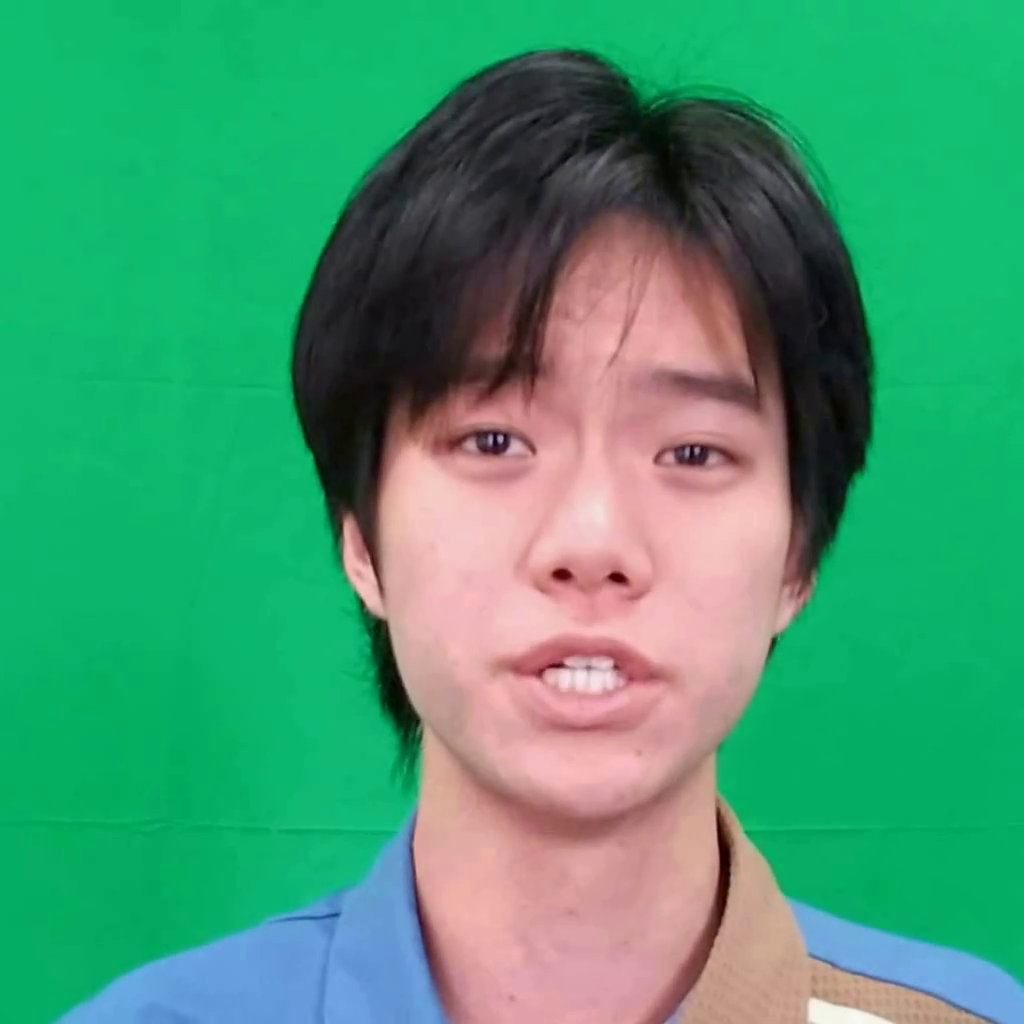}
    \end{subfigure}
     \hspace{-4pt}
        \begin{subfigure}{0.12\linewidth}
        \includegraphics[width=\linewidth]{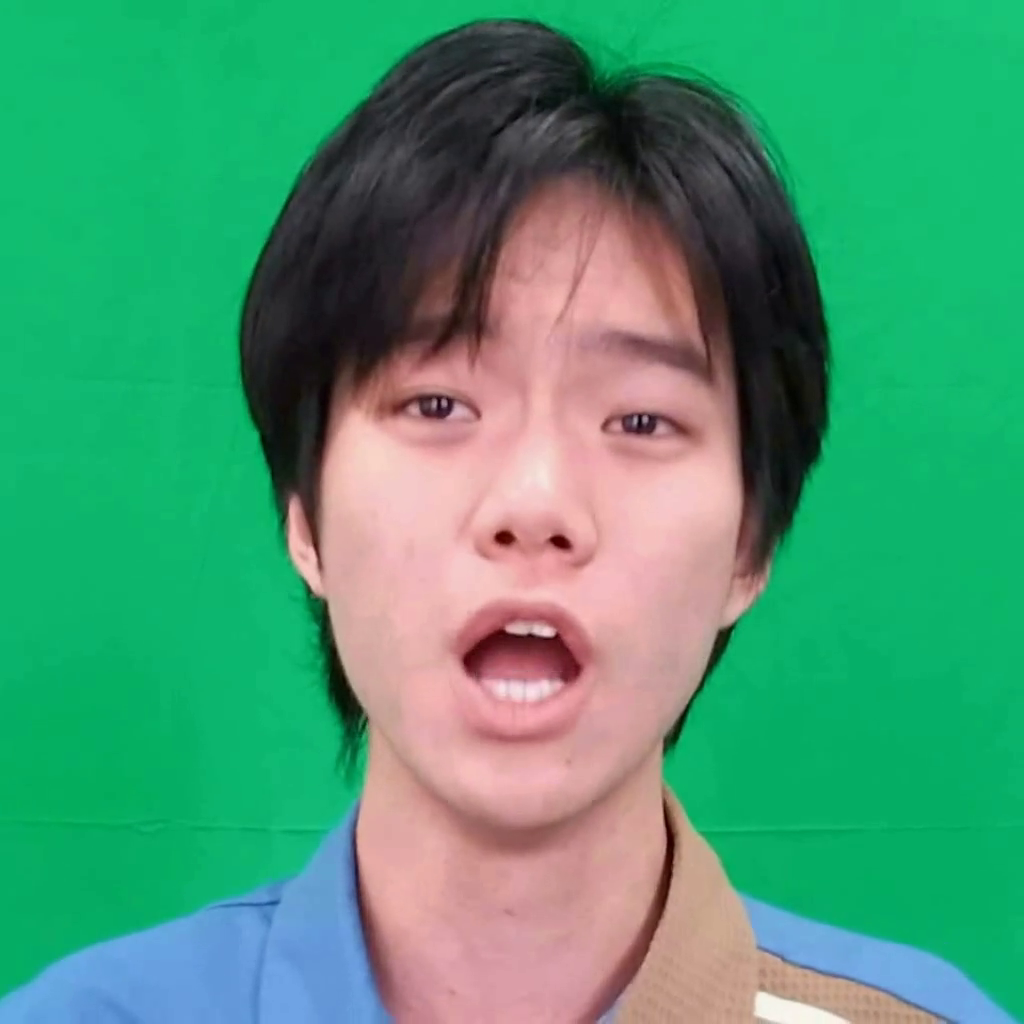}
    \end{subfigure}
     \hspace{-4pt}
        \begin{subfigure}{0.12\linewidth}
        \includegraphics[width=\linewidth]{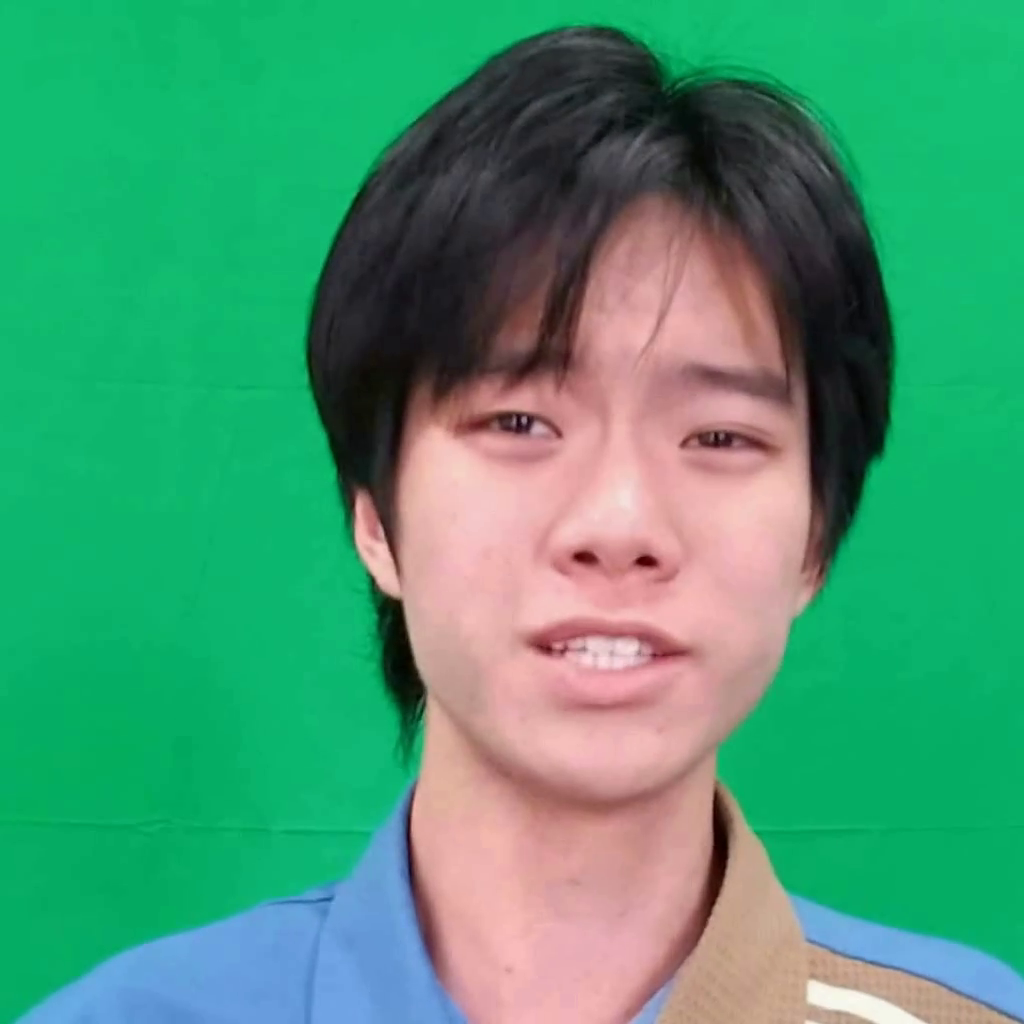}
    \end{subfigure}
    \hspace{-4pt}
        \begin{subfigure}{0.12\linewidth}
        \includegraphics[width=\linewidth]{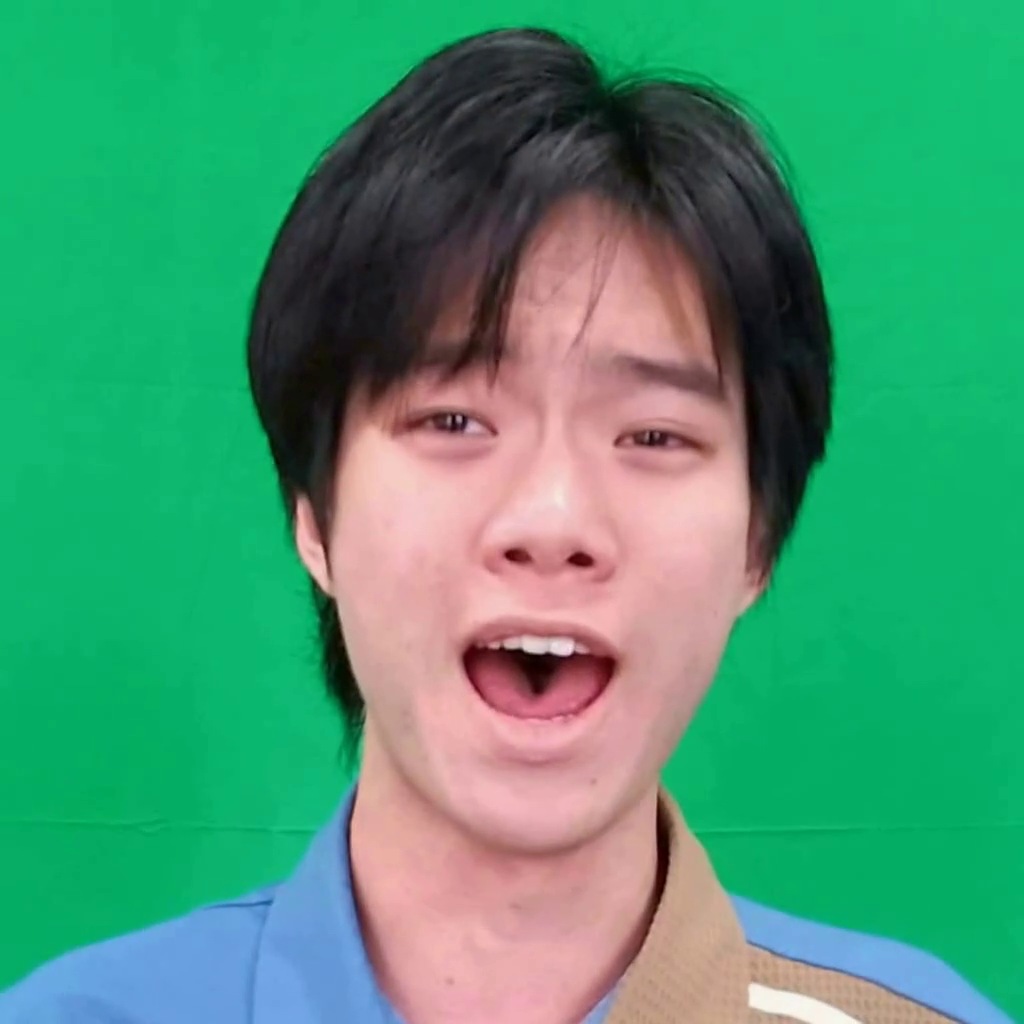}
    \end{subfigure}
    \hspace{-4pt}
        \begin{subfigure}{0.12\linewidth}
        \includegraphics[width=\linewidth]{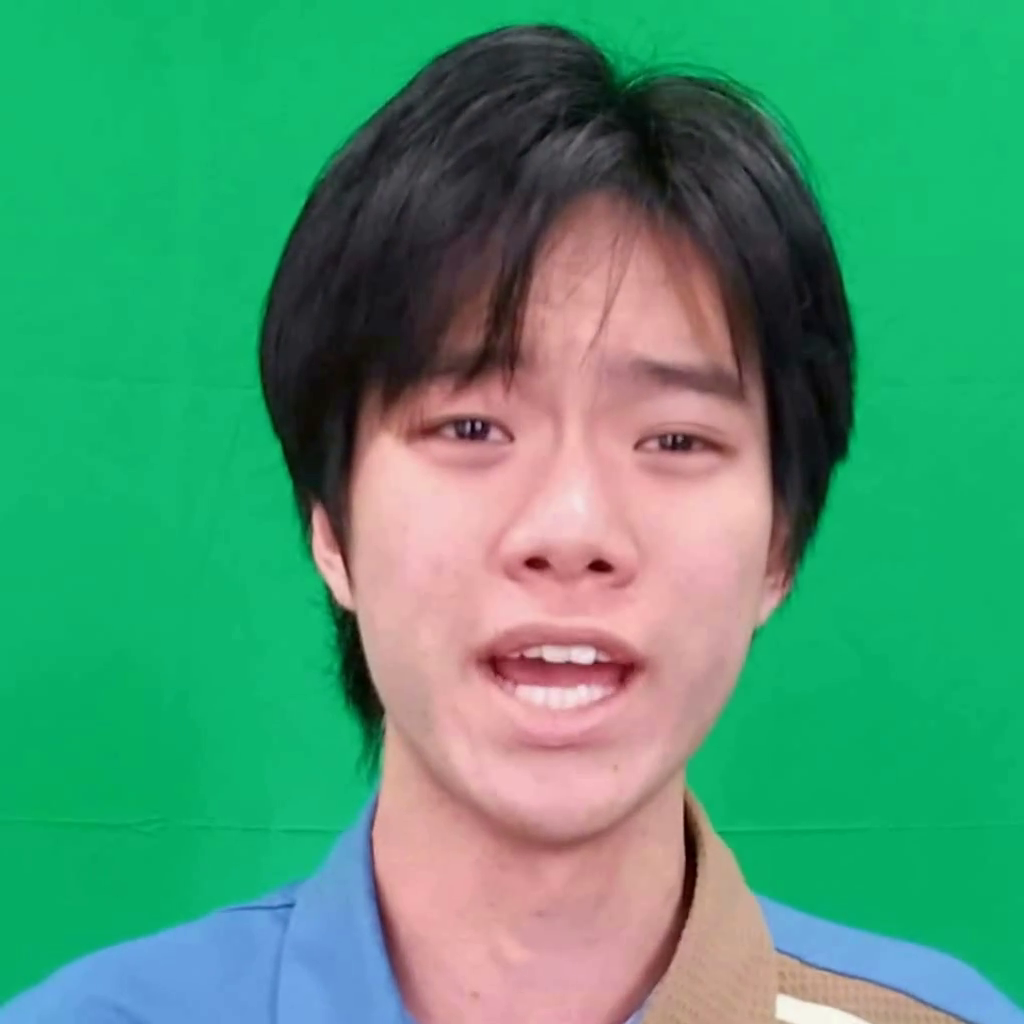}
    \end{subfigure}
    \end{minipage}

    \centering
        \begin{minipage}{0.02\linewidth}
    \centering
        \rotatebox{90}{Audio2Head}
    \end{minipage}
    \begin{minipage}{0.97\linewidth}
    \begin{subfigure}{0.12\linewidth}
        \includegraphics[width=\linewidth]{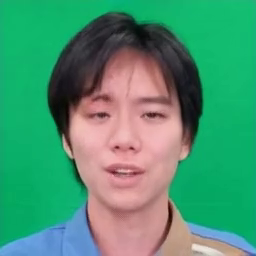}
    \end{subfigure}
    \hspace{-4pt}
        \begin{subfigure}{0.12\linewidth}
        \includegraphics[width=\linewidth]{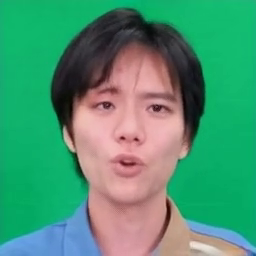}
    \end{subfigure}
     \hspace{-4pt}
        \begin{subfigure}{0.12\linewidth}
        \includegraphics[width=\linewidth]{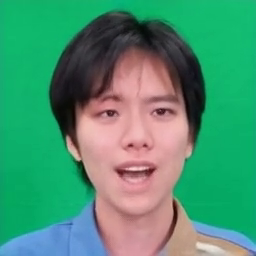}
    \end{subfigure}
     \hspace{-4pt}
        \begin{subfigure}{0.12\linewidth}
        \includegraphics[width=\linewidth]{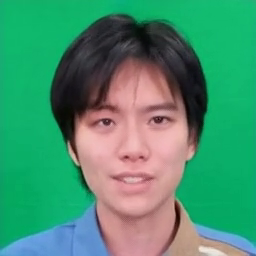}
    \end{subfigure}
     \hspace{-4pt}
        \begin{subfigure}{0.12\linewidth}
        \includegraphics[width=\linewidth]{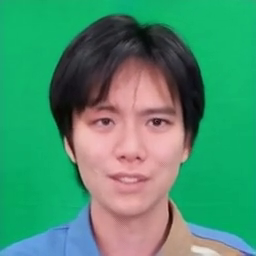}
    \end{subfigure}
     \hspace{-4pt}
        \begin{subfigure}{0.12\linewidth}
        \includegraphics[width=\linewidth]{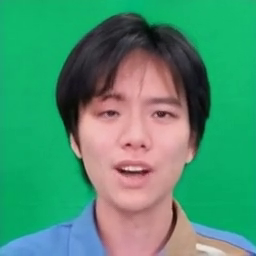}
    \end{subfigure}
    \hspace{-4pt}
        \begin{subfigure}{0.12\linewidth}
        \includegraphics[width=\linewidth]{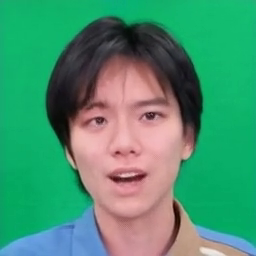}
    \end{subfigure}
    \hspace{-4pt}
        \begin{subfigure}{0.12\linewidth}
        \includegraphics[width=\linewidth]{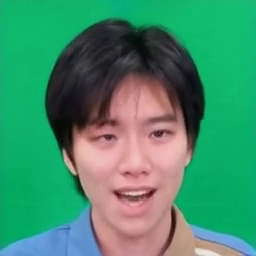}
    \end{subfigure}
    \end{minipage}

    \begin{minipage}{0.02\linewidth}
    \centering
        \rotatebox{90}{SadTalker}
    \end{minipage}
    \begin{minipage}{0.97\linewidth}
    \begin{subfigure}{0.12\linewidth}
        \includegraphics[width=\linewidth]{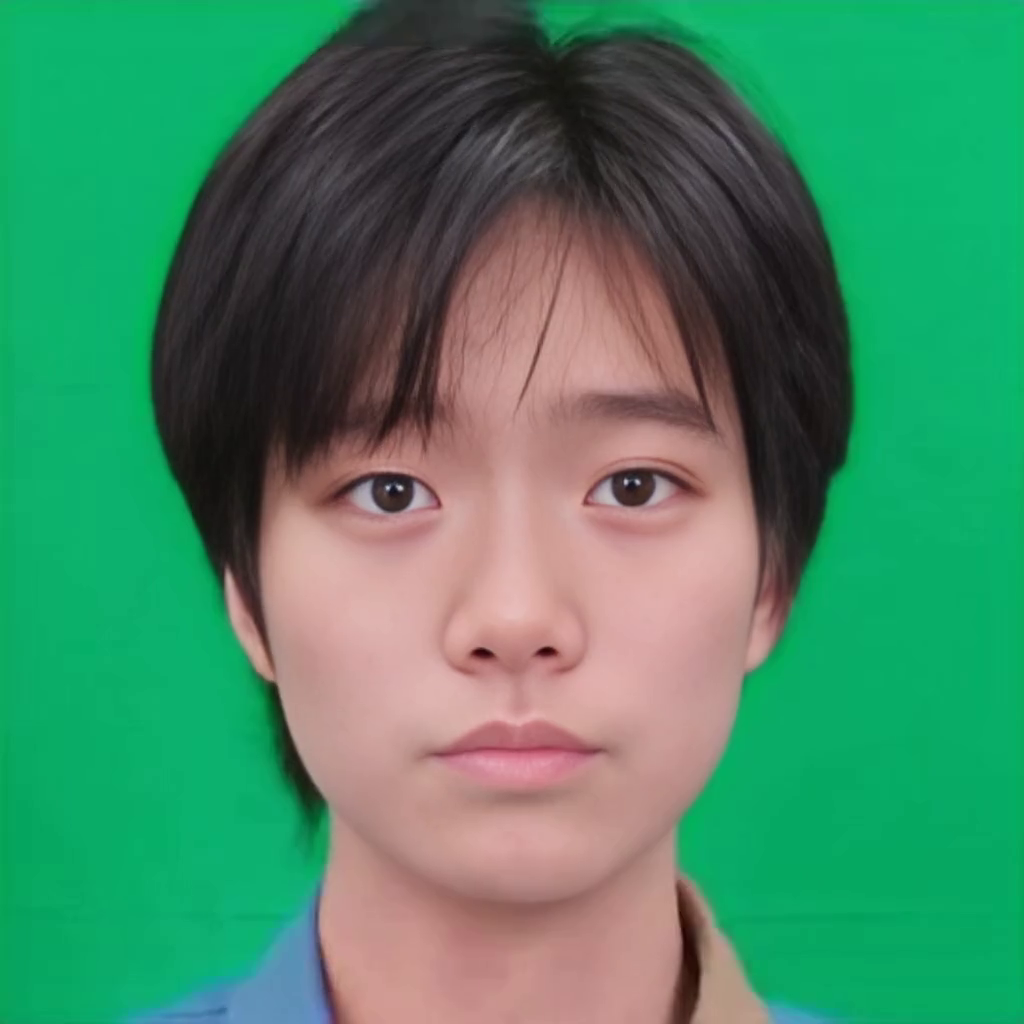}
    \end{subfigure}
    \hspace{-4pt}
        \begin{subfigure}{0.12\linewidth}
        \includegraphics[width=\linewidth]{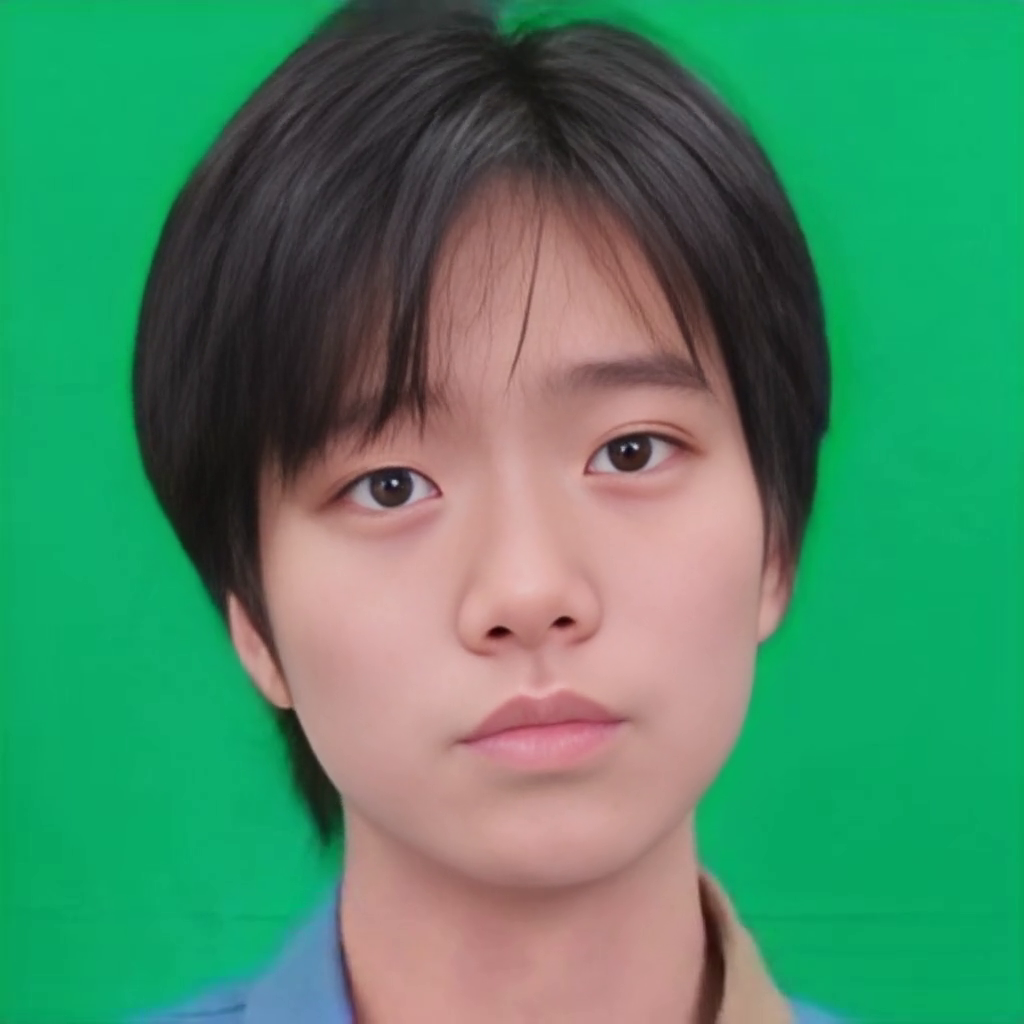}
    \end{subfigure}
     \hspace{-4pt}
        \begin{subfigure}{0.12\linewidth}
        \includegraphics[width=\linewidth]{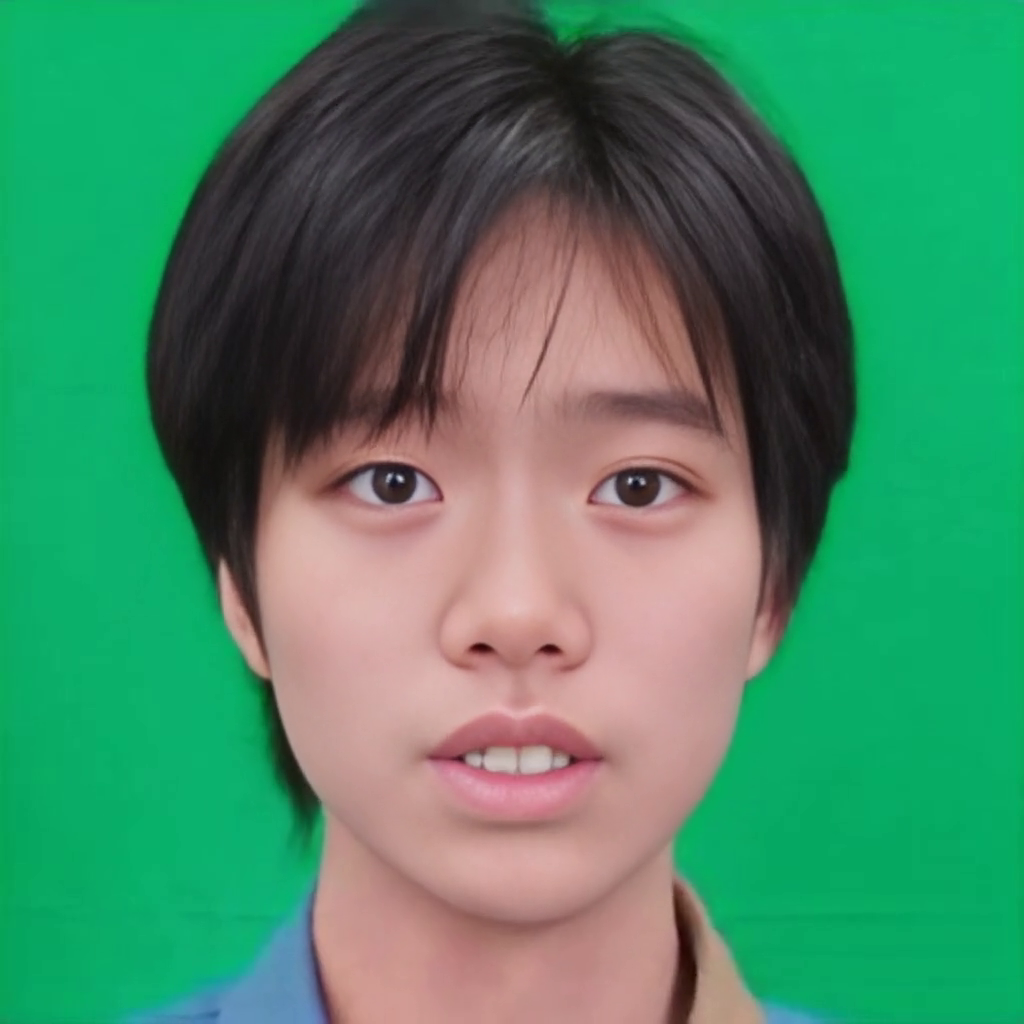}
    \end{subfigure}
     \hspace{-4pt}
        \begin{subfigure}{0.12\linewidth}
        \includegraphics[width=\linewidth]{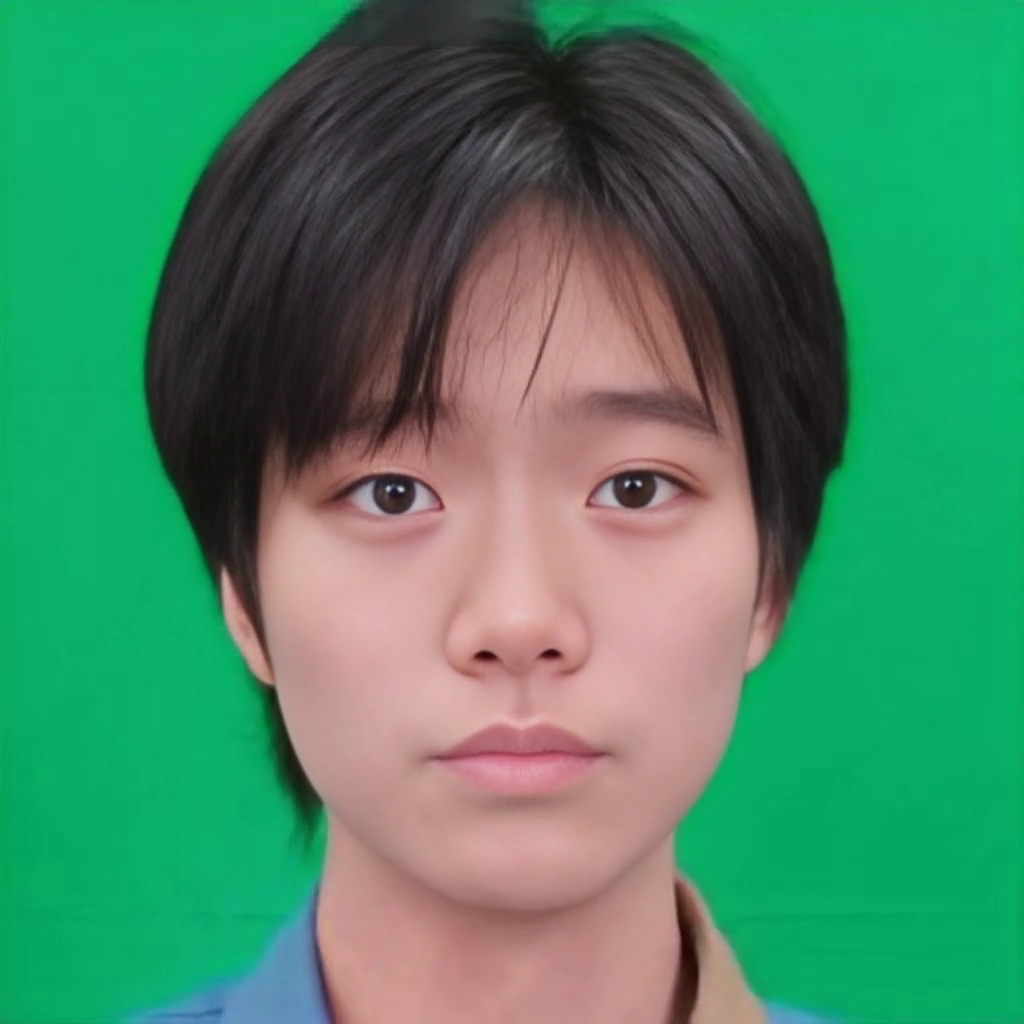}
    \end{subfigure}
     \hspace{-4pt}
        \begin{subfigure}{0.12\linewidth}
        \includegraphics[width=\linewidth]{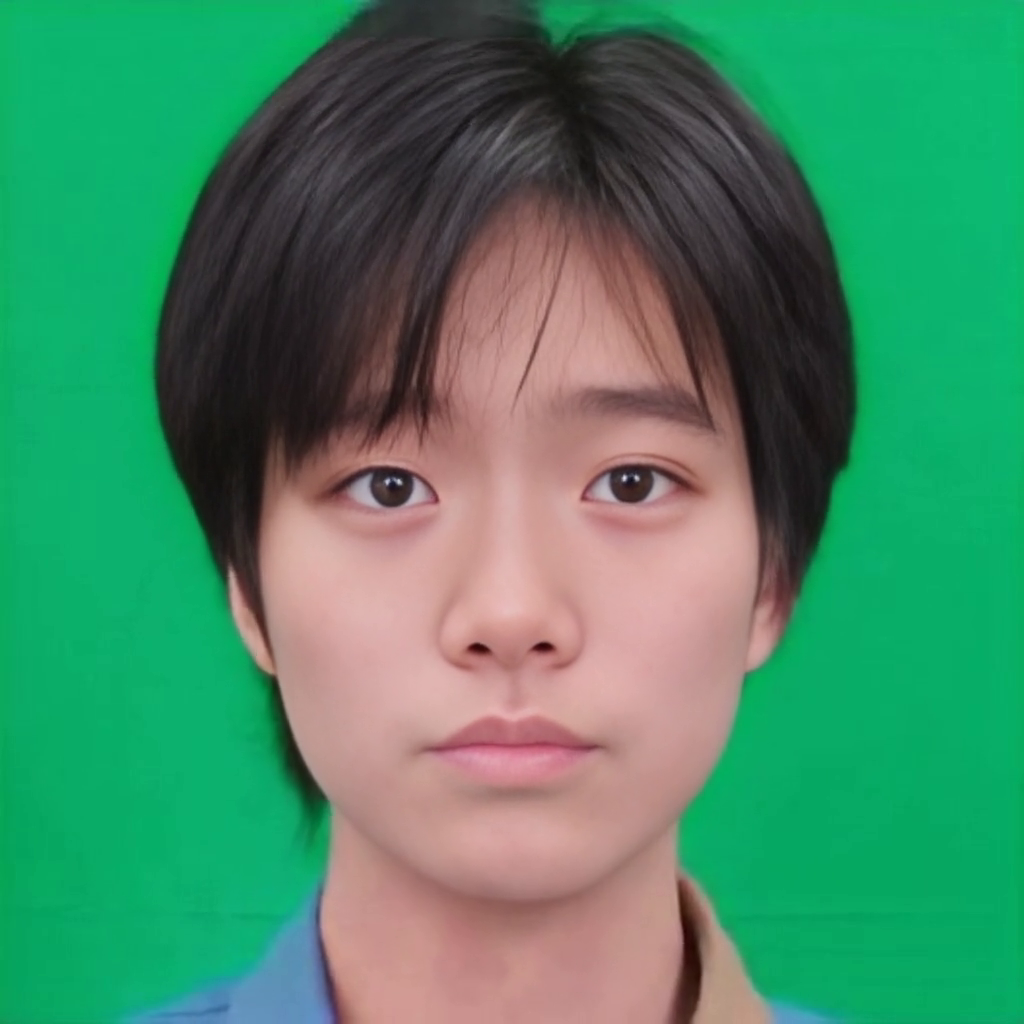}
    \end{subfigure}
     \hspace{-4pt}
        \begin{subfigure}{0.12\linewidth}
        \includegraphics[width=\linewidth]{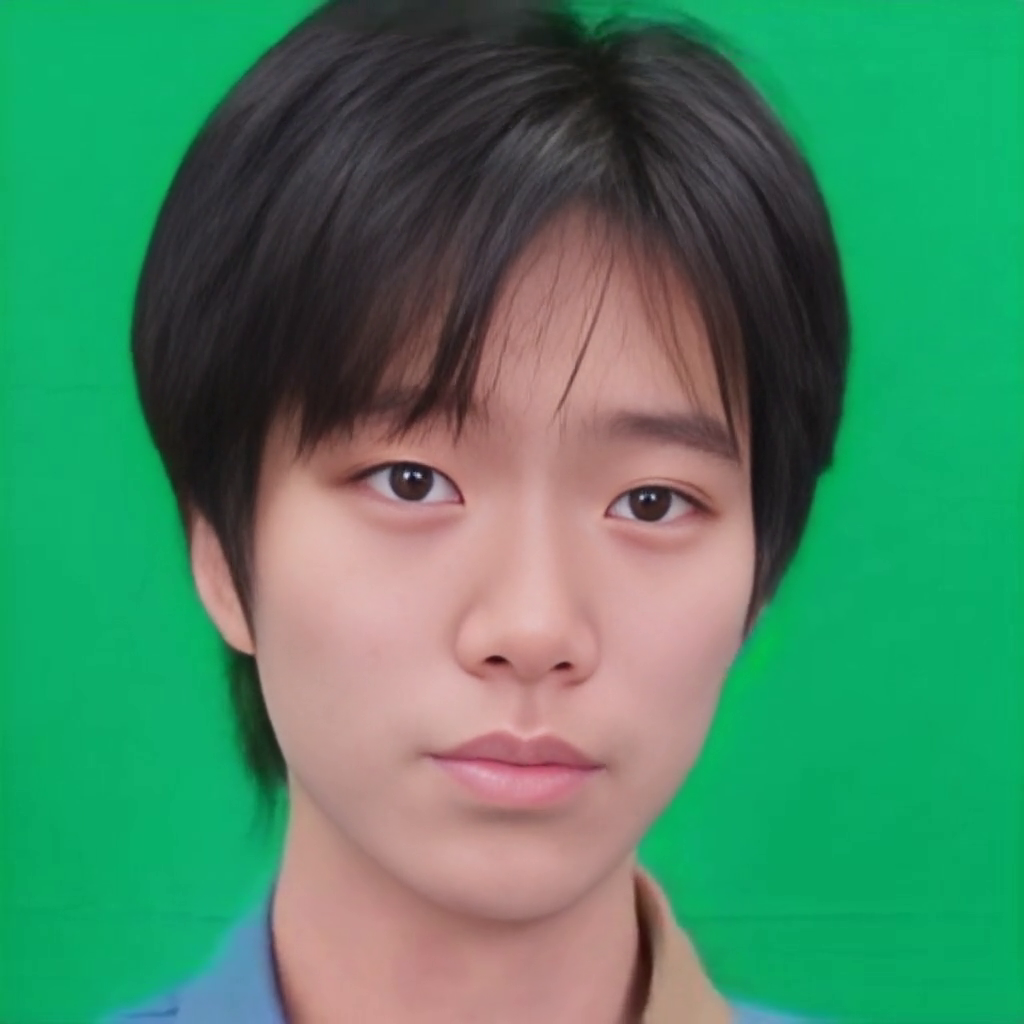}
    \end{subfigure}
    \hspace{-4pt}
        \begin{subfigure}{0.12\linewidth}
        \includegraphics[width=\linewidth]{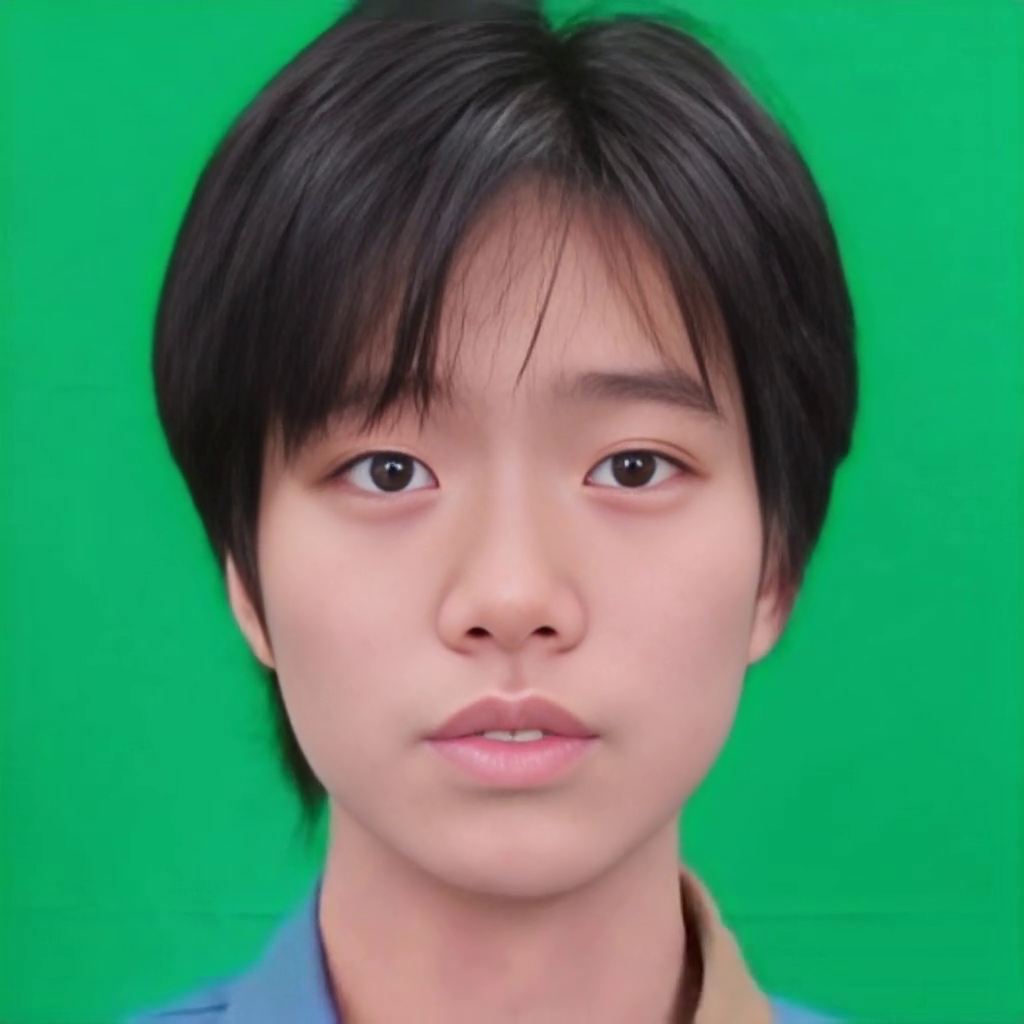}
    \end{subfigure}
    \hspace{-4pt}
        \begin{subfigure}{0.12\linewidth}
        \includegraphics[width=\linewidth]{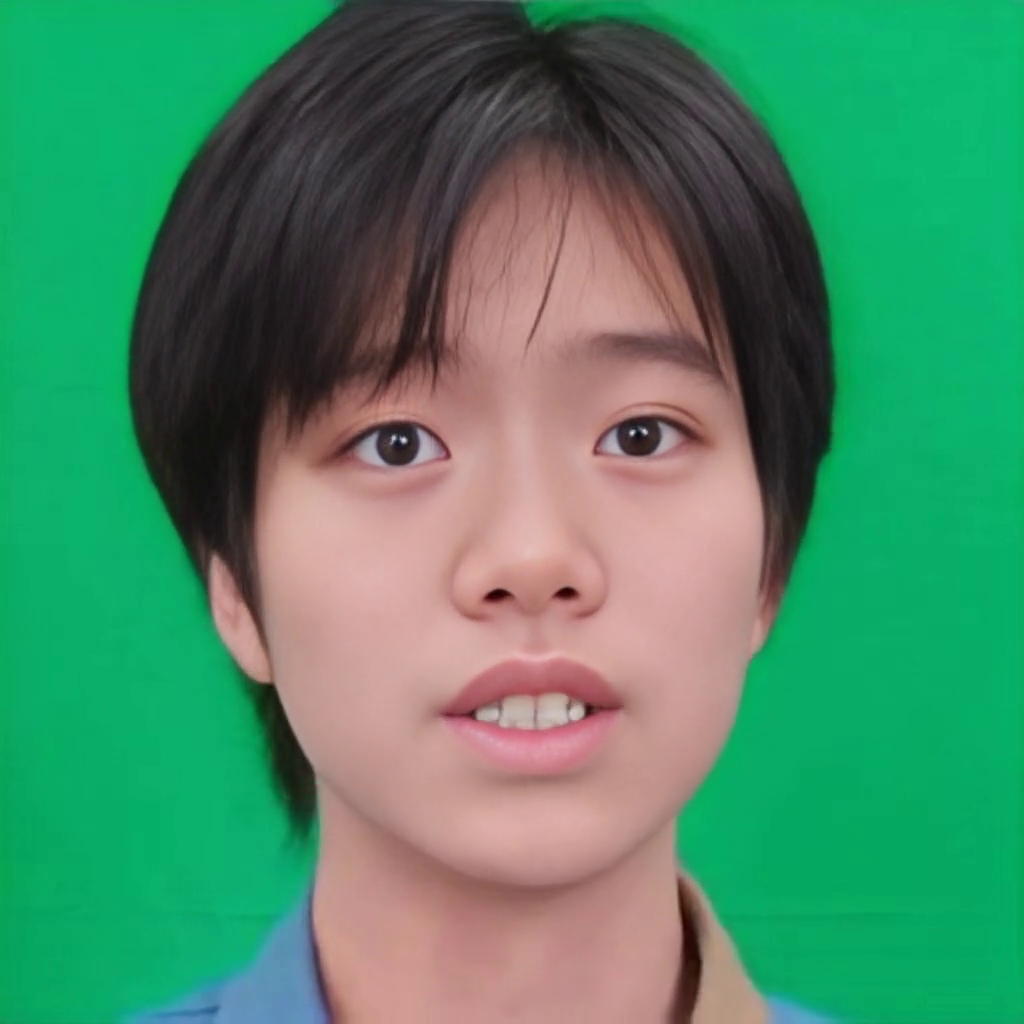}
    \end{subfigure}
    \end{minipage}

    \begin{minipage}{0.02\linewidth}
    \centering
        \rotatebox{90}{MuseTalk}
    \end{minipage}
    \begin{minipage}{0.97\linewidth}
    \begin{subfigure}{0.12\linewidth}
        \includegraphics[width=\linewidth]{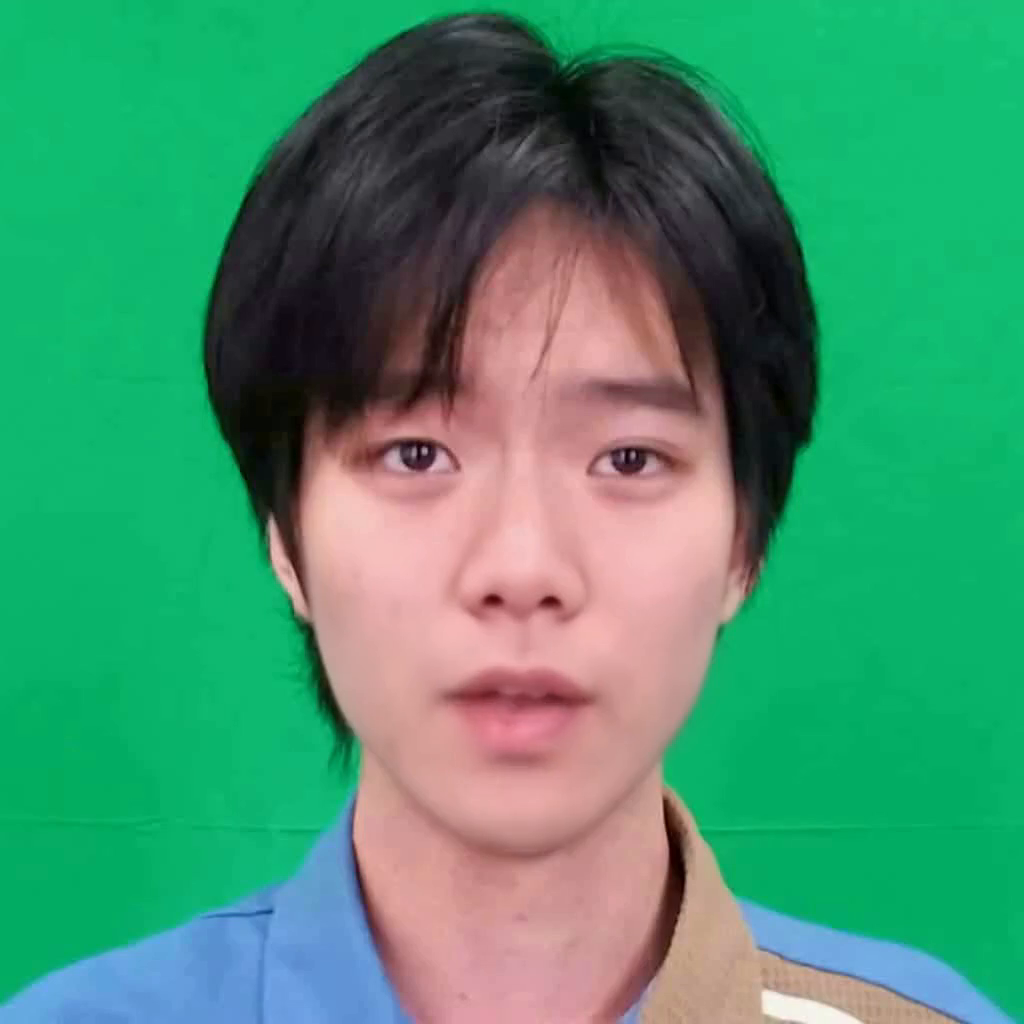}
    \end{subfigure}
    \hspace{-4pt}
        \begin{subfigure}{0.12\linewidth}
        \includegraphics[width=\linewidth]{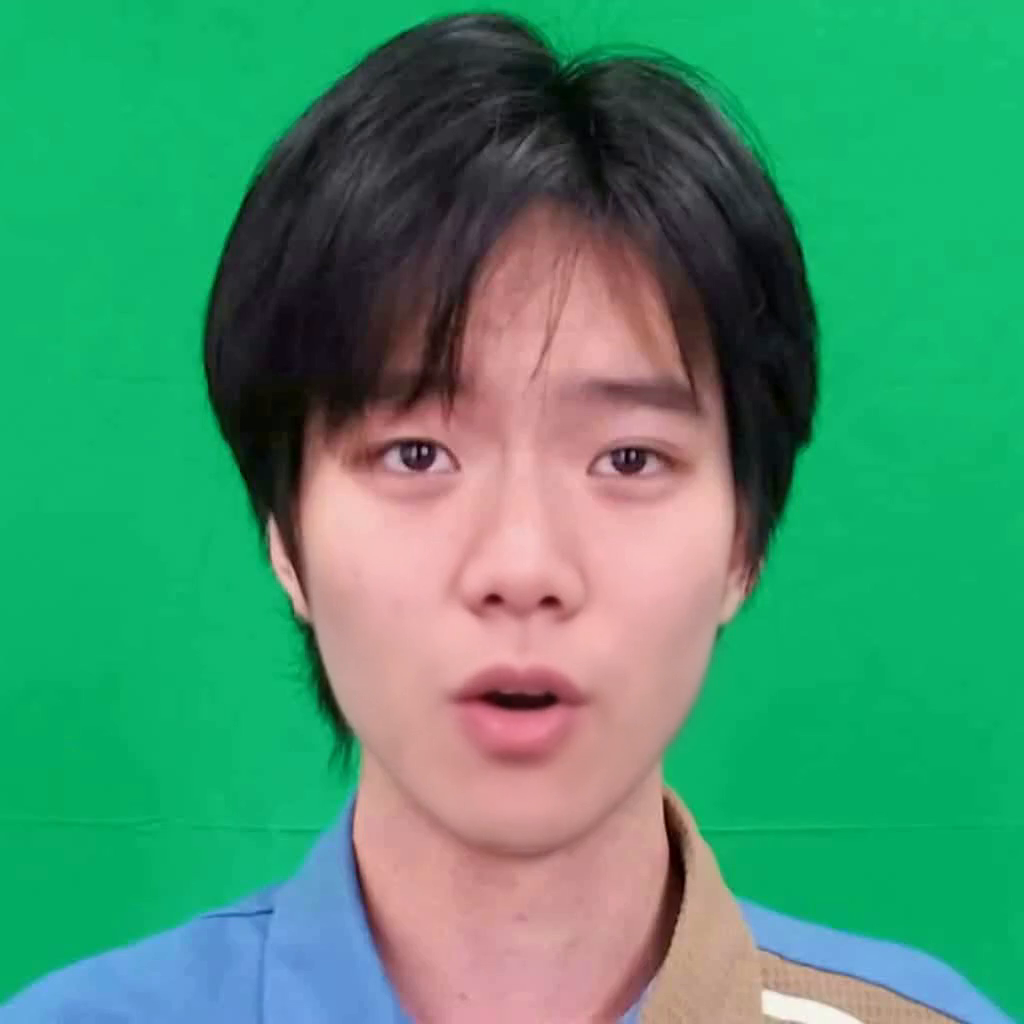}
    \end{subfigure}
     \hspace{-4pt}
        \begin{subfigure}{0.12\linewidth}
        \includegraphics[width=\linewidth]{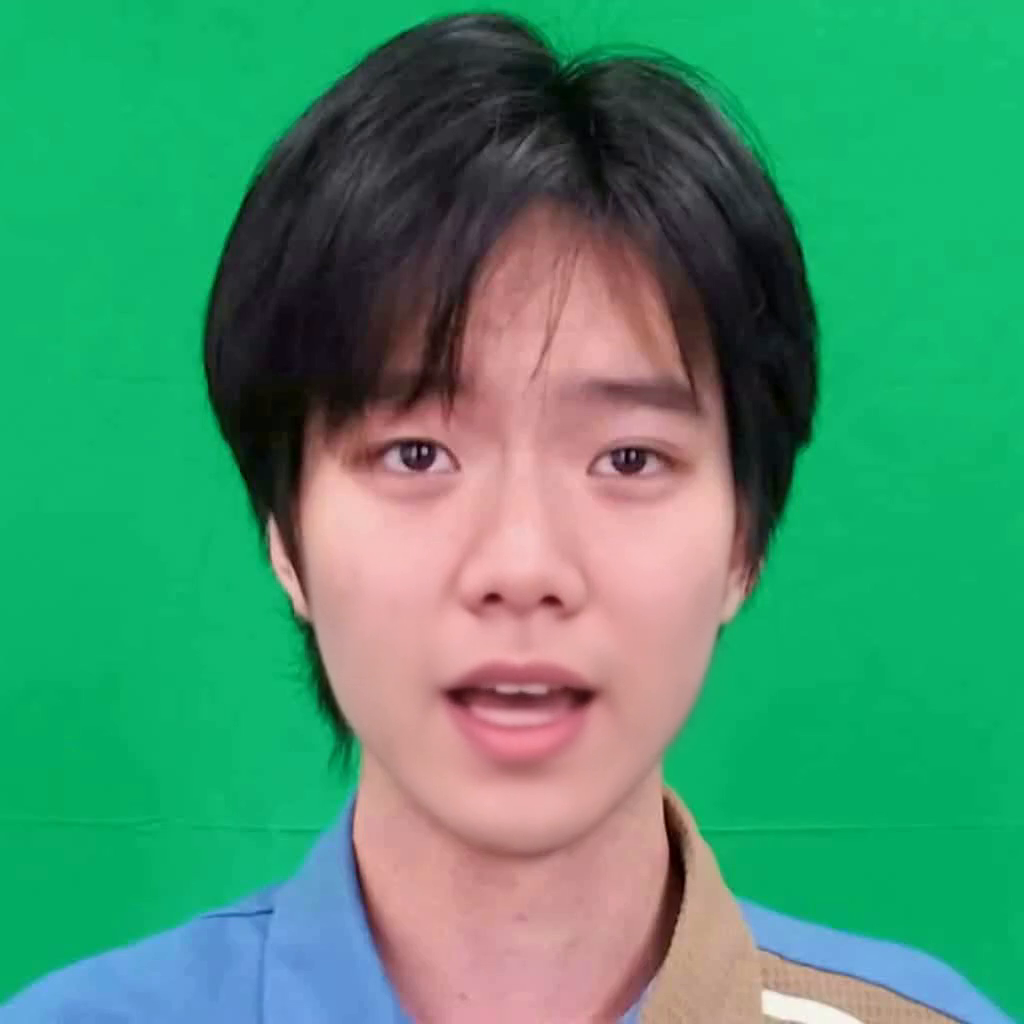}
    \end{subfigure}
     \hspace{-4pt}
        \begin{subfigure}{0.12\linewidth}
        \includegraphics[width=\linewidth]{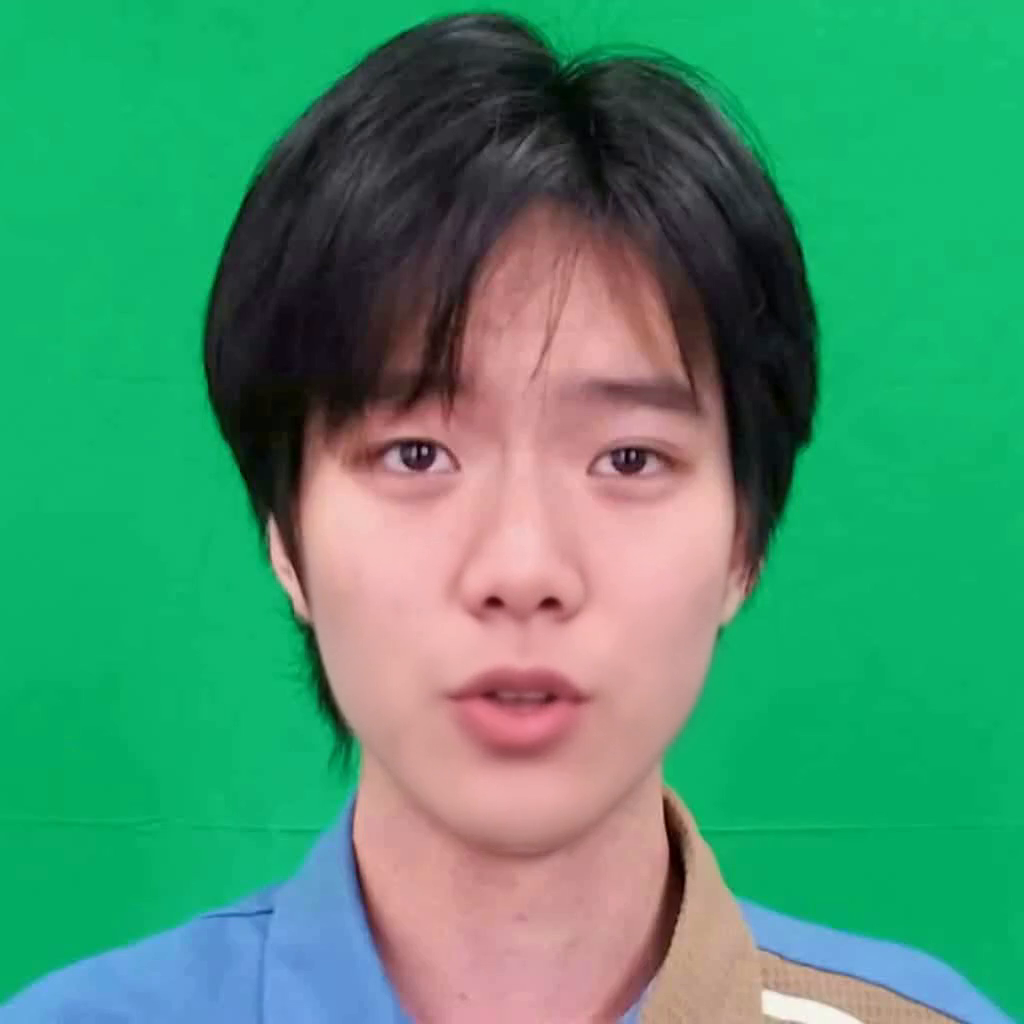}
    \end{subfigure}
     \hspace{-4pt}
        \begin{subfigure}{0.12\linewidth}
        \includegraphics[width=\linewidth]{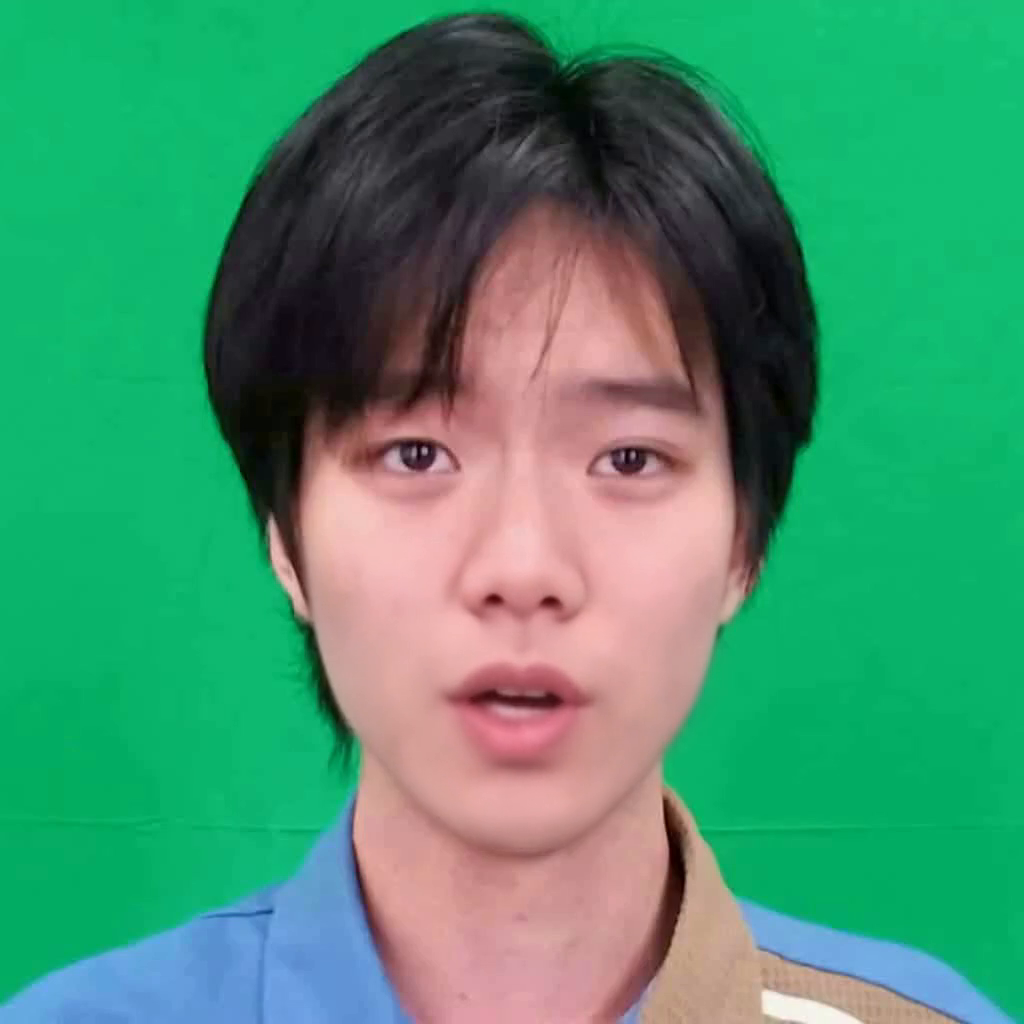}
    \end{subfigure}
     \hspace{-4pt}
        \begin{subfigure}{0.12\linewidth}
        \includegraphics[width=\linewidth]{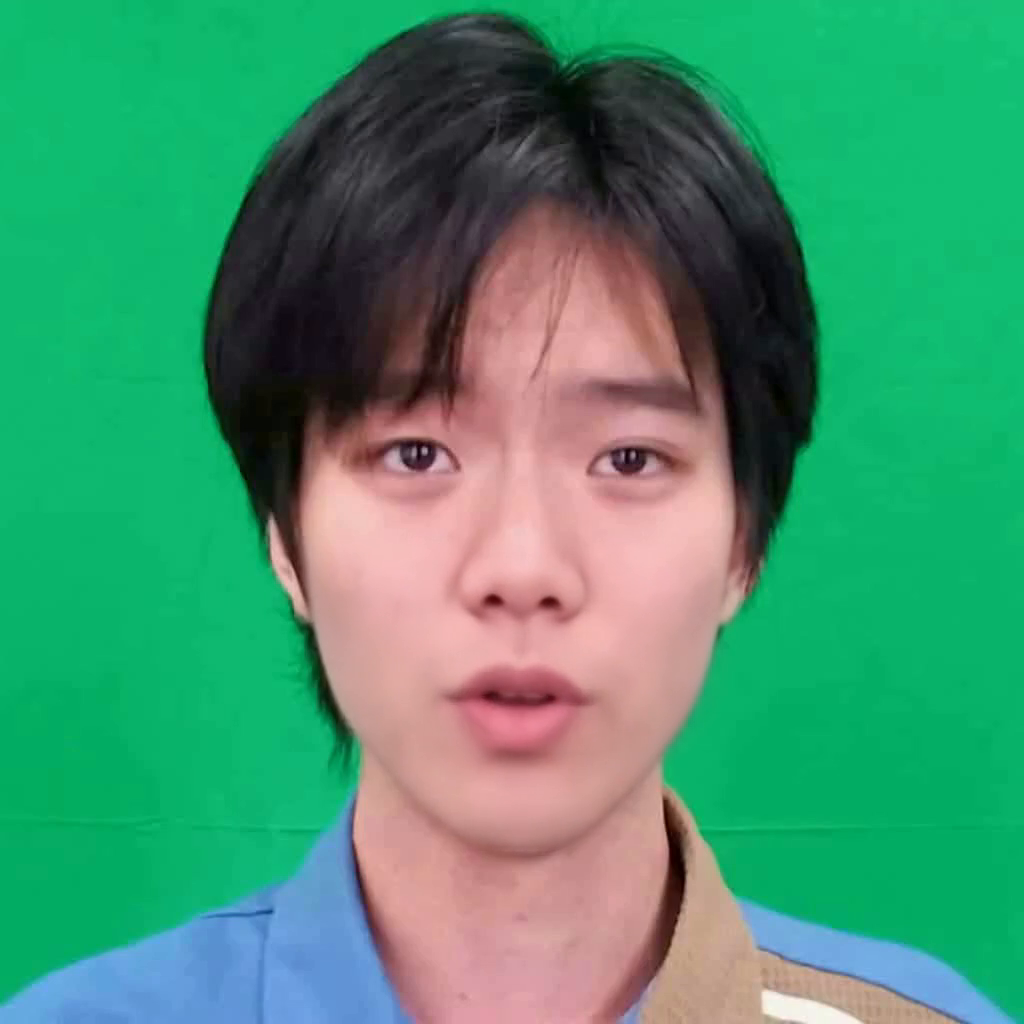}
    \end{subfigure}
    \hspace{-4pt}
        \begin{subfigure}{0.12\linewidth}
        \includegraphics[width=\linewidth]{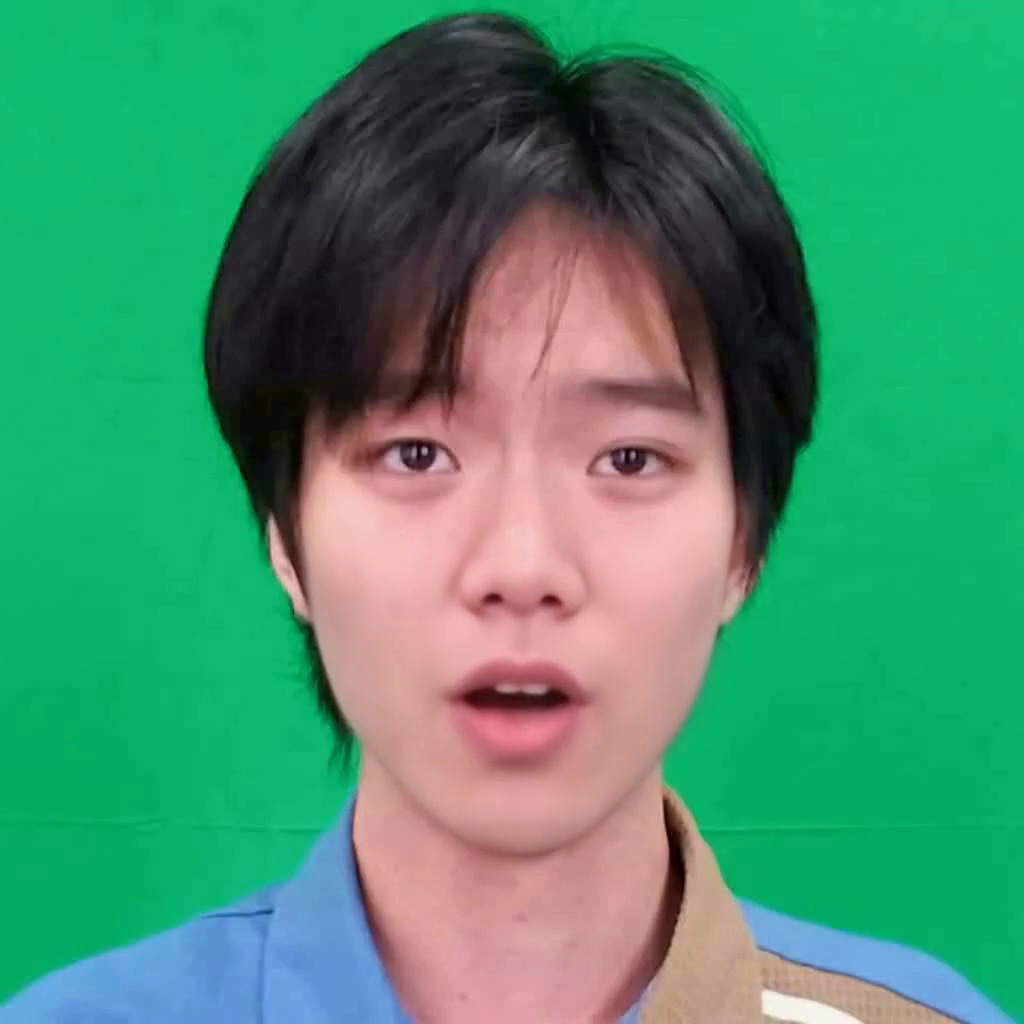}
    \end{subfigure}
    \hspace{-4pt}
        \begin{subfigure}{0.12\linewidth}
        \includegraphics[width=\linewidth]{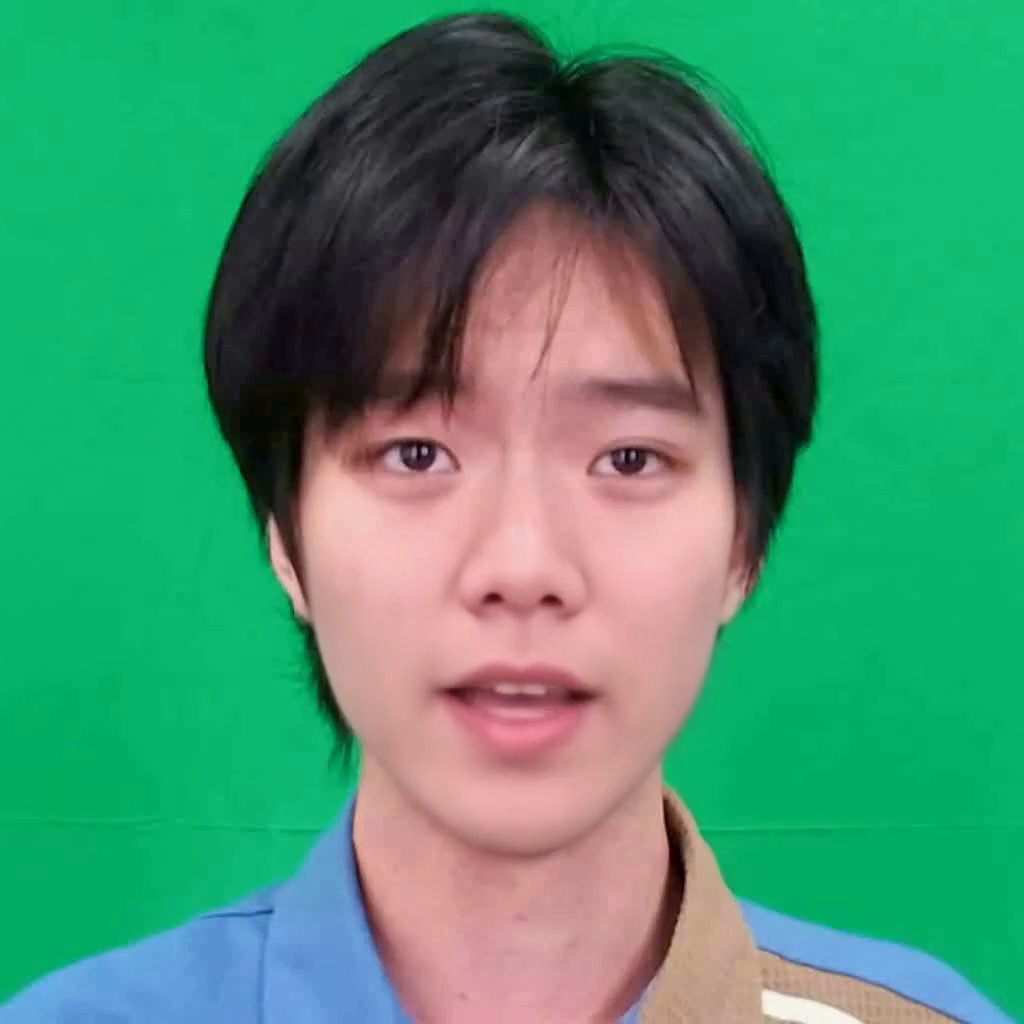}
    \end{subfigure}
    \end{minipage}

      \begin{minipage}{0.02\linewidth}
    \centering
        \rotatebox{90}{AniPortrait}
    \end{minipage}
    \begin{minipage}{0.97\linewidth}
    \begin{subfigure}{0.12\linewidth}
        \includegraphics[width=\linewidth]{figures/show_cases_chinese/musetalk1/id11_26_5_1_0001.png}
    \end{subfigure}
    \hspace{-4pt}
        \begin{subfigure}{0.12\linewidth}
        \includegraphics[width=\linewidth]{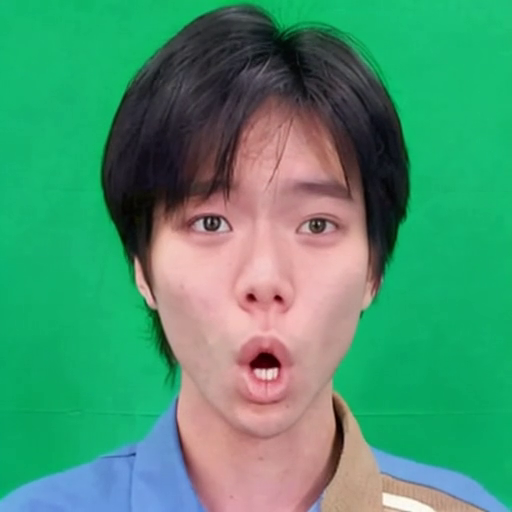}
    \end{subfigure}
     \hspace{-4pt}
        \begin{subfigure}{0.12\linewidth}
        \includegraphics[width=\linewidth]{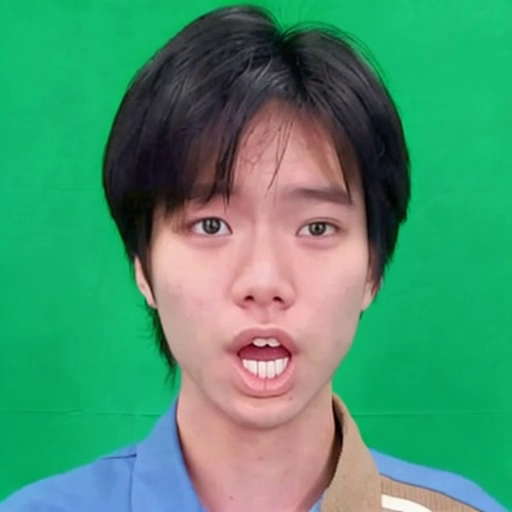}
    \end{subfigure}
     \hspace{-4pt}
        \begin{subfigure}{0.12\linewidth}
        \includegraphics[width=\linewidth]{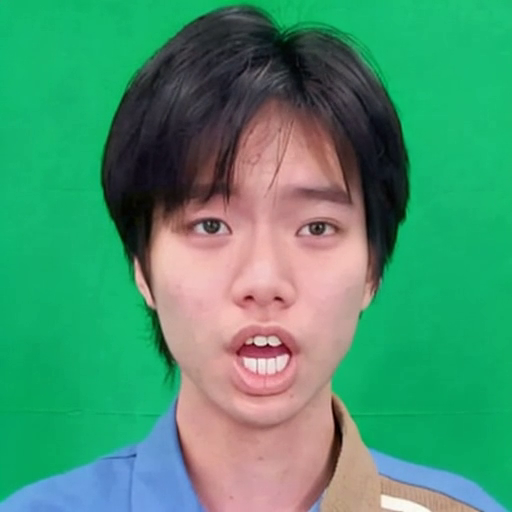}
    \end{subfigure}
     \hspace{-4pt}
        \begin{subfigure}{0.12\linewidth}
        \includegraphics[width=\linewidth]{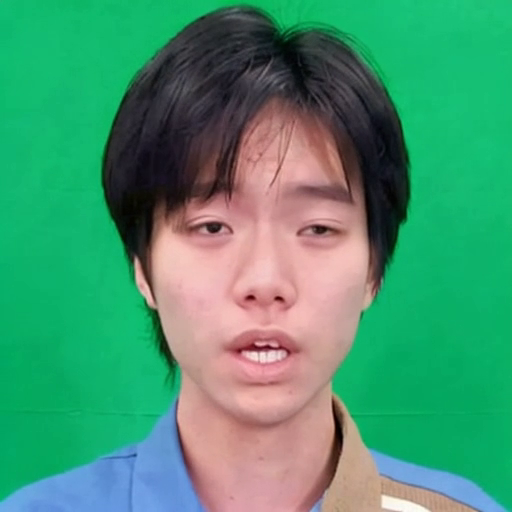}
    \end{subfigure}
     \hspace{-4pt}
        \begin{subfigure}{0.12\linewidth}
        \includegraphics[width=\linewidth]{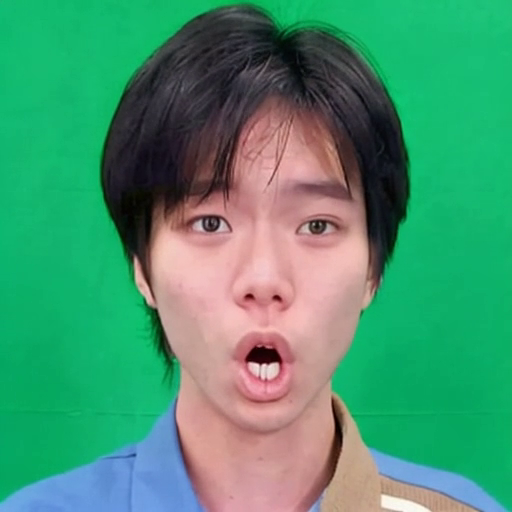}
    \end{subfigure}
    \hspace{-4pt}
        \begin{subfigure}{0.12\linewidth}
        \includegraphics[width=\linewidth]{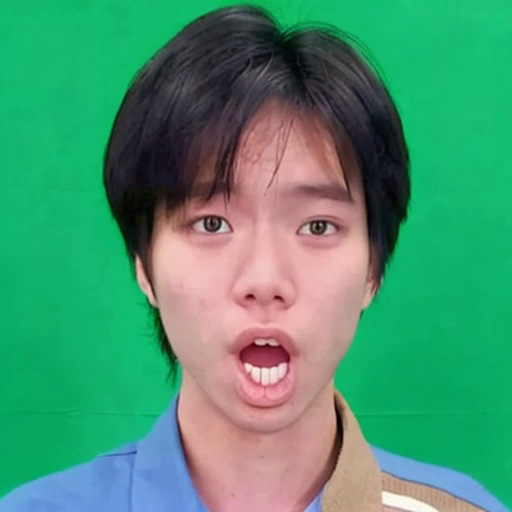}
    \end{subfigure}
    \hspace{-4pt}
        \begin{subfigure}{0.12\linewidth}
        \includegraphics[width=\linewidth]{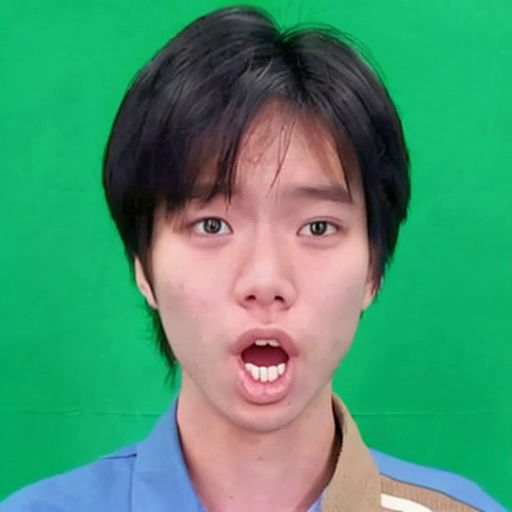}
    \end{subfigure}
    \end{minipage}

    \begin{minipage}{0.02\linewidth}
    \centering
        \rotatebox{90}{Echomimic}
    \end{minipage}
    \begin{minipage}{0.97\linewidth}
    \begin{subfigure}{0.12\linewidth}
        \includegraphics[width=\linewidth]{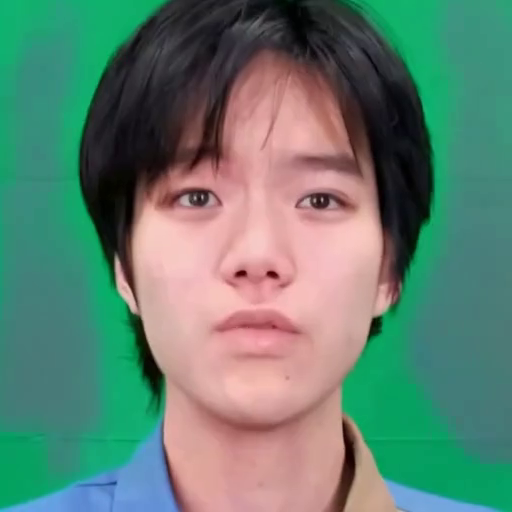}
    \end{subfigure}
    \hspace{-4pt}
        \begin{subfigure}{0.12\linewidth}
        \includegraphics[width=\linewidth]{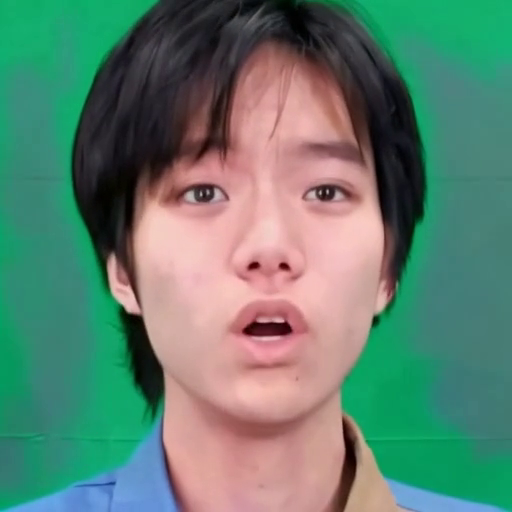}
    \end{subfigure}
     \hspace{-4pt}
        \begin{subfigure}{0.12\linewidth}
        \includegraphics[width=\linewidth]{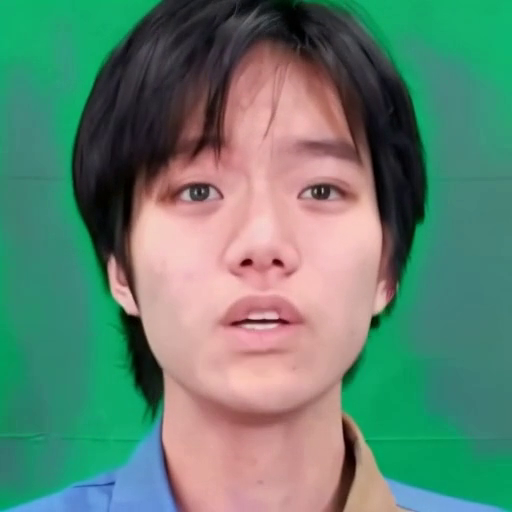}
    \end{subfigure}
     \hspace{-4pt}
        \begin{subfigure}{0.12\linewidth}
        \includegraphics[width=\linewidth]{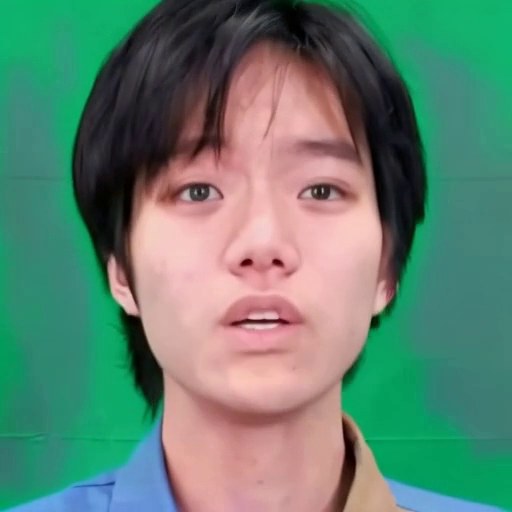}
    \end{subfigure}
     \hspace{-4pt}
        \begin{subfigure}{0.12\linewidth}
        \includegraphics[width=\linewidth]{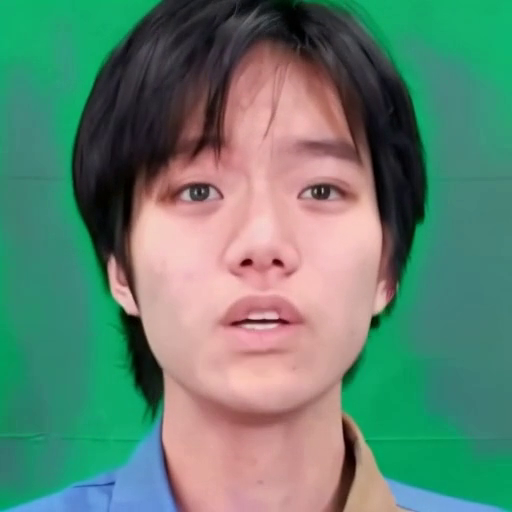}
    \end{subfigure}
     \hspace{-4pt}
        \begin{subfigure}{0.12\linewidth}
        \includegraphics[width=\linewidth]{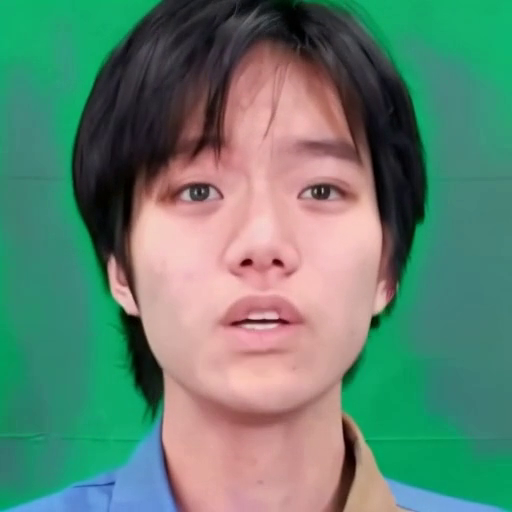}
    \end{subfigure}
    \hspace{-4pt}
        \begin{subfigure}{0.12\linewidth}
        \includegraphics[width=\linewidth]{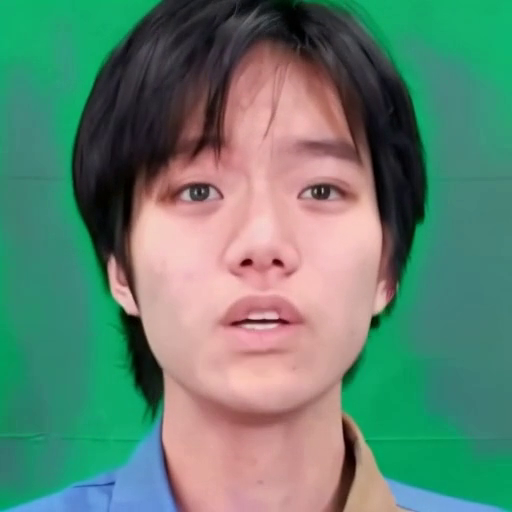}
    \end{subfigure}
    \hspace{-4pt}
        \begin{subfigure}{0.12\linewidth}
        \includegraphics[width=\linewidth]{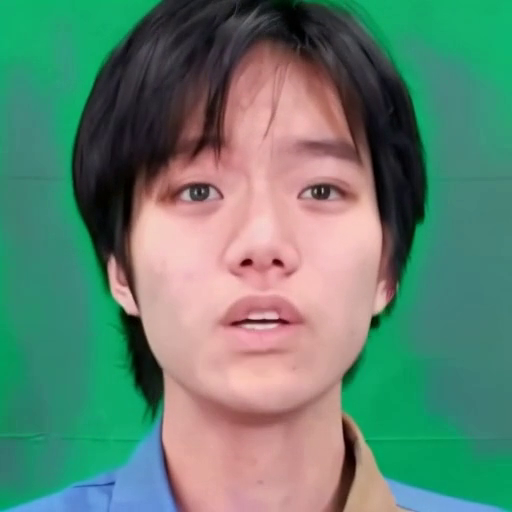}
    \end{subfigure}
    \end{minipage}

      \begin{minipage}{0.02\linewidth}
    \centering
        \rotatebox{90}{Hallo}
    \end{minipage}
    \begin{minipage}{0.97\linewidth}
    \begin{subfigure}{0.12\linewidth}
        \includegraphics[width=\linewidth]{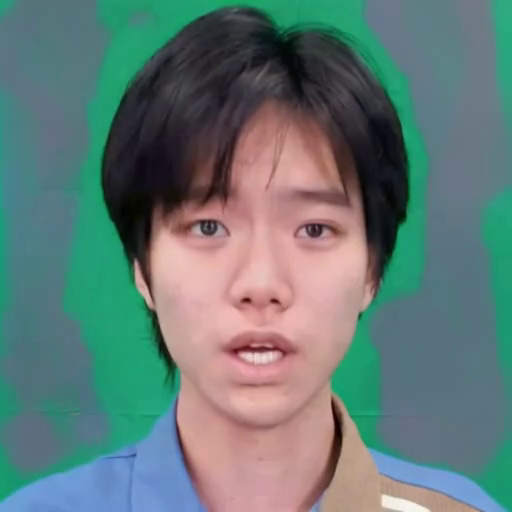}
    \end{subfigure}
    \hspace{-4pt}
        \begin{subfigure}{0.12\linewidth}
        \includegraphics[width=\linewidth]{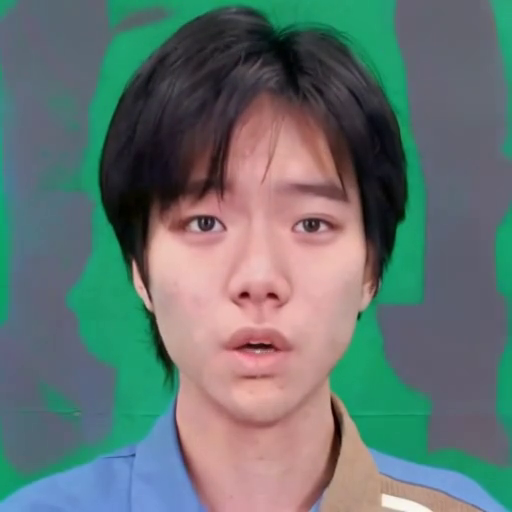}
    \end{subfigure}
     \hspace{-4pt}
        \begin{subfigure}{0.12\linewidth}
        \includegraphics[width=\linewidth]{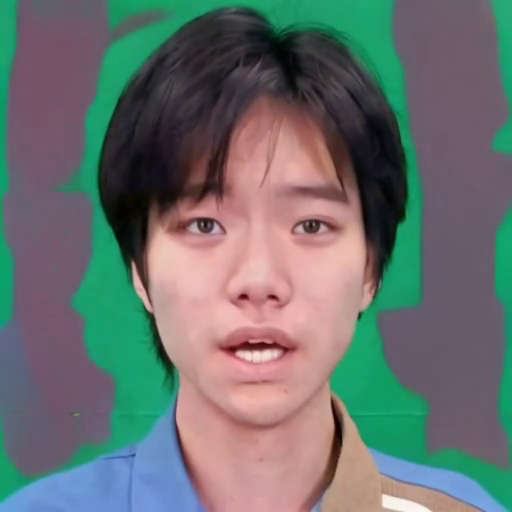}
    \end{subfigure}
     \hspace{-4pt}
        \begin{subfigure}{0.12\linewidth}
        \includegraphics[width=\linewidth]{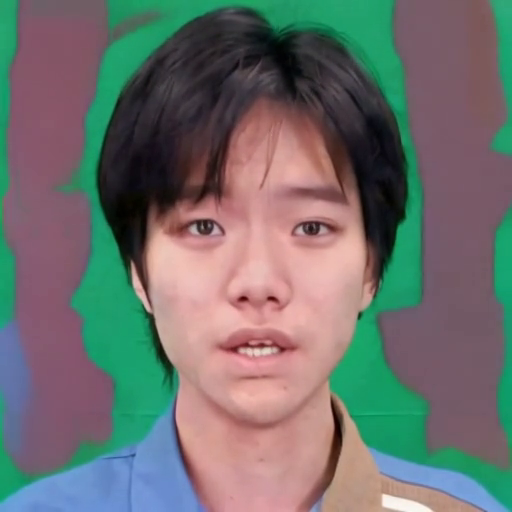}
    \end{subfigure}
     \hspace{-4pt}
        \begin{subfigure}{0.12\linewidth}
        \includegraphics[width=\linewidth]{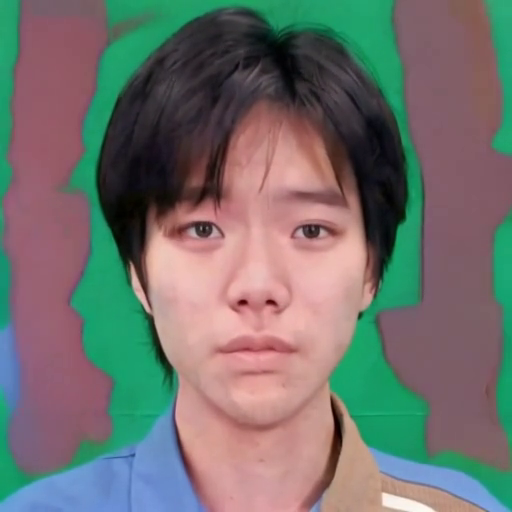}
    \end{subfigure}
     \hspace{-4pt}
        \begin{subfigure}{0.12\linewidth}
        \includegraphics[width=\linewidth]{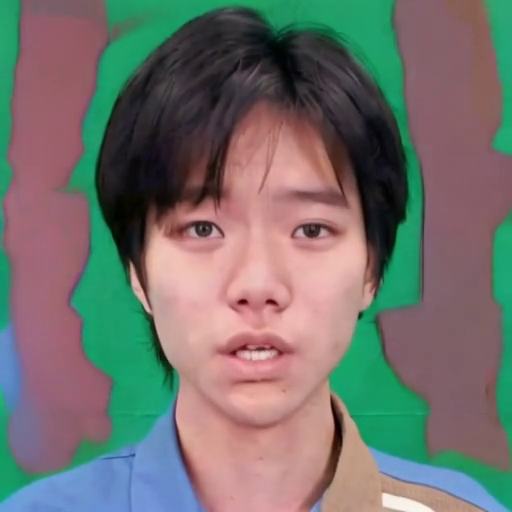}
    \end{subfigure}
    \hspace{-4pt}
        \begin{subfigure}{0.12\linewidth}
        \includegraphics[width=\linewidth]{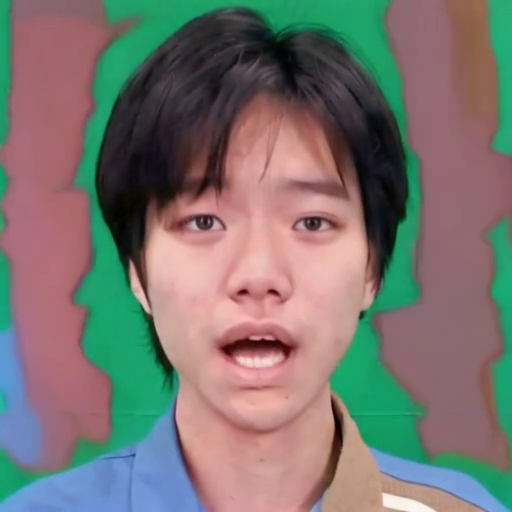}
    \end{subfigure}
    \hspace{-4pt}
        \begin{subfigure}{0.12\linewidth}
        \includegraphics[width=\linewidth]{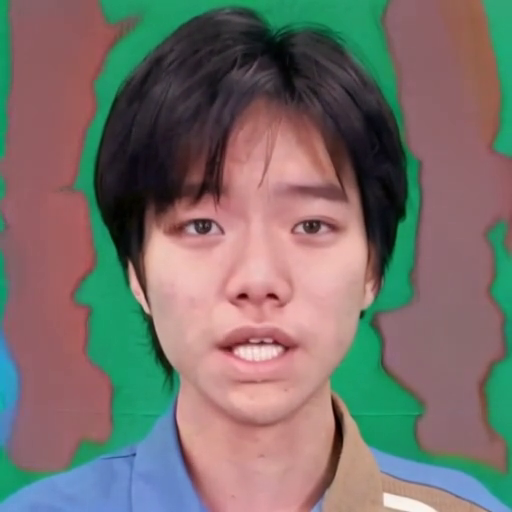}
    \end{subfigure}
    \end{minipage}

      \begin{minipage}{0.02\linewidth}
    \centering
        \rotatebox{90}{Hallo2}
    \end{minipage}
    \begin{minipage}{0.97\linewidth}
    \begin{subfigure}{0.12\linewidth}
        \includegraphics[width=\linewidth]{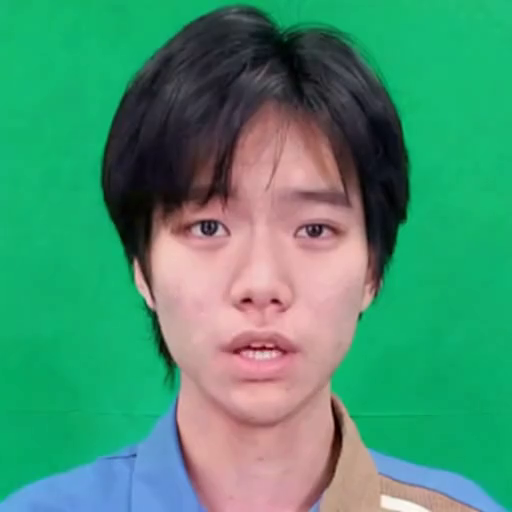}
    \end{subfigure}
    \hspace{-4pt}
        \begin{subfigure}{0.12\linewidth}
        \includegraphics[width=\linewidth]{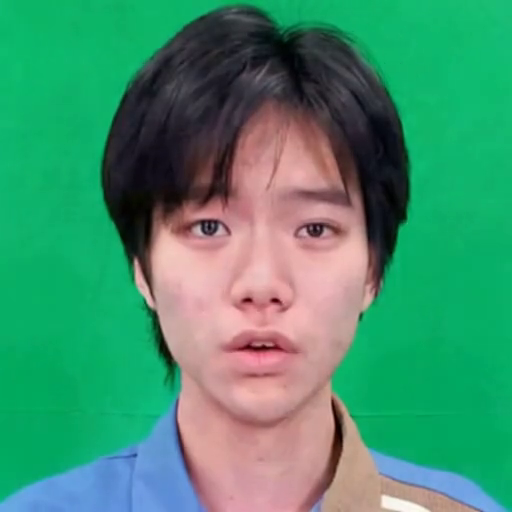}
    \end{subfigure}
     \hspace{-4pt}
        \begin{subfigure}{0.12\linewidth}
        \includegraphics[width=\linewidth]{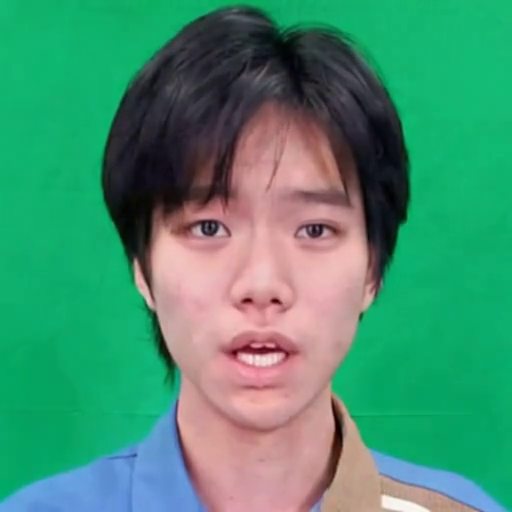}
    \end{subfigure}
     \hspace{-4pt}
        \begin{subfigure}{0.12\linewidth}
        \includegraphics[width=\linewidth]{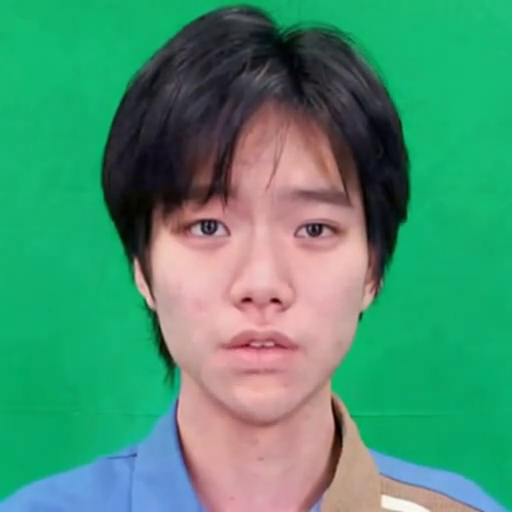}
    \end{subfigure}
     \hspace{-4pt}
        \begin{subfigure}{0.12\linewidth}
        \includegraphics[width=\linewidth]{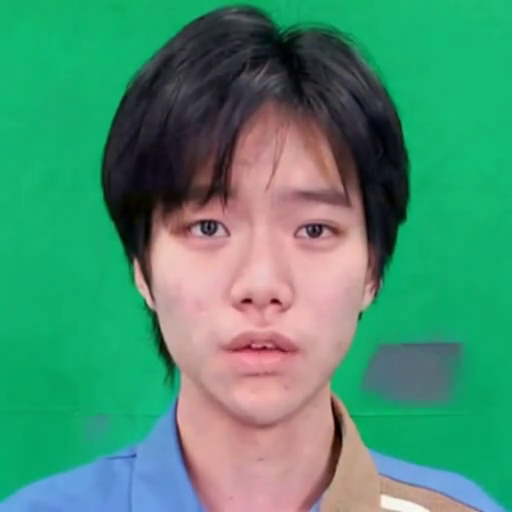}
    \end{subfigure}
     \hspace{-4pt}
        \begin{subfigure}{0.12\linewidth}
        \includegraphics[width=\linewidth]{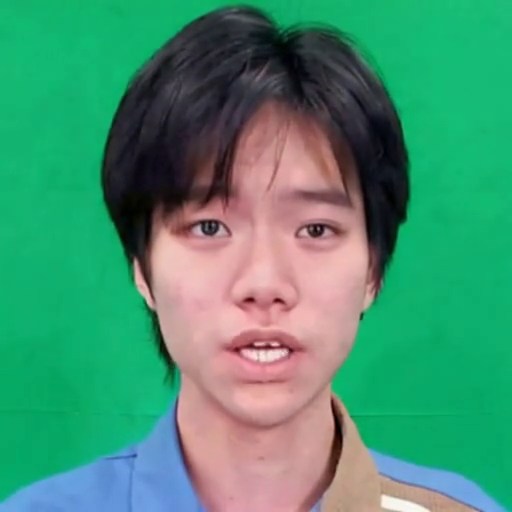}
    \end{subfigure}
    \hspace{-4pt}
        \begin{subfigure}{0.12\linewidth}
        \includegraphics[width=\linewidth]{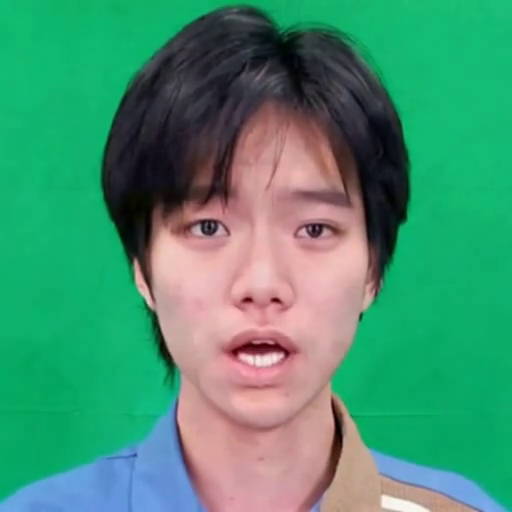}
    \end{subfigure}
    \hspace{-4pt}
        \begin{subfigure}{0.12\linewidth}
        \includegraphics[width=\linewidth]{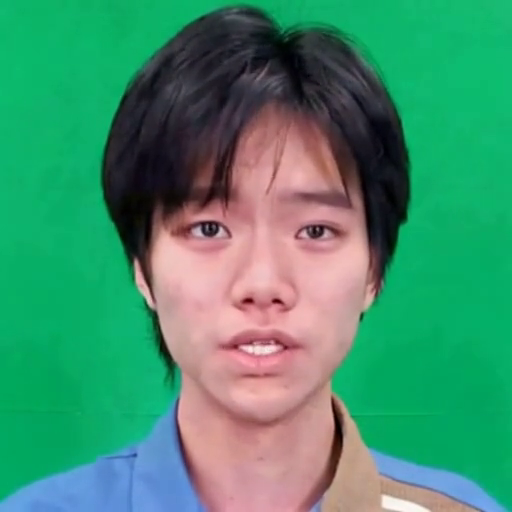}
    \end{subfigure}
    \end{minipage}

    \begin{minipage}{0.02\linewidth}
    \centering
        \rotatebox{90}{\model}
    \end{minipage}
    \begin{minipage}{0.97\linewidth}
    \begin{subfigure}{0.12\linewidth}
        \includegraphics[width=\linewidth]{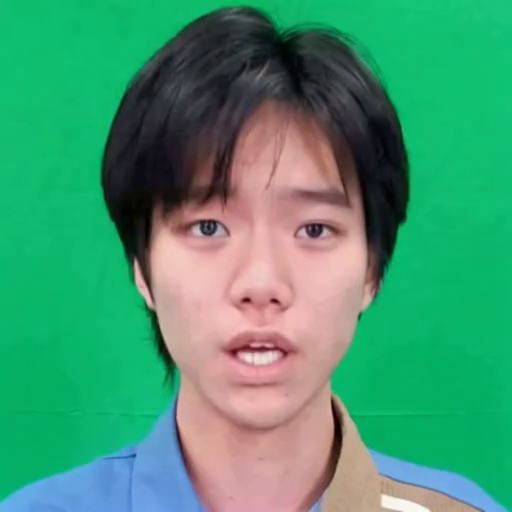}
    \end{subfigure}
    \hspace{-4pt}
        \begin{subfigure}{0.12\linewidth}
        \includegraphics[width=\linewidth]{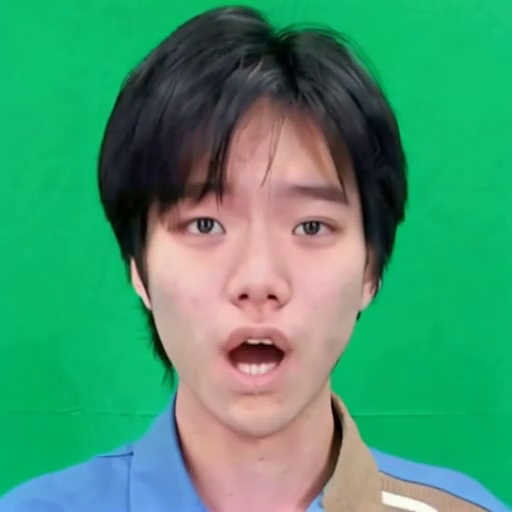}
    \end{subfigure}
     \hspace{-4pt}
        \begin{subfigure}{0.12\linewidth}
        \includegraphics[width=\linewidth]{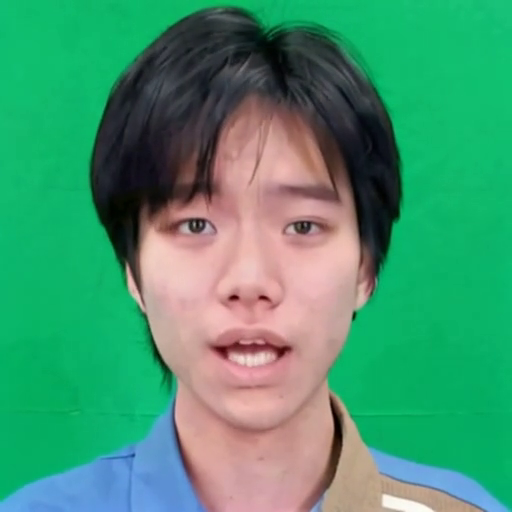}
    \end{subfigure}
     \hspace{-4pt}
        \begin{subfigure}{0.12\linewidth}
        \includegraphics[width=\linewidth]{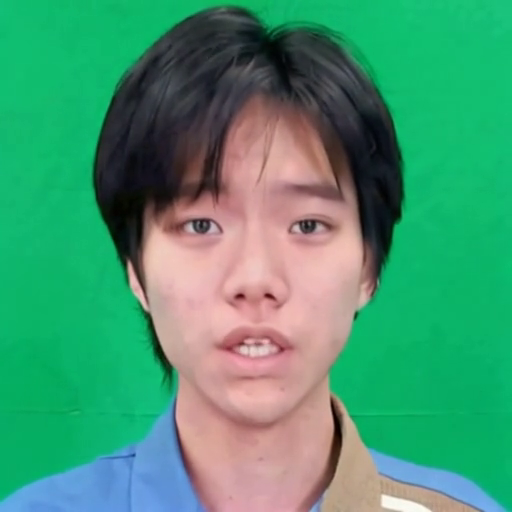}
    \end{subfigure}
     \hspace{-4pt}
        \begin{subfigure}{0.12\linewidth}
        \includegraphics[width=\linewidth]{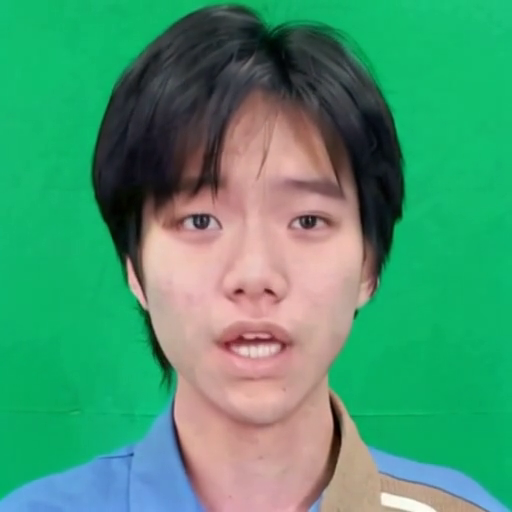}
    \end{subfigure}
     \hspace{-4pt}
        \begin{subfigure}{0.12\linewidth}
        \includegraphics[width=\linewidth]{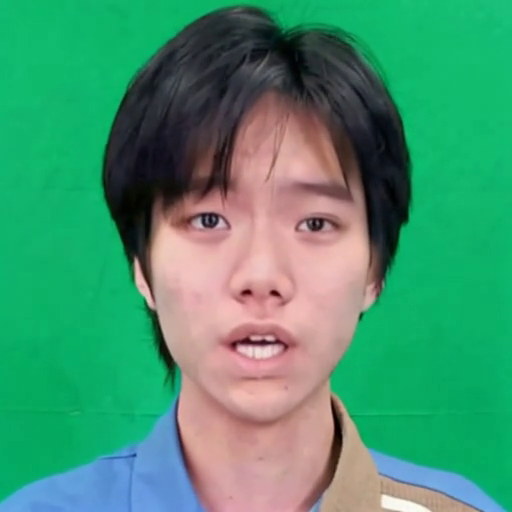}
    \end{subfigure}
    \hspace{-4pt}
        \begin{subfigure}{0.12\linewidth}
        \includegraphics[width=\linewidth]{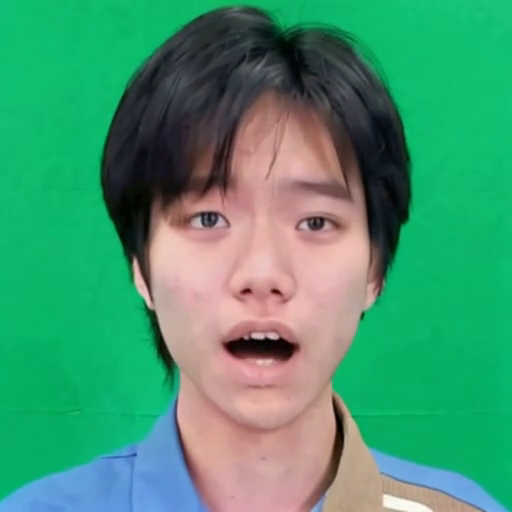}
    \end{subfigure}
    \hspace{-4pt}
        \begin{subfigure}{0.12\linewidth}
        \includegraphics[width=\linewidth]{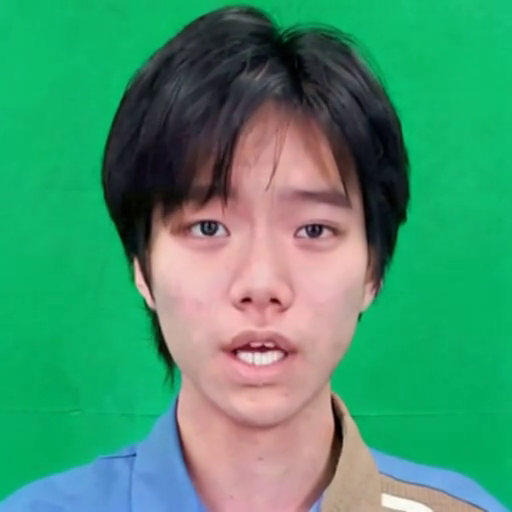}
    \end{subfigure}
    \end{minipage}
    \caption{The visualization of generated singing videos by baseline methods and our \model. }
    \label{fig:full_comprison_app_singing}
\end{figure*}

To provide a more reliable evaluation of the performance of different methods on the SingingHead dataset, we present a comprehensive comparison of the generated videos in Figure~\ref{fig:full_comprison_app_singing}. The results clearly demonstrate that our method, \model, generates more realistic and accurate singing videos, particularly in terms of lip synchronization and motion diversity. Compared to the baseline methods, our approach captures finer details in the facial expressions and head movements, ensuring more natural and expressive video generation. Notably, while the baseline methods struggle with maintaining consistent lip synchronization and accurate head poses, our method consistently produces high-quality results, even in challenging scenarios. This demonstrates the effectiveness of our design, particularly the Multi-scale Spectral Module and Self-adaptive Filter Module, which enhance the alignment between the audio and visual features for a more vivid and coherent performance.

\begin{figure*}[h]
\begin{subfigure}{0.24\linewidth}
    \includegraphics[width=\linewidth]{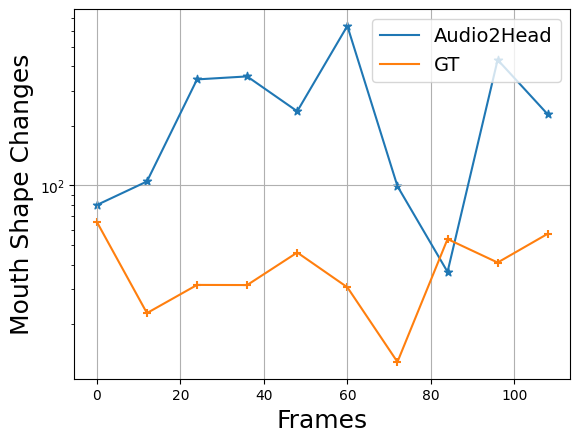}
    \caption{Audio2Head.}
\end{subfigure}
\begin{subfigure}{0.24\linewidth}
    \includegraphics[width=\linewidth]{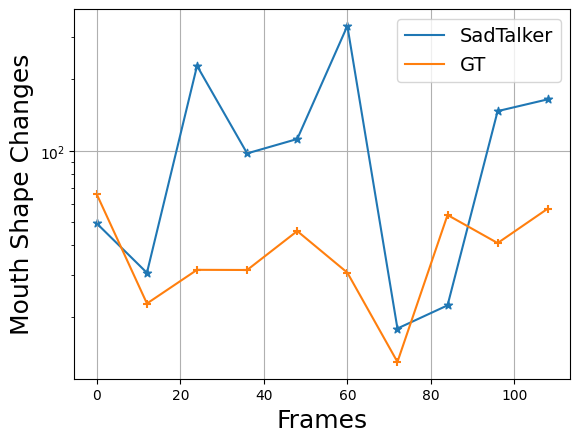}
    \caption{SadTalker.}
\end{subfigure}
\begin{subfigure}{0.24\linewidth}
    \includegraphics[width=\linewidth]{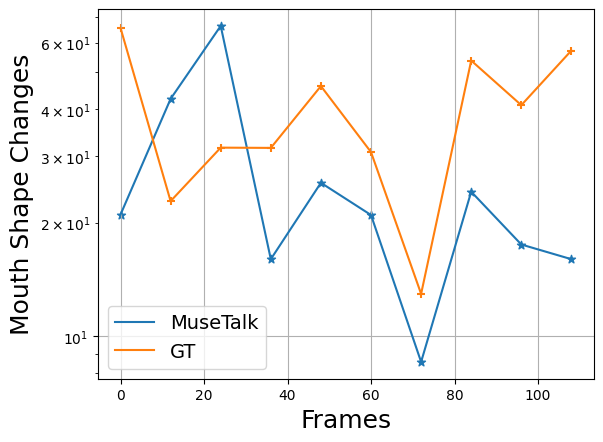}
    \caption{MuseTalk.}
\end{subfigure}
\begin{subfigure}{0.24\linewidth}
    \includegraphics[width=\linewidth]{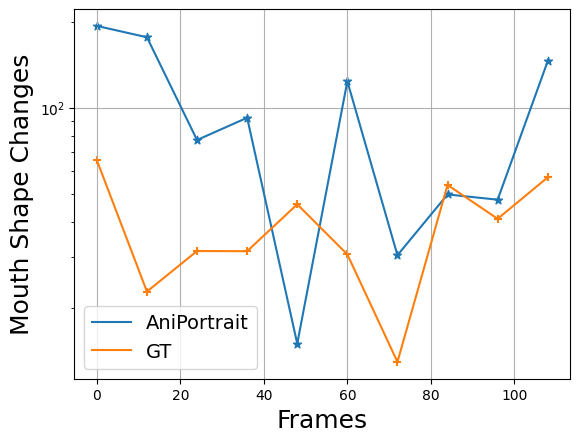}
    \caption{AniPortrait.}
\end{subfigure}

\begin{subfigure}{0.24\linewidth}
    \includegraphics[width=\linewidth]{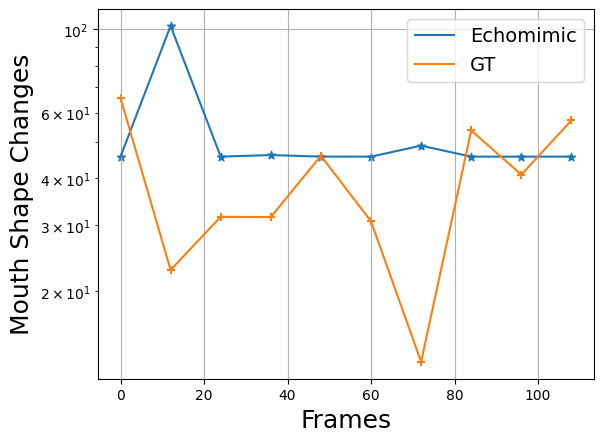}
    \caption{Echomimic.}
\end{subfigure}
\begin{subfigure}{0.24\linewidth}
    \includegraphics[width=\linewidth]{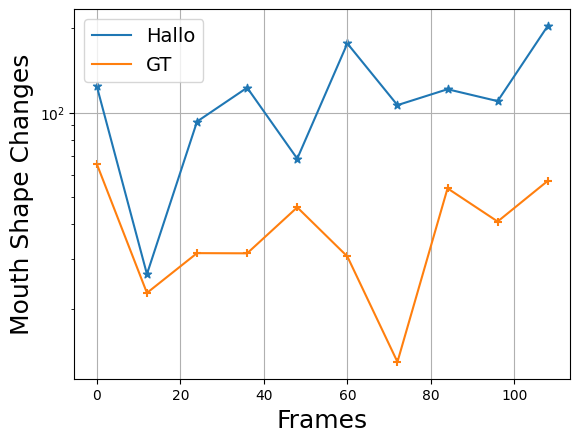}
    \caption{Hallo.}
\end{subfigure}
\begin{subfigure}{0.24\linewidth}
    \includegraphics[width=\linewidth]{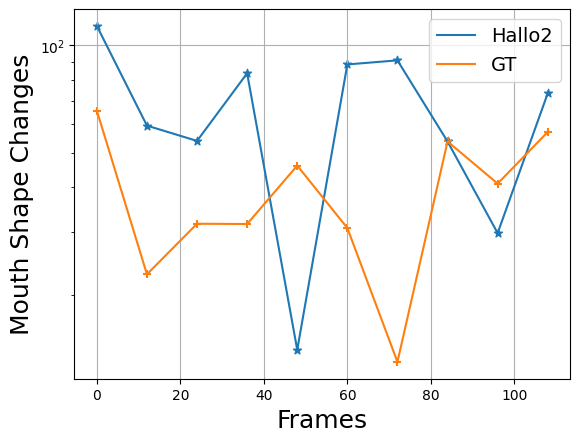}
    \caption{Hallo2.}
\end{subfigure}
\begin{subfigure}{0.24\linewidth}
    \includegraphics[width=\linewidth]{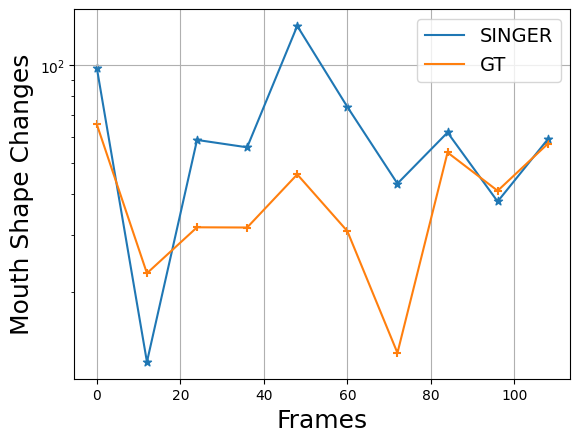}
    \caption{\model.}
\end{subfigure}

\caption{The mouth shape changes of videos generated by different methods as shown Figure~\ref{fig:full_comprison_app_singing}. }
\label{fig:app_mouth_shape_change_singing}
\end{figure*}

\begin{figure*}[h]
\begin{subfigure}{0.24\linewidth}
    \includegraphics[width=\linewidth]{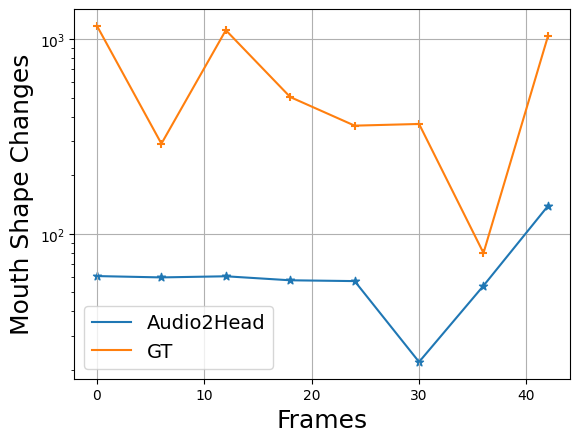}
    \caption{Audio2Head.}
\end{subfigure}
\begin{subfigure}{0.24\linewidth}
    \includegraphics[width=\linewidth]{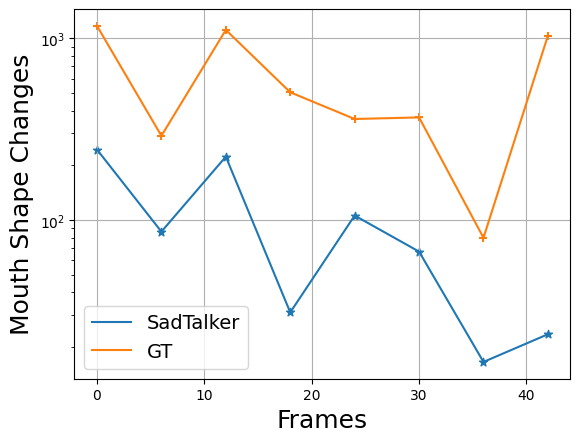}
    \caption{SadTalker.}
\end{subfigure}
\begin{subfigure}{0.24\linewidth}
    \includegraphics[width=\linewidth]{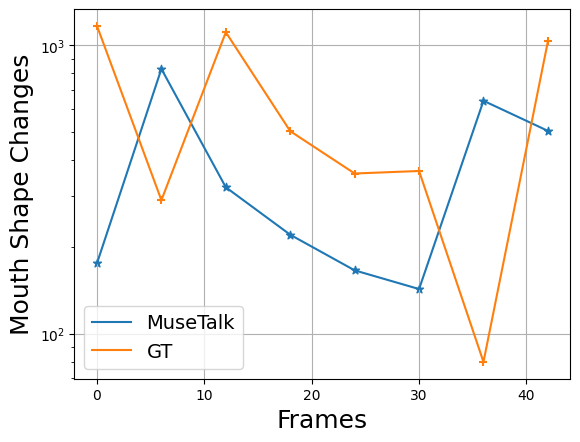}
    \caption{MuseTalk.}
\end{subfigure}
\begin{subfigure}{0.24\linewidth}
    \includegraphics[width=\linewidth]{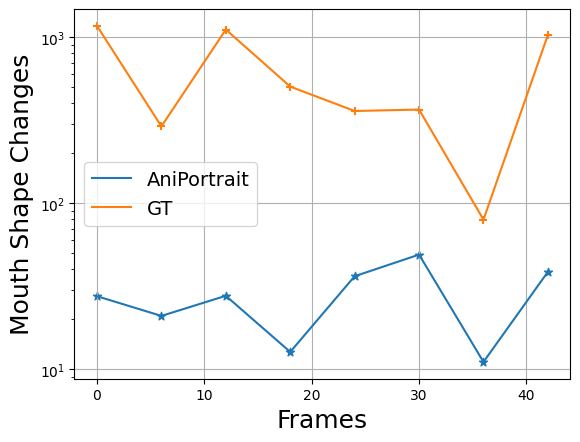}
    \caption{AniPortrait.}
\end{subfigure}

\begin{subfigure}{0.24\linewidth}
    \includegraphics[width=\linewidth]{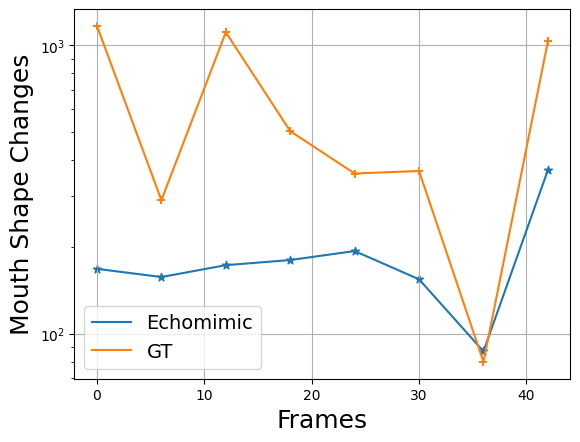}
    \caption{Echomimic.}
\end{subfigure}
\begin{subfigure}{0.24\linewidth}
    \includegraphics[width=\linewidth]{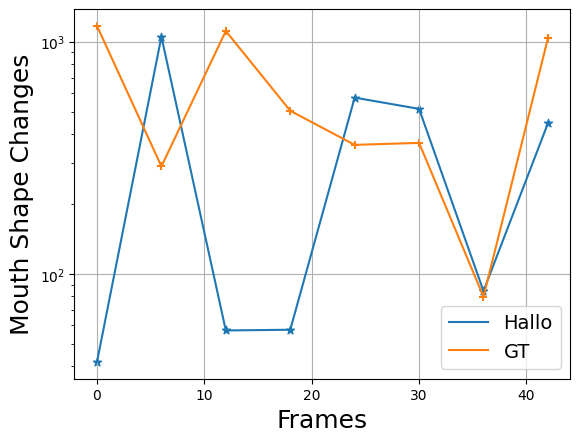}
    \caption{Hallo.}
\end{subfigure}
\begin{subfigure}{0.24\linewidth}
    \includegraphics[width=\linewidth]{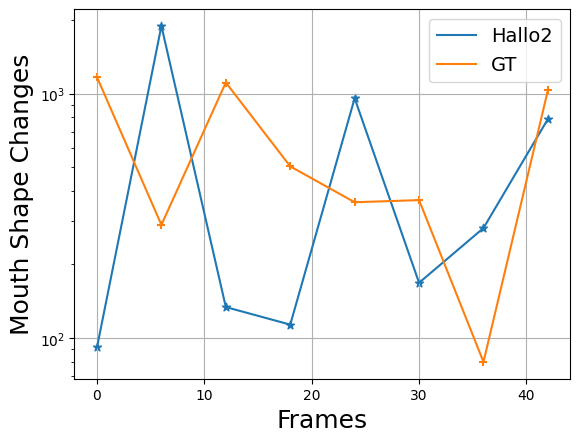}
    \caption{Hallo2.}
\end{subfigure}
\begin{subfigure}{0.24\linewidth}
    \includegraphics[width=\linewidth]{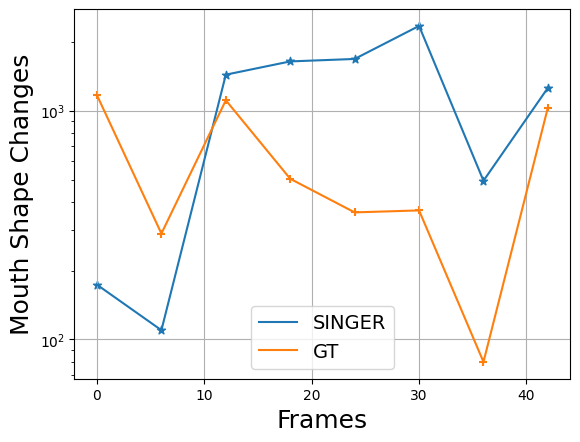}
    \caption{\model.}
\end{subfigure}

\caption{The mouth shape changes of videos generated by different methods as shown Figure~\ref{fig:full_comprison_app}. }
\label{fig:app_mouth_shape_change}
\end{figure*}

\subsubsection{Mouth Shape Changes} 
To further demonstrate the effectiveness of our method, we also present the mouth shape changes of the generated videos, as shown in Figure~\ref{fig:app_mouth_shape_change_singing}. The results indicate that our method produces more accurate and natural mouth shape changes compared to the baseline methods. The generated mouth movements closely align with the singing audio, ensuring better lip synchronization and enhancing the realism of the generated videos. This improvement highlights the advantage of our approach in capturing subtle variations in mouth shape during singing, which is crucial for achieving high-quality singing video generation.

\subsection{Full Comparison on SHV} ~\label{app:fullcomp}
\subsubsection{The mouth shape change}
We also present the mouth shape changes of the generated videos in Figure~\ref{fig:full_comprison_app} by different generation methods, as shown in Figure~\ref{fig:app_mouth_shape_change}. The results reveal that the baseline methods fail to generate accurate mouth shapes based on the given audio, whereas our method closely follows the same change trend as the ground truth video. This demonstrates the effectiveness of our proposed approach in achieving realistic and accurate lip synchronization, further highlighting the superior performance of our method in generating high-quality singing videos.

\subsection{Visualization}
\begin{figure*}[h]
    \begin{minipage}{0.02\linewidth}
    \centering
        \rotatebox{90}{GT}
    \end{minipage}
    \begin{minipage}{0.97\linewidth}
    \begin{subfigure}{0.12\linewidth}
        \includegraphics[width=\linewidth]{figures/gt/0001.png}
    \end{subfigure}
    \hspace{-4pt}
        \begin{subfigure}{0.12\linewidth}
        \includegraphics[width=\linewidth]{figures/gt/0007.png}
    \end{subfigure}
     \hspace{-4pt}
        \begin{subfigure}{0.12\linewidth}
        \includegraphics[width=\linewidth]{figures/gt/0010.png}
    \end{subfigure}
     \hspace{-4pt}
        \begin{subfigure}{0.12\linewidth}
        \includegraphics[width=\linewidth]{figures/gt/0022.png}
    \end{subfigure}
     \hspace{-4pt}
        \begin{subfigure}{0.12\linewidth}
        \includegraphics[width=\linewidth]{figures/gt/0031.png}
    \end{subfigure}
     \hspace{-4pt}
        \begin{subfigure}{0.12\linewidth}
        \includegraphics[width=\linewidth]{figures/gt/0036.png}
    \end{subfigure}
    \hspace{-4pt}
        \begin{subfigure}{0.12\linewidth}
        \includegraphics[width=\linewidth]{figures/gt/0038.png}
    \end{subfigure}
    \hspace{-4pt}
        \begin{subfigure}{0.12\linewidth}
        \includegraphics[width=\linewidth]{figures/gt/0048.png}
    \end{subfigure}
    \end{minipage}
  
    \centering
        \begin{minipage}{0.02\linewidth}
    \centering
        \rotatebox{90}{Audio2Head}
    \end{minipage}
    \begin{minipage}{0.97\linewidth}
    \begin{subfigure}{0.12\linewidth}
        \includegraphics[width=\linewidth]{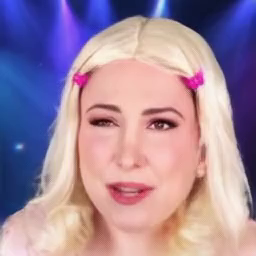}
    \end{subfigure}
    \hspace{-4pt}
        \begin{subfigure}{0.12\linewidth}
        \includegraphics[width=\linewidth]{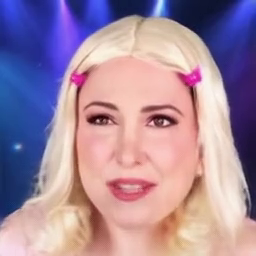}
    \end{subfigure}
     \hspace{-4pt}
        \begin{subfigure}{0.12\linewidth}
        \includegraphics[width=\linewidth]{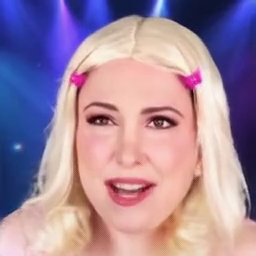}
    \end{subfigure}
     \hspace{-4pt}
        \begin{subfigure}{0.12\linewidth}
        \includegraphics[width=\linewidth]{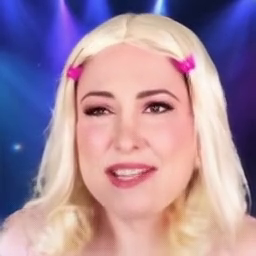}
    \end{subfigure}
     \hspace{-4pt}
        \begin{subfigure}{0.12\linewidth}
        \includegraphics[width=\linewidth]{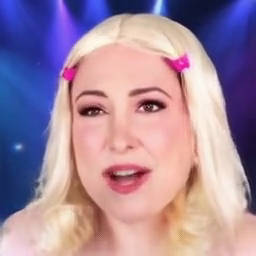}
    \end{subfigure}
     \hspace{-4pt}
        \begin{subfigure}{0.12\linewidth}
        \includegraphics[width=\linewidth]{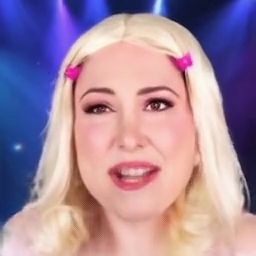}
    \end{subfigure}
    \hspace{-4pt}
        \begin{subfigure}{0.12\linewidth}
        \includegraphics[width=\linewidth]{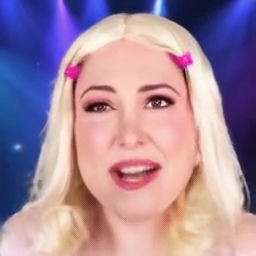}
    \end{subfigure}
    \hspace{-4pt}
        \begin{subfigure}{0.12\linewidth}
        \includegraphics[width=\linewidth]{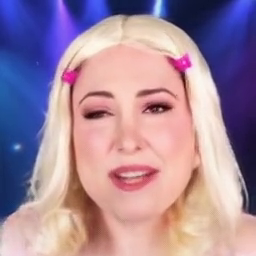}
    \end{subfigure}
    \end{minipage}
    \begin{minipage}{0.02\linewidth}
    \centering
        \rotatebox{90}{SadTalker}
    \end{minipage}
    \begin{minipage}{0.97\linewidth}
    \begin{subfigure}{0.12\linewidth}
        \includegraphics[width=\linewidth]{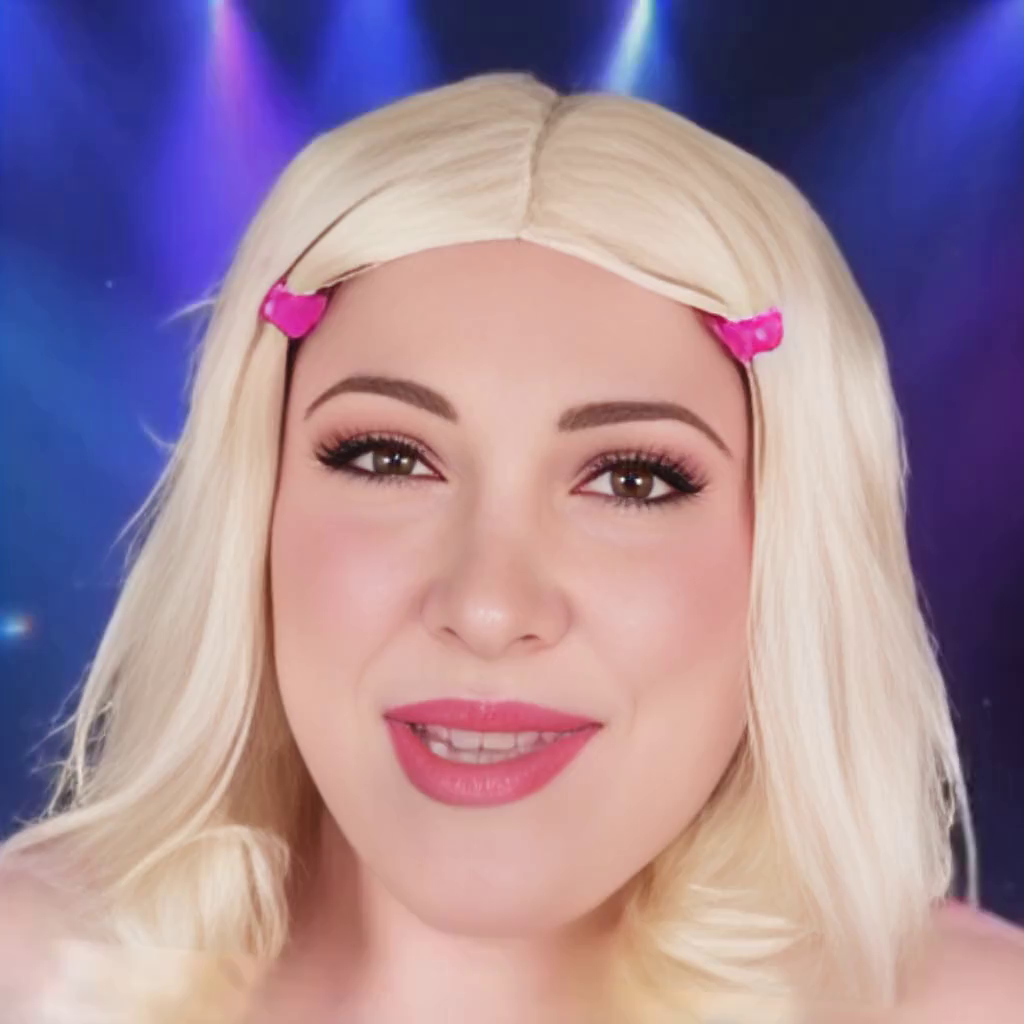}
    \end{subfigure}
    \hspace{-4pt}
        \begin{subfigure}{0.12\linewidth}
        \includegraphics[width=\linewidth]{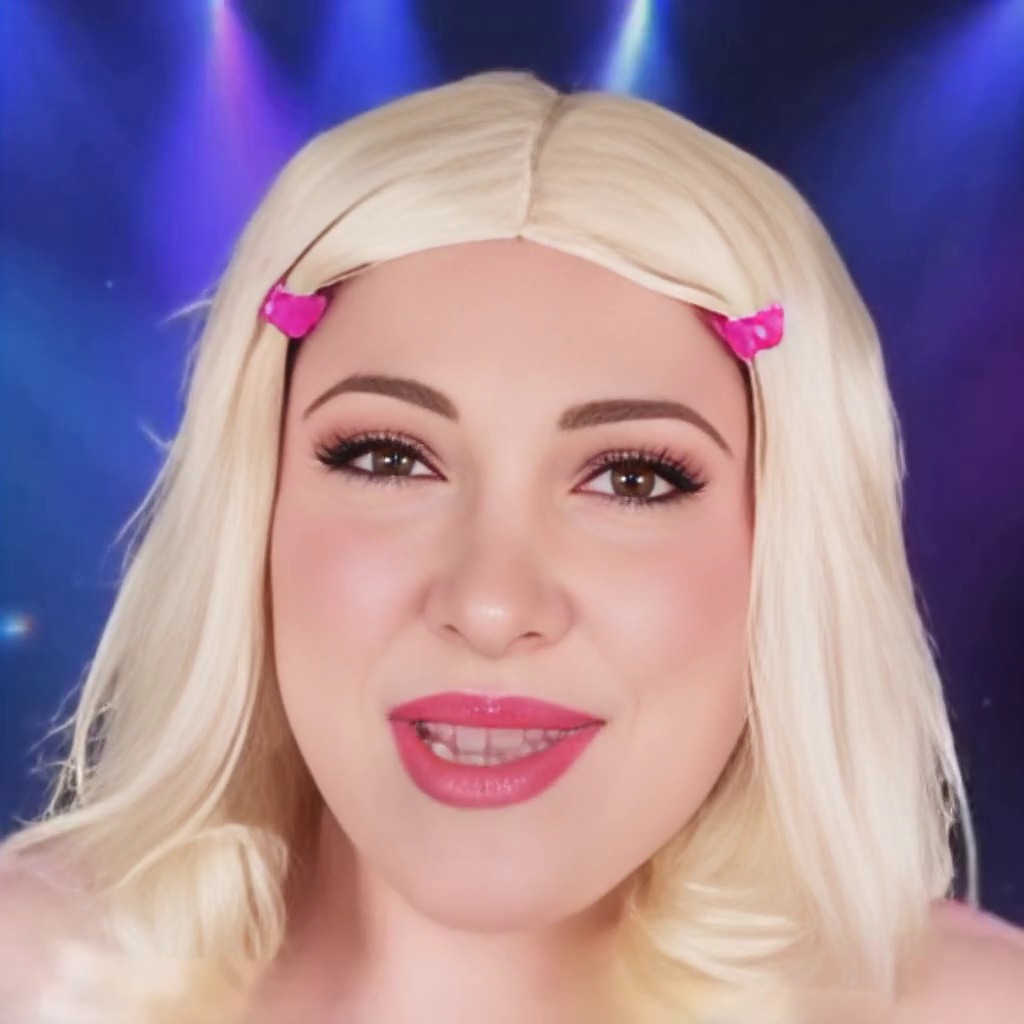}
    \end{subfigure}
     \hspace{-4pt}
        \begin{subfigure}{0.12\linewidth}
        \includegraphics[width=\linewidth]{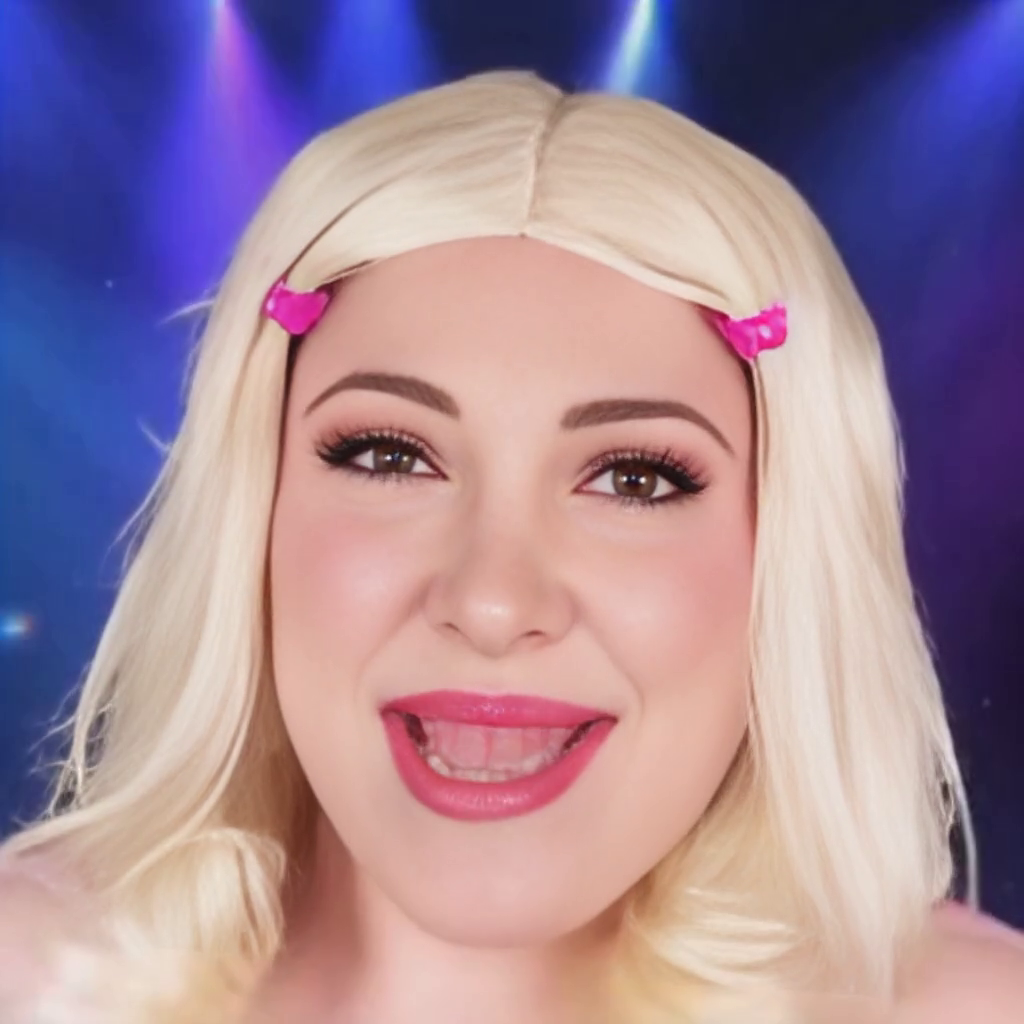}
    \end{subfigure}
     \hspace{-4pt}
        \begin{subfigure}{0.12\linewidth}
        \includegraphics[width=\linewidth]{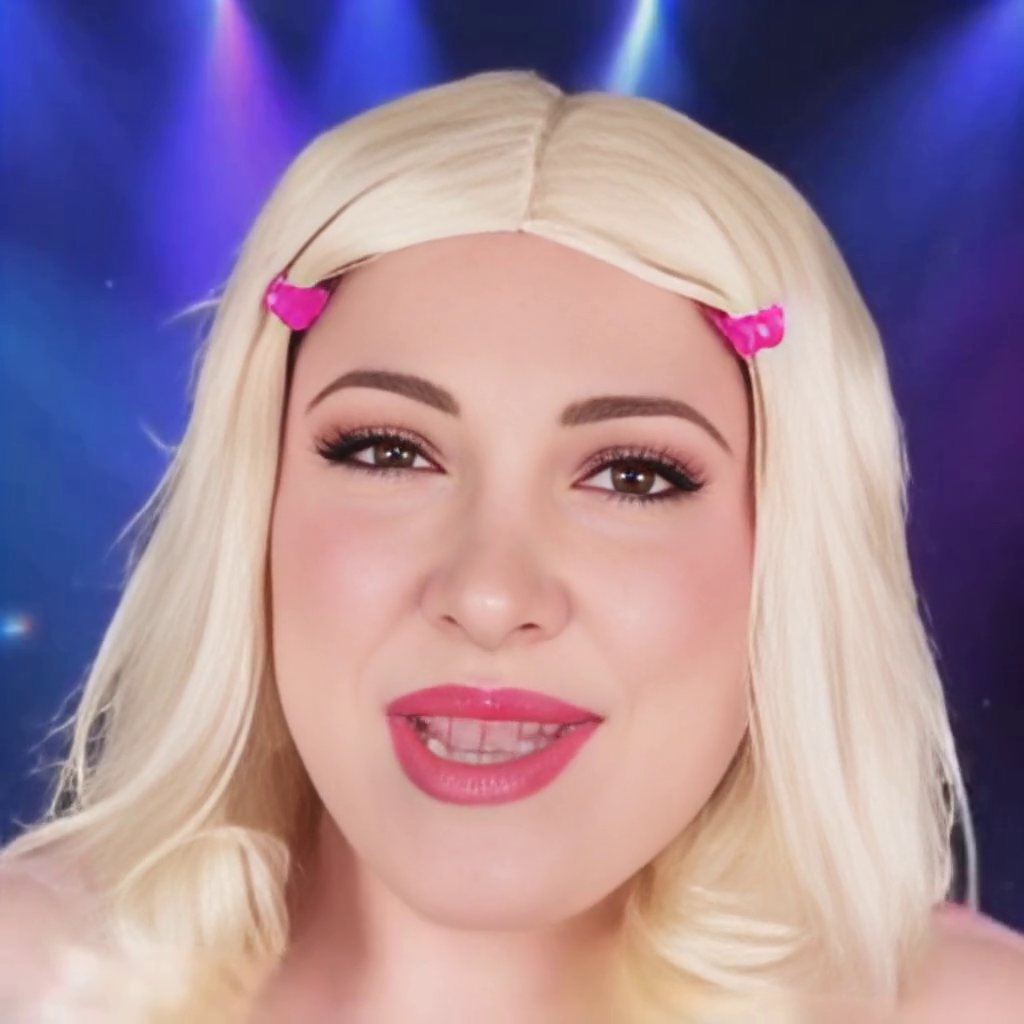}
    \end{subfigure}
     \hspace{-4pt}
        \begin{subfigure}{0.12\linewidth}
        \includegraphics[width=\linewidth]{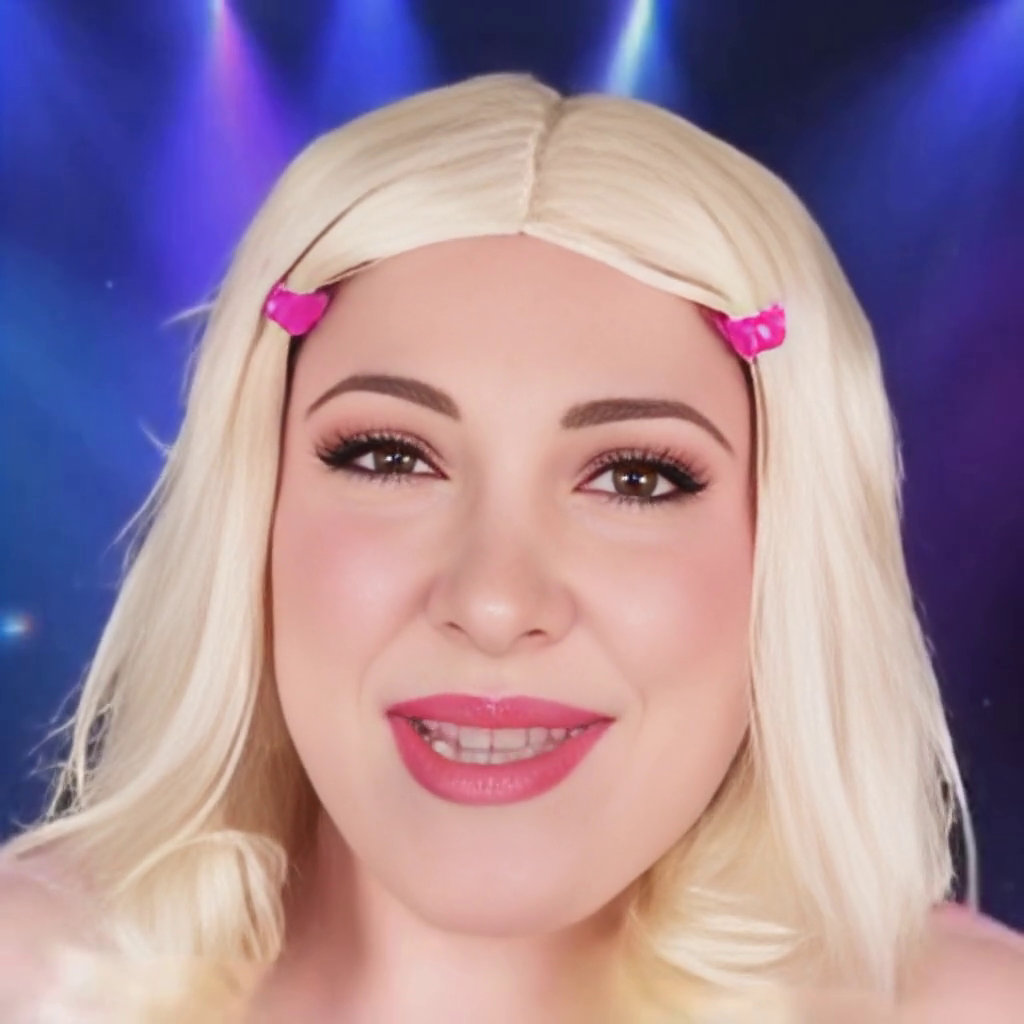}
    \end{subfigure}
     \hspace{-4pt}
        \begin{subfigure}{0.12\linewidth}
        \includegraphics[width=\linewidth]{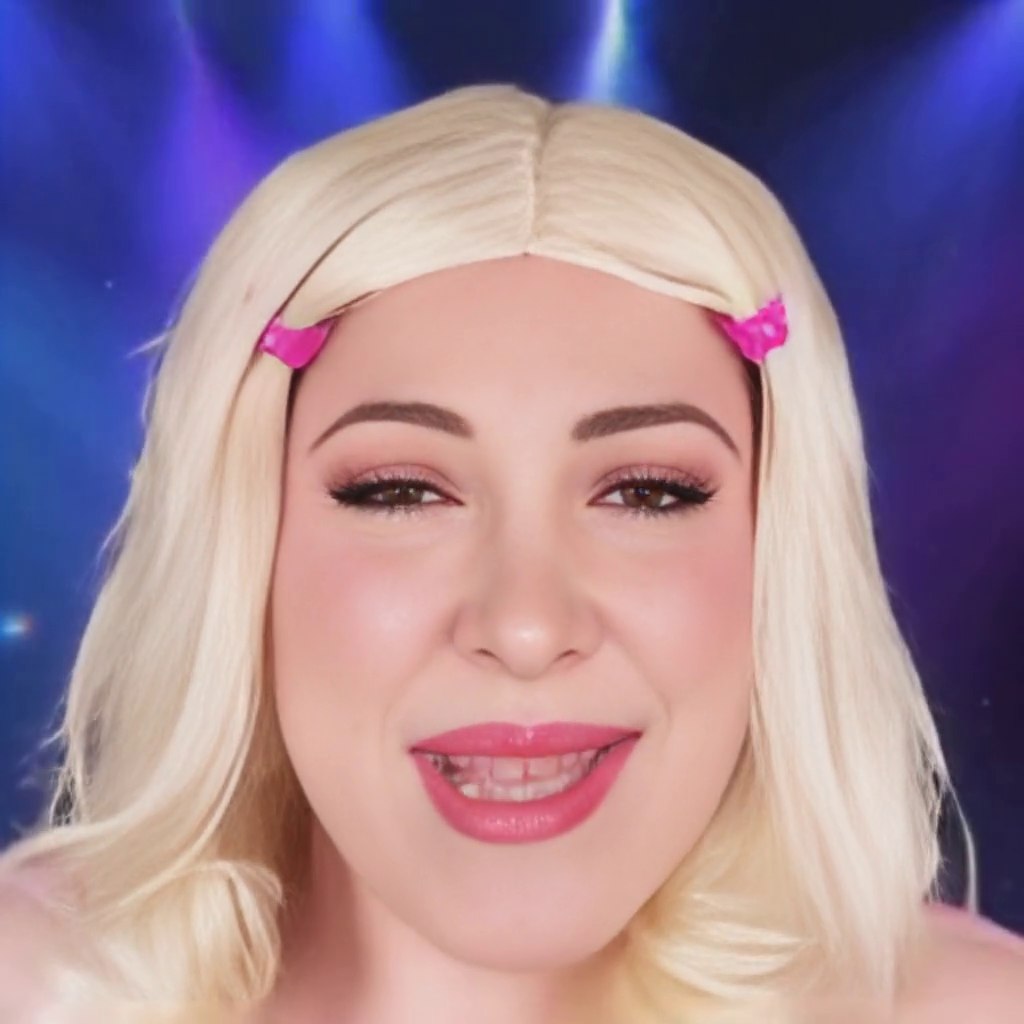}
    \end{subfigure}
    \hspace{-4pt}
        \begin{subfigure}{0.12\linewidth}
        \includegraphics[width=\linewidth]{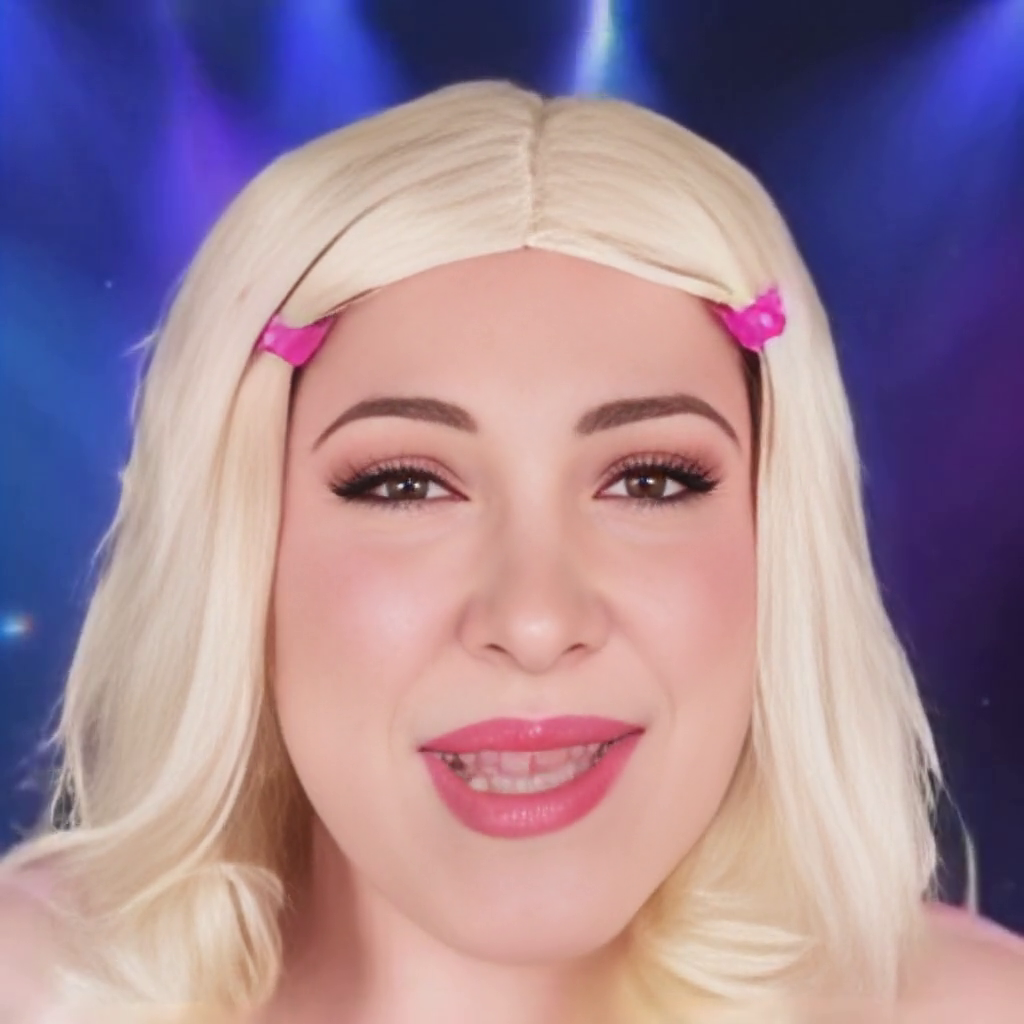}
    \end{subfigure}
    \hspace{-4pt}
        \begin{subfigure}{0.12\linewidth}
        \includegraphics[width=\linewidth]{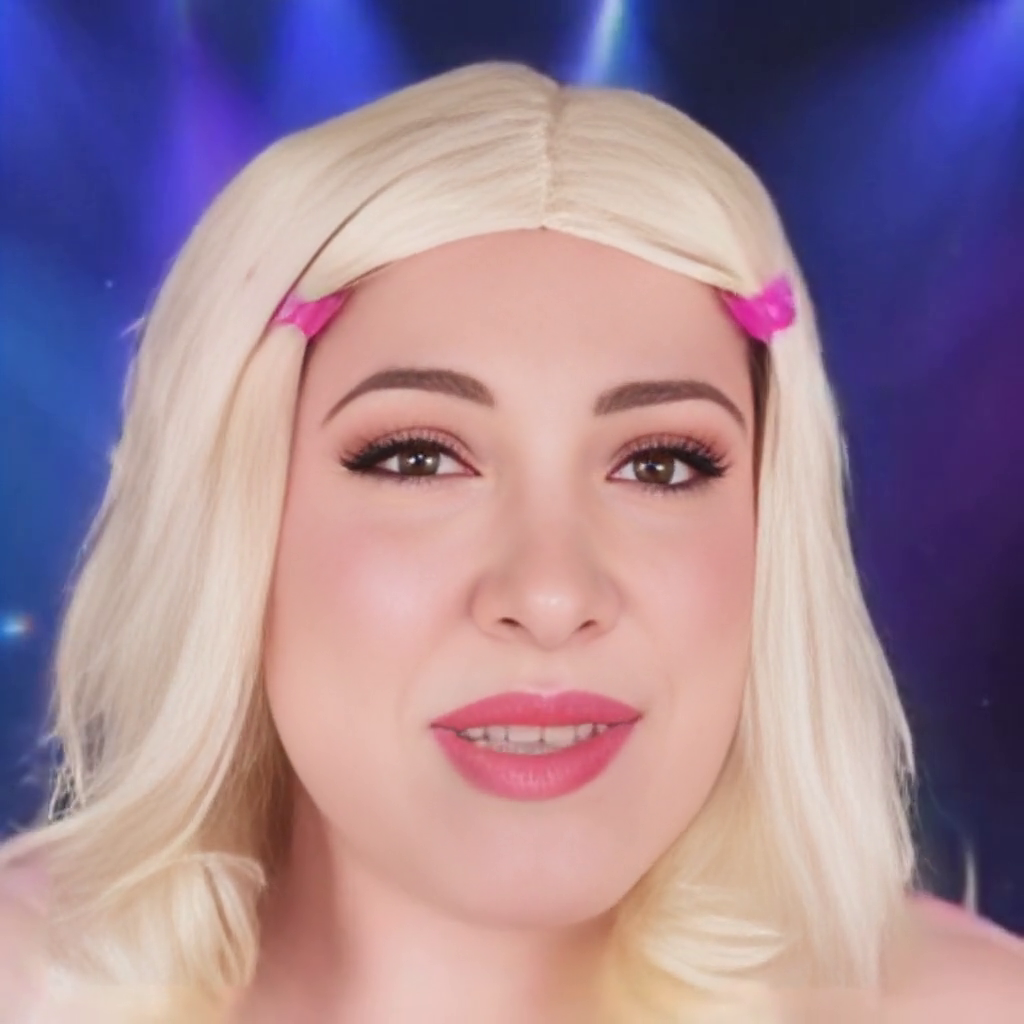}
    \end{subfigure}
    \end{minipage}
      \begin{minipage}{0.02\linewidth}
    \centering
        \rotatebox{90}{MuseTalk}
    \end{minipage}
    \begin{minipage}{0.97\linewidth}
    \begin{subfigure}{0.12\linewidth}
        \includegraphics[width=\linewidth]{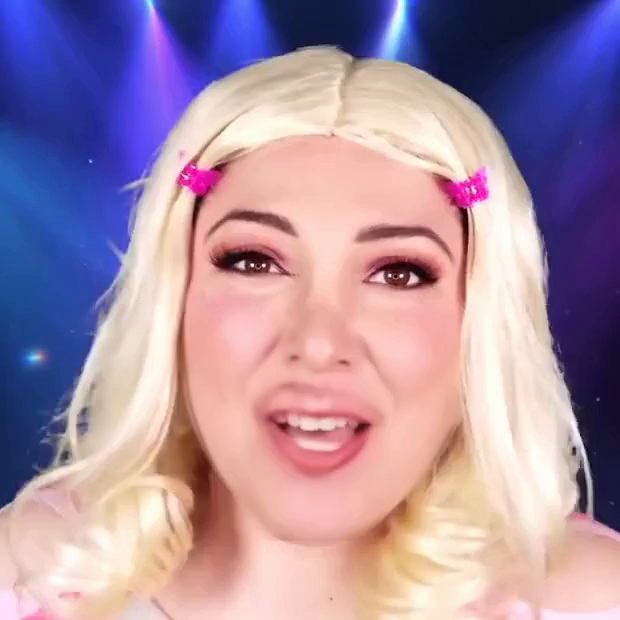}
    \end{subfigure}
    \hspace{-4pt}
        \begin{subfigure}{0.12\linewidth}
        \includegraphics[width=\linewidth]{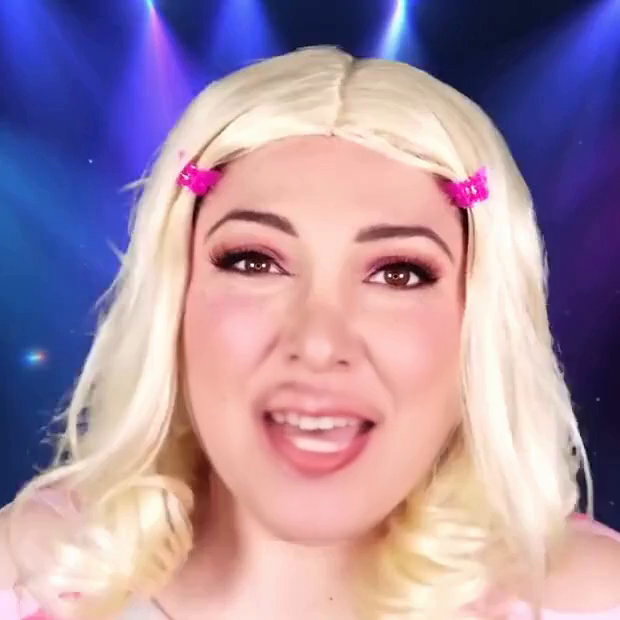}
    \end{subfigure}
     \hspace{-4pt}
        \begin{subfigure}{0.12\linewidth}
        \includegraphics[width=\linewidth]{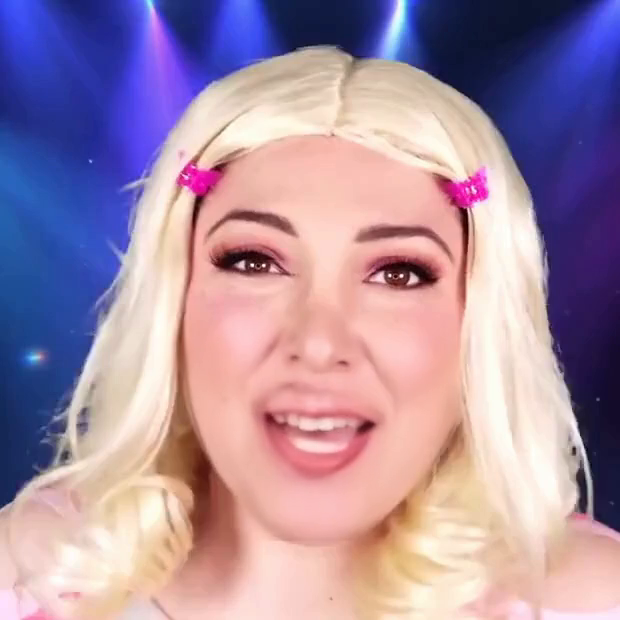}
    \end{subfigure}
     \hspace{-4pt}
        \begin{subfigure}{0.12\linewidth}
        \includegraphics[width=\linewidth]{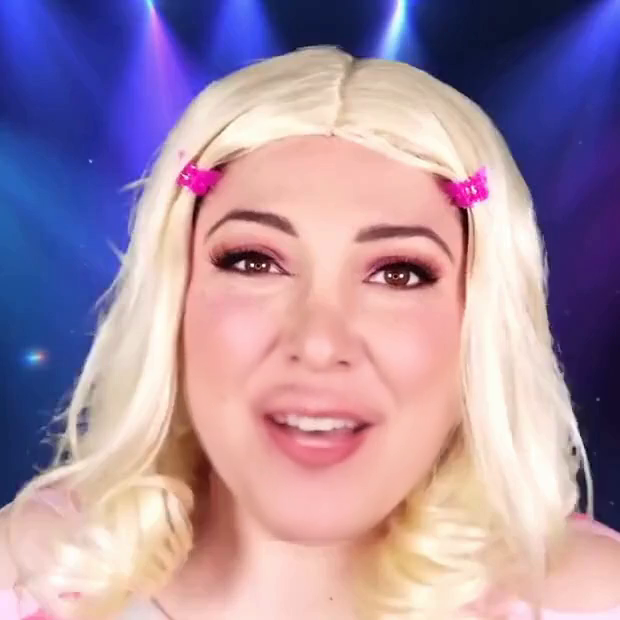}
    \end{subfigure}
     \hspace{-4pt}
        \begin{subfigure}{0.12\linewidth}
        \includegraphics[width=\linewidth]{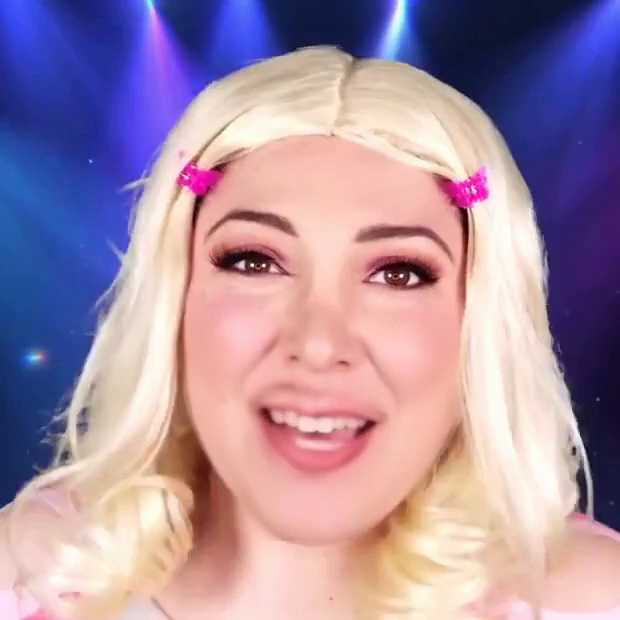}
    \end{subfigure}
     \hspace{-4pt}
        \begin{subfigure}{0.12\linewidth}
        \includegraphics[width=\linewidth]{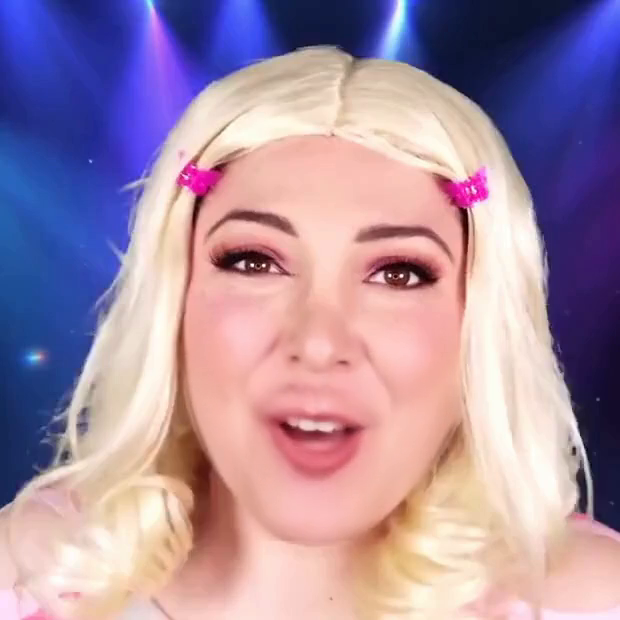}
    \end{subfigure}
    \hspace{-4pt}
        \begin{subfigure}{0.12\linewidth}
        \includegraphics[width=\linewidth]{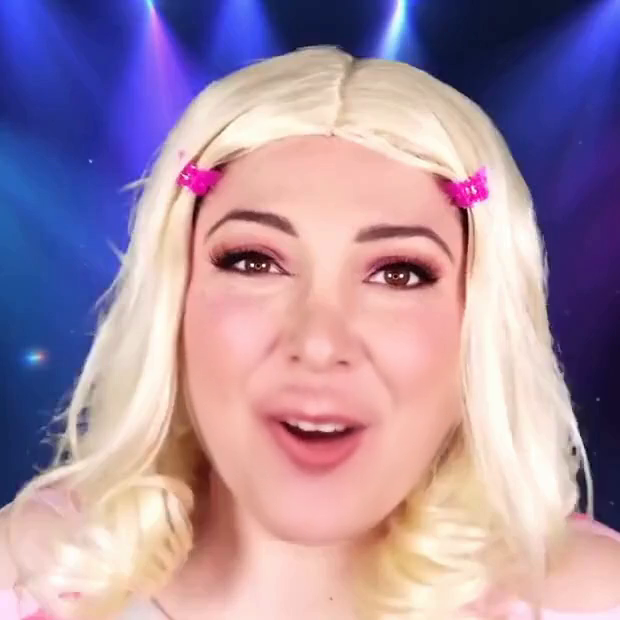}
    \end{subfigure}
    \hspace{-4pt}
        \begin{subfigure}{0.12\linewidth}
        \includegraphics[width=\linewidth]{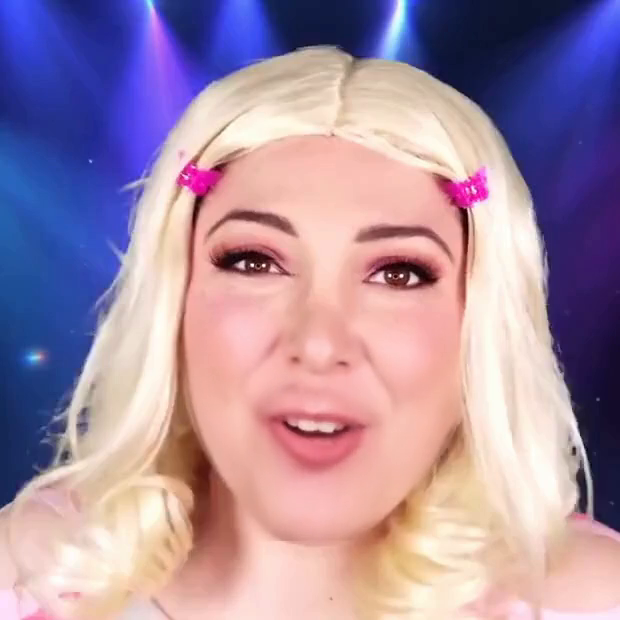}
    \end{subfigure}
    \end{minipage}
      \begin{minipage}{0.02\linewidth}
    \centering
        \rotatebox{90}{AniPortrait}
    \end{minipage}
    \begin{minipage}{0.97\linewidth}
    \begin{subfigure}{0.12\linewidth}
        \includegraphics[width=\linewidth]{figures/aniport/2bCItaHq3JQ-008-2_0001.png}
    \end{subfigure}
    \hspace{-4pt}
        \begin{subfigure}{0.12\linewidth}
        \includegraphics[width=\linewidth]{figures/aniport/2bCItaHq3JQ-008-2_0007.png}
    \end{subfigure}
     \hspace{-4pt}
        \begin{subfigure}{0.12\linewidth}
        \includegraphics[width=\linewidth]{figures/aniport/2bCItaHq3JQ-008-2_0010.png}
    \end{subfigure}
     \hspace{-4pt}
        \begin{subfigure}{0.12\linewidth}
        \includegraphics[width=\linewidth]{figures/aniport/2bCItaHq3JQ-008-2_0022.png}
    \end{subfigure}
     \hspace{-4pt}
        \begin{subfigure}{0.12\linewidth}
        \includegraphics[width=\linewidth]{figures/aniport/2bCItaHq3JQ-008-2_0031.png}
    \end{subfigure}
     \hspace{-4pt}
        \begin{subfigure}{0.12\linewidth}
        \includegraphics[width=\linewidth]{figures/aniport/2bCItaHq3JQ-008-2_0036.png}
    \end{subfigure}
    \hspace{-4pt}
        \begin{subfigure}{0.12\linewidth}
        \includegraphics[width=\linewidth]{figures/aniport/2bCItaHq3JQ-008-2_0038.png}
    \end{subfigure}
    \hspace{-4pt}
        \begin{subfigure}{0.12\linewidth}
        \includegraphics[width=\linewidth]{figures/aniport/2bCItaHq3JQ-008-2_0048.png}
    \end{subfigure}
    \end{minipage}
      \begin{minipage}{0.02\linewidth}
    \centering
        \rotatebox{90}{Echomimic}
    \end{minipage}
    \begin{minipage}{0.97\linewidth}
    \begin{subfigure}{0.12\linewidth}
        \includegraphics[width=\linewidth]{figures/echomimic/2bCItaHq3JQ-008-2_0001.png}
    \end{subfigure}
    \hspace{-4pt}
        \begin{subfigure}{0.12\linewidth}
        \includegraphics[width=\linewidth]{figures/echomimic/2bCItaHq3JQ-008-2_0007.png}
    \end{subfigure}
     \hspace{-4pt}
        \begin{subfigure}{0.12\linewidth}
        \includegraphics[width=\linewidth]{figures/echomimic/2bCItaHq3JQ-008-2_0010.png}
    \end{subfigure}
     \hspace{-4pt}
        \begin{subfigure}{0.12\linewidth}
        \includegraphics[width=\linewidth]{figures/echomimic/2bCItaHq3JQ-008-2_0022.png}
    \end{subfigure}
     \hspace{-4pt}
        \begin{subfigure}{0.12\linewidth}
        \includegraphics[width=\linewidth]{figures/echomimic/2bCItaHq3JQ-008-2_0031.png}
    \end{subfigure}
     \hspace{-4pt}
        \begin{subfigure}{0.12\linewidth}
        \includegraphics[width=\linewidth]{figures/echomimic/2bCItaHq3JQ-008-2_0036.png}
    \end{subfigure}
    \hspace{-4pt}
        \begin{subfigure}{0.12\linewidth}
        \includegraphics[width=\linewidth]{figures/echomimic/2bCItaHq3JQ-008-2_0038.png}
    \end{subfigure}
    \hspace{-4pt}
        \begin{subfigure}{0.12\linewidth}
        \includegraphics[width=\linewidth]{figures/echomimic/2bCItaHq3JQ-008-2_0048.png}
    \end{subfigure}
    \end{minipage}
      \begin{minipage}{0.02\linewidth}
    \centering
        \rotatebox{90}{Hallo}
    \end{minipage}
    \begin{minipage}{0.97\linewidth}
    \begin{subfigure}{0.12\linewidth}
        \includegraphics[width=\linewidth]{figures/hallo/2bCItaHq3JQ-008-2_0001.png}
    \end{subfigure}
    \hspace{-4pt}
        \begin{subfigure}{0.12\linewidth}
        \includegraphics[width=\linewidth]{figures/hallo/2bCItaHq3JQ-008-2_0007.png}
    \end{subfigure}
     \hspace{-4pt}
        \begin{subfigure}{0.12\linewidth}
        \includegraphics[width=\linewidth]{figures/hallo/2bCItaHq3JQ-008-2_0010.png}
    \end{subfigure}
     \hspace{-4pt}
        \begin{subfigure}{0.12\linewidth}
        \includegraphics[width=\linewidth]{figures/hallo/2bCItaHq3JQ-008-2_0022.png}
    \end{subfigure}
     \hspace{-4pt}
        \begin{subfigure}{0.12\linewidth}
        \includegraphics[width=\linewidth]{figures/hallo/2bCItaHq3JQ-008-2_0031.png}
    \end{subfigure}
     \hspace{-4pt}
        \begin{subfigure}{0.12\linewidth}
        \includegraphics[width=\linewidth]{figures/hallo/2bCItaHq3JQ-008-2_0036.png}
    \end{subfigure}
    \hspace{-4pt}
        \begin{subfigure}{0.12\linewidth}
        \includegraphics[width=\linewidth]{figures/hallo/2bCItaHq3JQ-008-2_0038.png}
    \end{subfigure}
    \hspace{-4pt}
        \begin{subfigure}{0.12\linewidth}
        \includegraphics[width=\linewidth]{figures/hallo/2bCItaHq3JQ-008-2_0048.png}
    \end{subfigure}
    \end{minipage}
      \begin{minipage}{0.02\linewidth}
    \centering
        \rotatebox{90}{Hallo2}
    \end{minipage}
    \begin{minipage}{0.97\linewidth}
    \begin{subfigure}{0.12\linewidth}
        \includegraphics[width=\linewidth]{figures/hallo2/2bCItaHq3JQ-008-2_0001.png}
    \end{subfigure}
    \hspace{-4pt}
        \begin{subfigure}{0.12\linewidth}
        \includegraphics[width=\linewidth]{figures/hallo2/2bCItaHq3JQ-008-2_0007.png}
    \end{subfigure}
     \hspace{-4pt}
        \begin{subfigure}{0.12\linewidth}
        \includegraphics[width=\linewidth]{figures/hallo2/2bCItaHq3JQ-008-2_0010.png}
    \end{subfigure}
     \hspace{-4pt}
        \begin{subfigure}{0.12\linewidth}
        \includegraphics[width=\linewidth]{figures/hallo2/2bCItaHq3JQ-008-2_0022.png}
    \end{subfigure}
     \hspace{-4pt}
        \begin{subfigure}{0.12\linewidth}
        \includegraphics[width=\linewidth]{figures/hallo2/2bCItaHq3JQ-008-2_0031.png}
    \end{subfigure}
     \hspace{-4pt}
        \begin{subfigure}{0.12\linewidth}
        \includegraphics[width=\linewidth]{figures/hallo2/2bCItaHq3JQ-008-2_0036.png}
    \end{subfigure}
    \hspace{-4pt}
        \begin{subfigure}{0.12\linewidth}
        \includegraphics[width=\linewidth]{figures/hallo2/2bCItaHq3JQ-008-2_0038.png}
    \end{subfigure}
    \hspace{-4pt}
        \begin{subfigure}{0.12\linewidth}
        \includegraphics[width=\linewidth]{figures/hallo2/2bCItaHq3JQ-008-2_0048.png}
    \end{subfigure}
    \end{minipage}
  
    \begin{minipage}{0.02\linewidth}
    \centering
        \rotatebox{90}{\model}
    \end{minipage}
    \begin{minipage}{0.97\linewidth}
    \begin{subfigure}{0.12\linewidth}
        \includegraphics[width=\linewidth]{figures/our/2bCItaHq3JQ-008-2_0001.png}
    \end{subfigure}
    \hspace{-4pt}
        \begin{subfigure}{0.12\linewidth}
        \includegraphics[width=\linewidth]{figures/our/2bCItaHq3JQ-008-2_0007.png}
    \end{subfigure}
     \hspace{-4pt}
        \begin{subfigure}{0.12\linewidth}
        \includegraphics[width=\linewidth]{figures/our/2bCItaHq3JQ-008-2_0010.png}
    \end{subfigure}
     \hspace{-4pt}
        \begin{subfigure}{0.12\linewidth}
        \includegraphics[width=\linewidth]{figures/our/2bCItaHq3JQ-008-2_0022.png}
    \end{subfigure}
     \hspace{-4pt}
        \begin{subfigure}{0.12\linewidth}
        \includegraphics[width=\linewidth]{figures/our/2bCItaHq3JQ-008-2_0031.png}
    \end{subfigure}
     \hspace{-4pt}
        \begin{subfigure}{0.12\linewidth}
        \includegraphics[width=\linewidth]{figures/our/2bCItaHq3JQ-008-2_0036.png}
    \end{subfigure}
    \hspace{-4pt}
        \begin{subfigure}{0.12\linewidth}
        \includegraphics[width=\linewidth]{figures/our/2bCItaHq3JQ-008-2_0038.png}
    \end{subfigure}
    \hspace{-4pt}
        \begin{subfigure}{0.12\linewidth}
        \includegraphics[width=\linewidth]{figures/our/2bCItaHq3JQ-008-2_0048.png}
    \end{subfigure}
    \end{minipage}
    \caption{The visualization of generated singing videos by baseline methods and our \model. }
    \label{fig:full_comprison_app}
\end{figure*}

\begin{figure*}[h]
    \begin{minipage}{0.02\linewidth}
    \centering
        \rotatebox{90}{GT}
    \end{minipage}
    \begin{minipage}{0.97\linewidth}
    \begin{subfigure}{0.12\linewidth}
        \includegraphics[width=\linewidth]{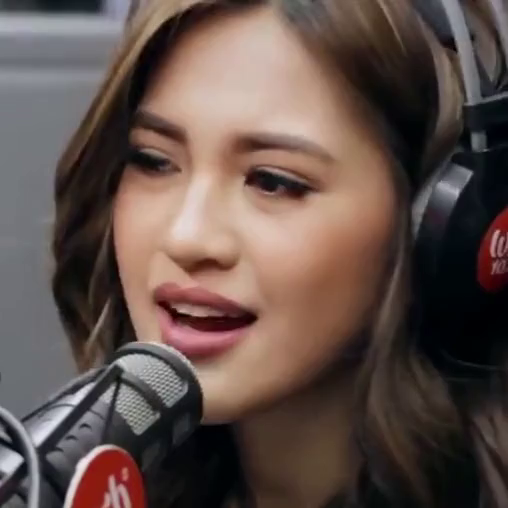}
    \end{subfigure}
    \hspace{-4pt}
        \begin{subfigure}{0.12\linewidth}
        \includegraphics[width=\linewidth]{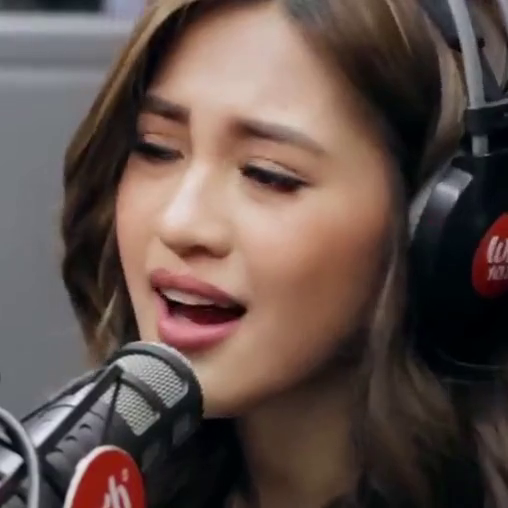}
    \end{subfigure}
     \hspace{-4pt}
        \begin{subfigure}{0.12\linewidth}
        \includegraphics[width=\linewidth]{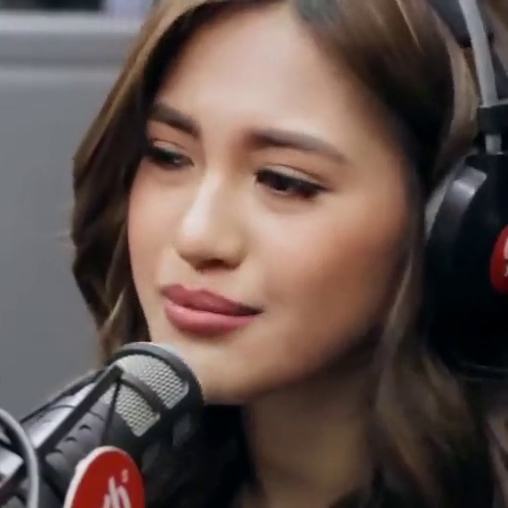}
    \end{subfigure}
     \hspace{-4pt}
        \begin{subfigure}{0.12\linewidth}
        \includegraphics[width=\linewidth]{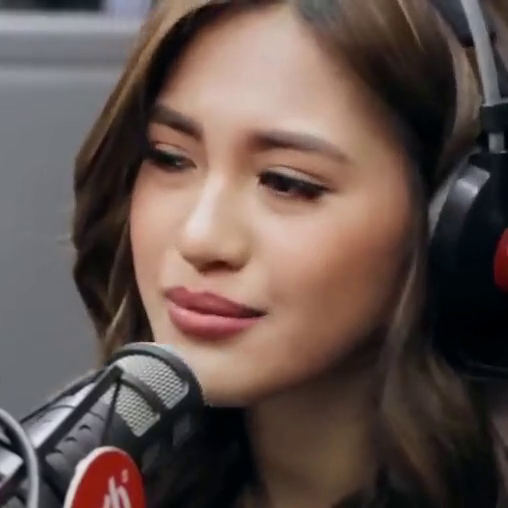}
    \end{subfigure}
     \hspace{-4pt}
        \begin{subfigure}{0.12\linewidth}
        \includegraphics[width=\linewidth]{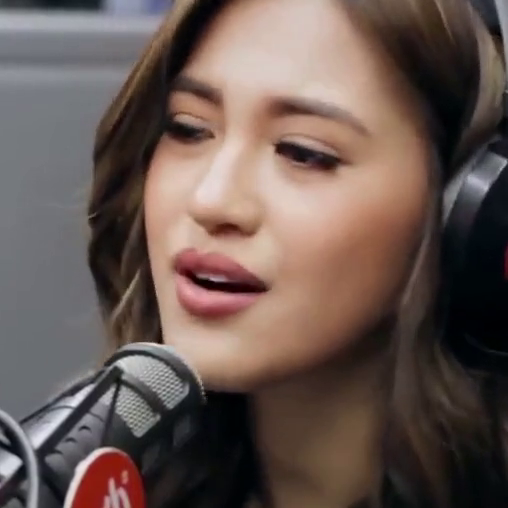}
    \end{subfigure}
     \hspace{-4pt}
        \begin{subfigure}{0.12\linewidth}
        \includegraphics[width=\linewidth]{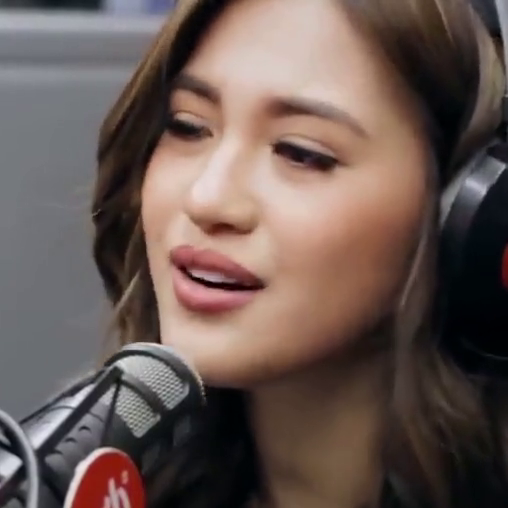}
    \end{subfigure}
    \hspace{-4pt}
        \begin{subfigure}{0.12\linewidth}
        \includegraphics[width=\linewidth]{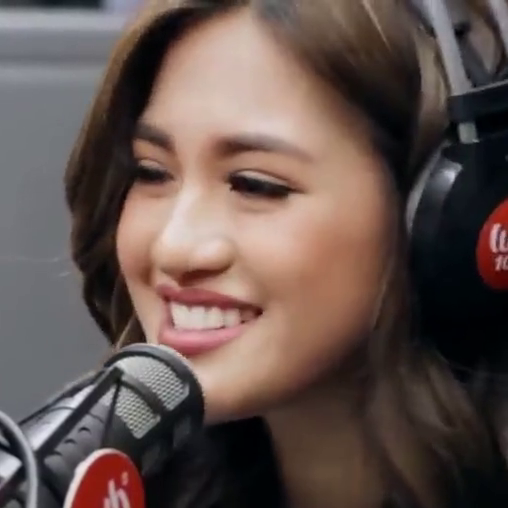}
    \end{subfigure}
    \hspace{-4pt}
        \begin{subfigure}{0.12\linewidth}
        \includegraphics[width=\linewidth]{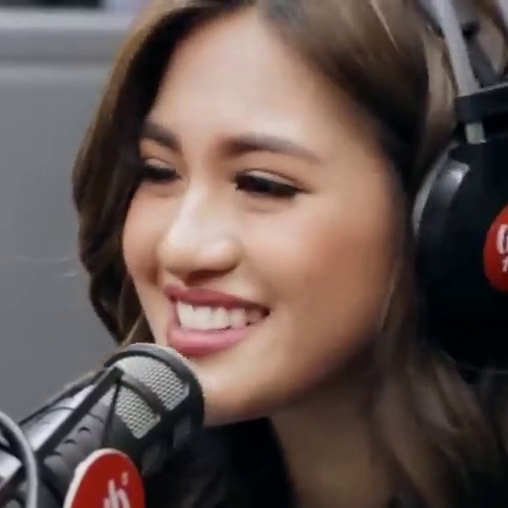}
    \end{subfigure}
    \end{minipage}
  
    \centering
        \begin{minipage}{0.02\linewidth}
    \centering
        \rotatebox{90}{Audio2Head}
    \end{minipage}
    \begin{minipage}{0.97\linewidth}
    \begin{subfigure}{0.12\linewidth}
        \includegraphics[width=\linewidth]{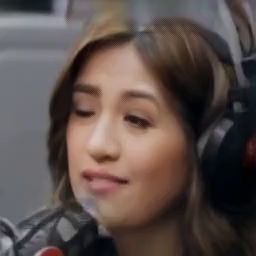}
    \end{subfigure}
    \hspace{-4pt}
        \begin{subfigure}{0.12\linewidth}
        \includegraphics[width=\linewidth]{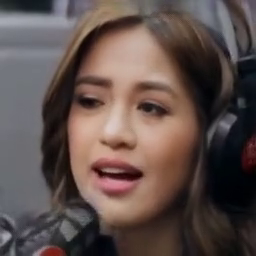}
    \end{subfigure}
     \hspace{-4pt}
        \begin{subfigure}{0.12\linewidth}
        \includegraphics[width=\linewidth]{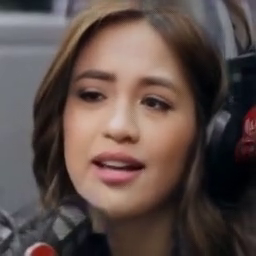}
    \end{subfigure}
     \hspace{-4pt}
        \begin{subfigure}{0.12\linewidth}
        \includegraphics[width=\linewidth]{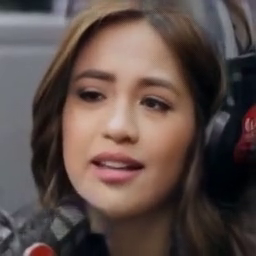}
    \end{subfigure}
     \hspace{-4pt}
        \begin{subfigure}{0.12\linewidth}
        \includegraphics[width=\linewidth]{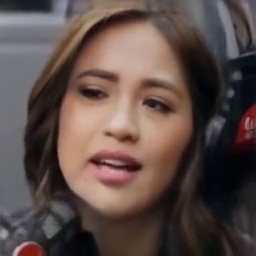}
    \end{subfigure}
     \hspace{-4pt}
        \begin{subfigure}{0.12\linewidth}
        \includegraphics[width=\linewidth]{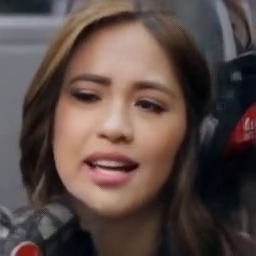}
    \end{subfigure}
    \hspace{-4pt}
        \begin{subfigure}{0.12\linewidth}
        \includegraphics[width=\linewidth]{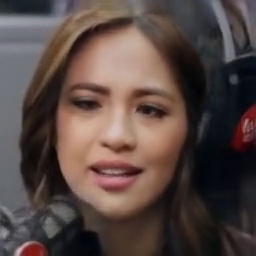}
    \end{subfigure}
    \hspace{-4pt}
        \begin{subfigure}{0.12\linewidth}
        \includegraphics[width=\linewidth]{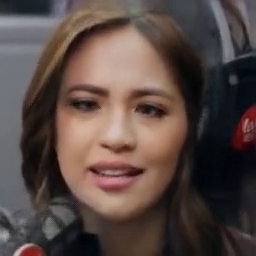}
    \end{subfigure}
    \end{minipage}
    \begin{minipage}{0.02\linewidth}
    \centering
        \rotatebox{90}{SadTalker}
    \end{minipage}
    \begin{minipage}{0.97\linewidth}
    \begin{subfigure}{0.12\linewidth}
        \includegraphics[width=\linewidth]{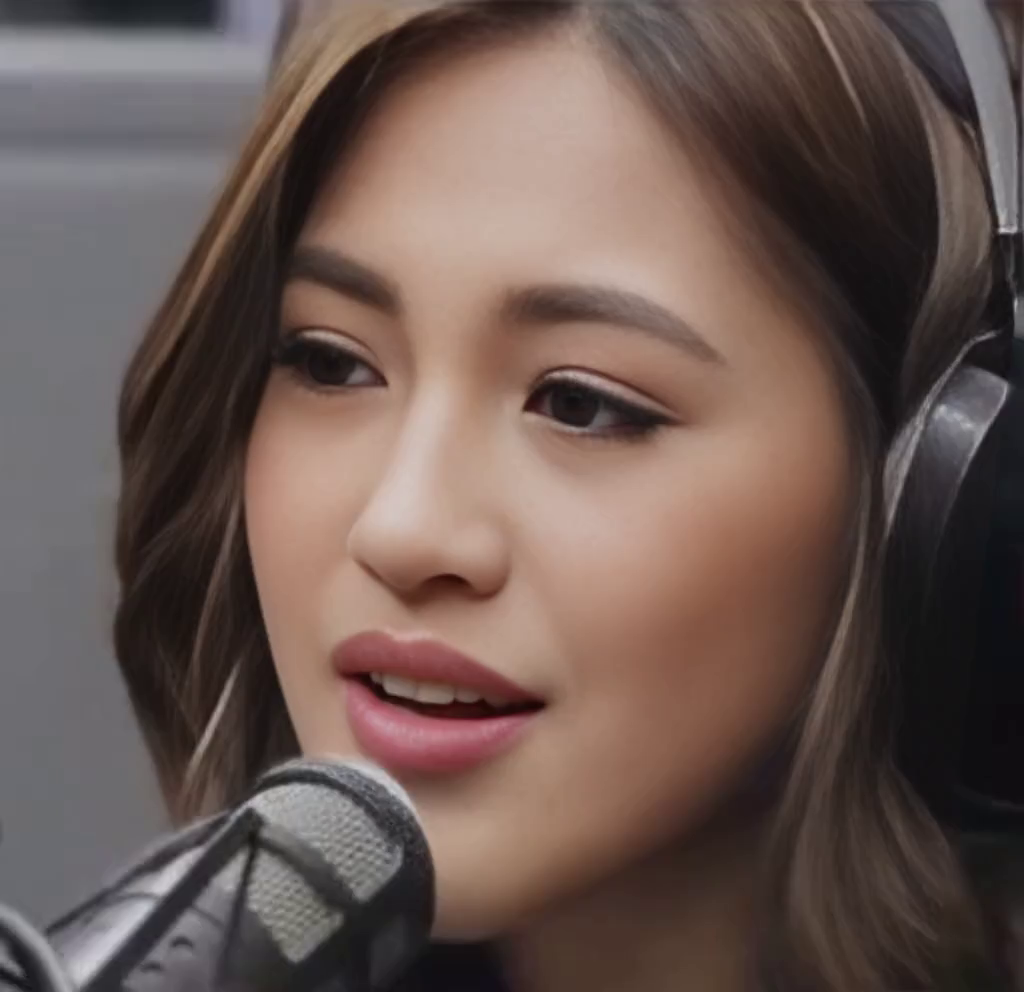}
    \end{subfigure}
    \hspace{-4pt}
        \begin{subfigure}{0.12\linewidth}
        \includegraphics[width=\linewidth]{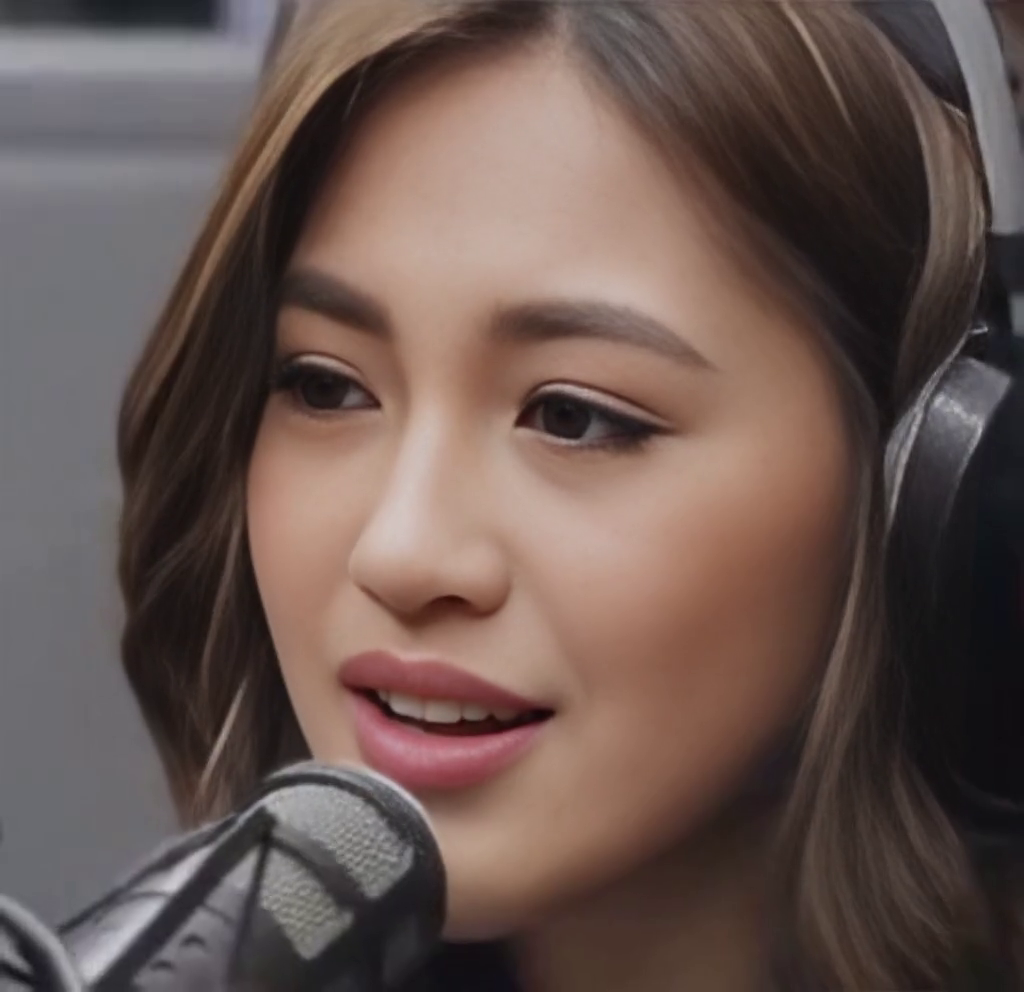}
    \end{subfigure}
     \hspace{-4pt}
        \begin{subfigure}{0.12\linewidth}
        \includegraphics[width=\linewidth]{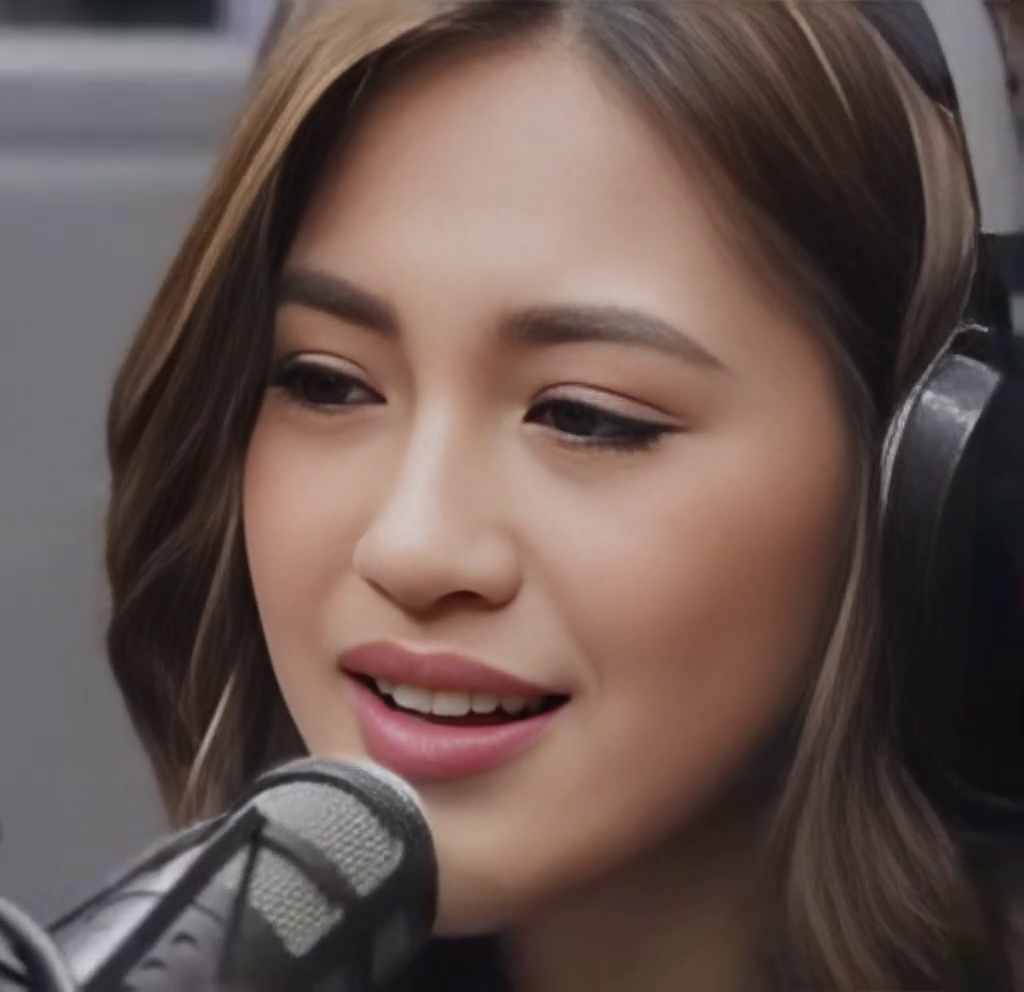}
    \end{subfigure}
     \hspace{-4pt}
        \begin{subfigure}{0.12\linewidth}
        \includegraphics[width=\linewidth]{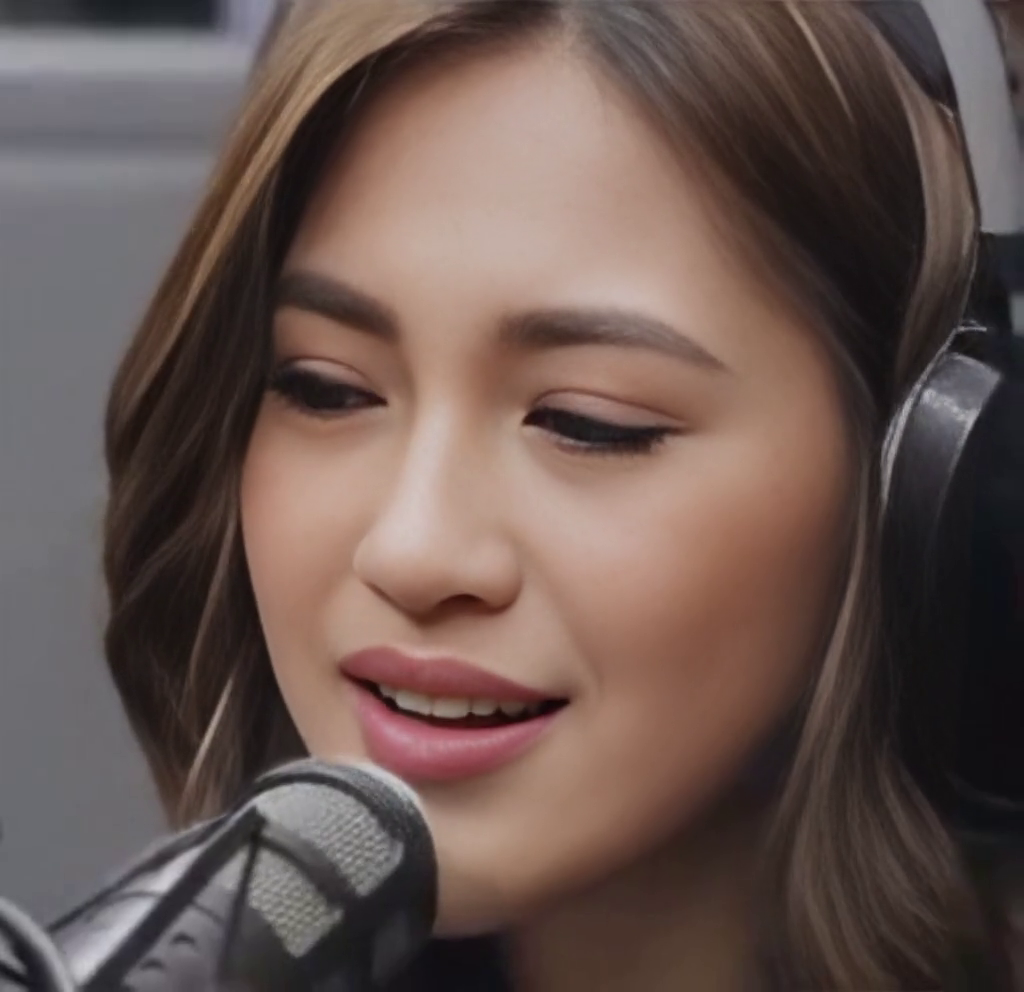}
    \end{subfigure}
     \hspace{-4pt}
        \begin{subfigure}{0.12\linewidth}
        \includegraphics[width=\linewidth]{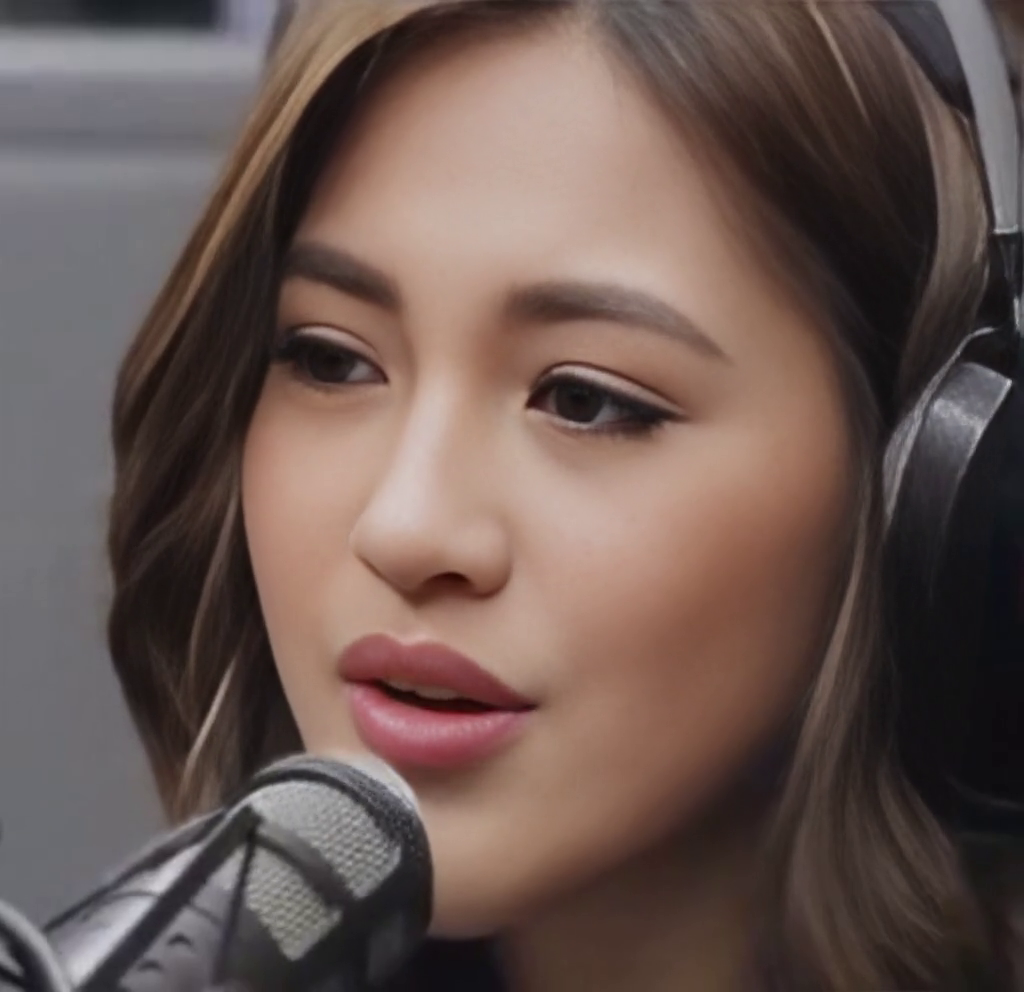}
    \end{subfigure}
     \hspace{-4pt}
        \begin{subfigure}{0.12\linewidth}
        \includegraphics[width=\linewidth]{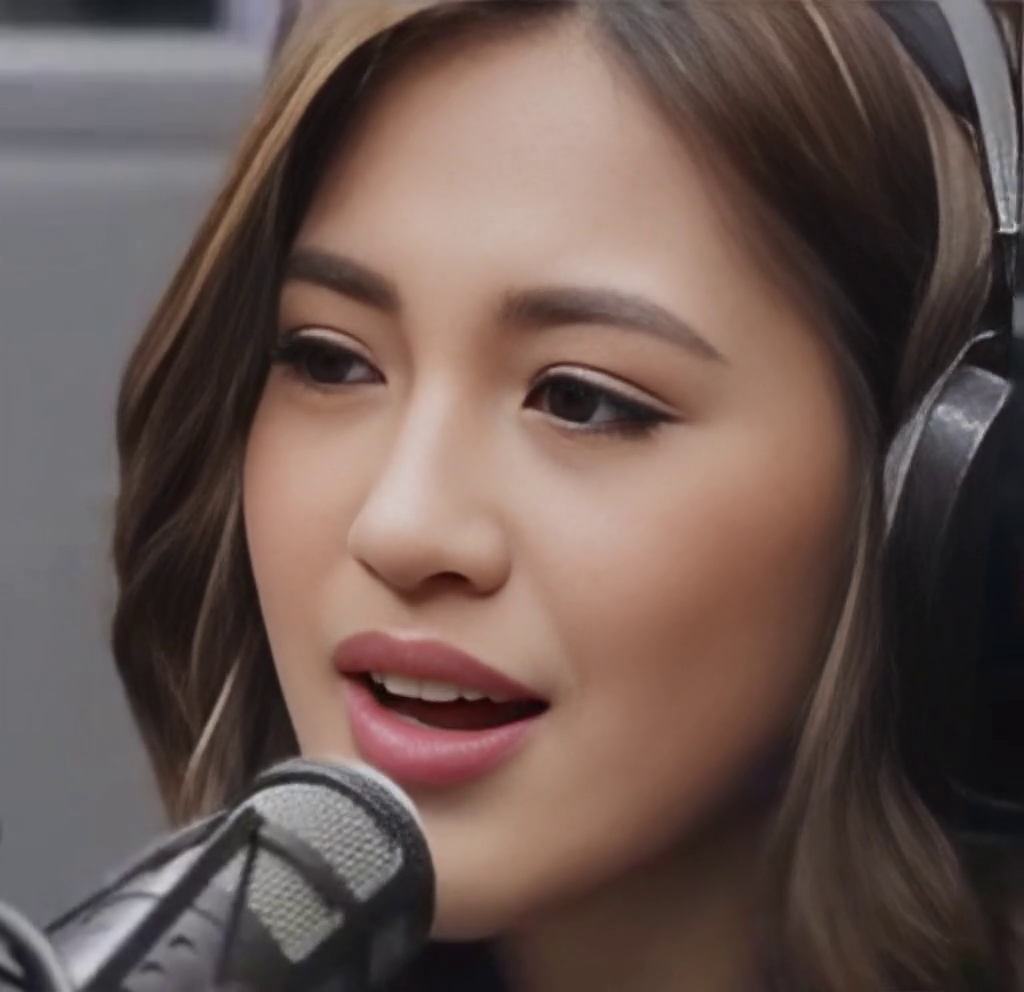}
    \end{subfigure}
    \hspace{-4pt}
        \begin{subfigure}{0.12\linewidth}
        \includegraphics[width=\linewidth]{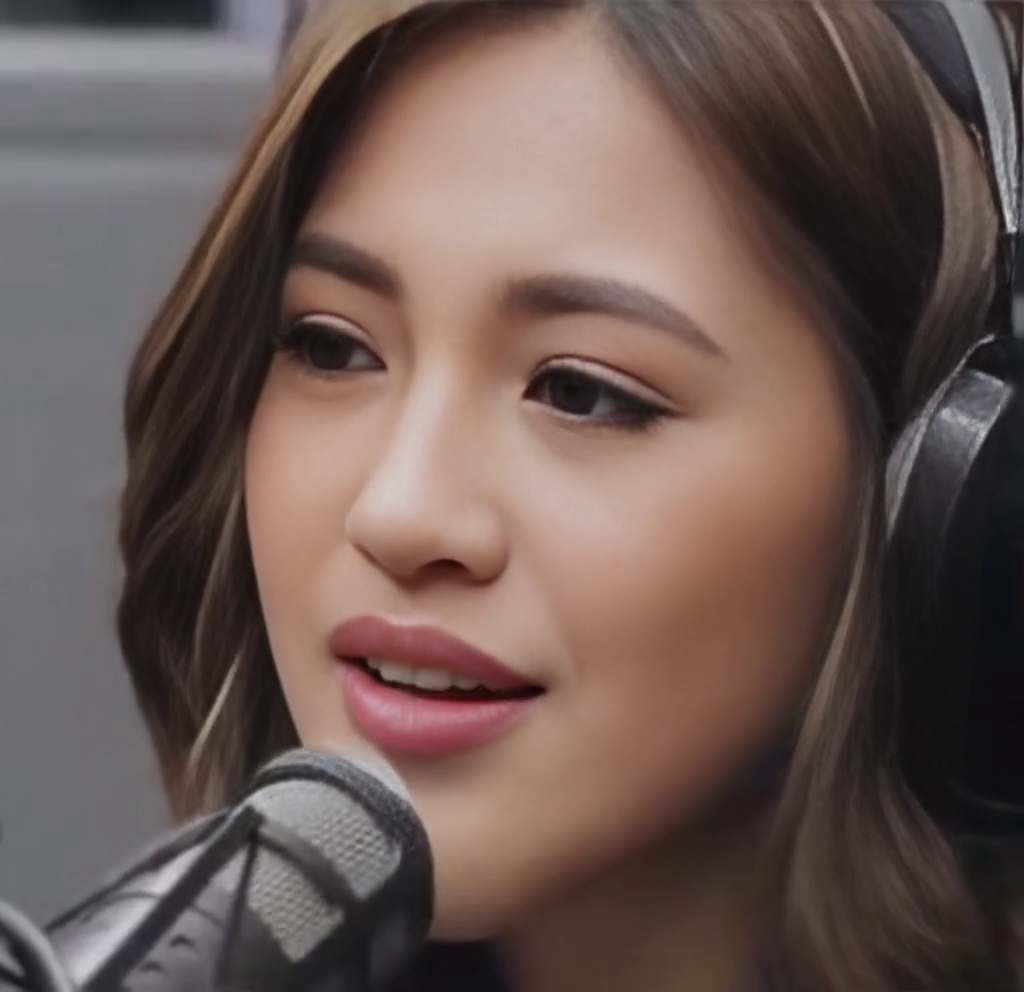}
    \end{subfigure}
    \hspace{-4pt}
        \begin{subfigure}{0.12\linewidth}
        \includegraphics[width=\linewidth]{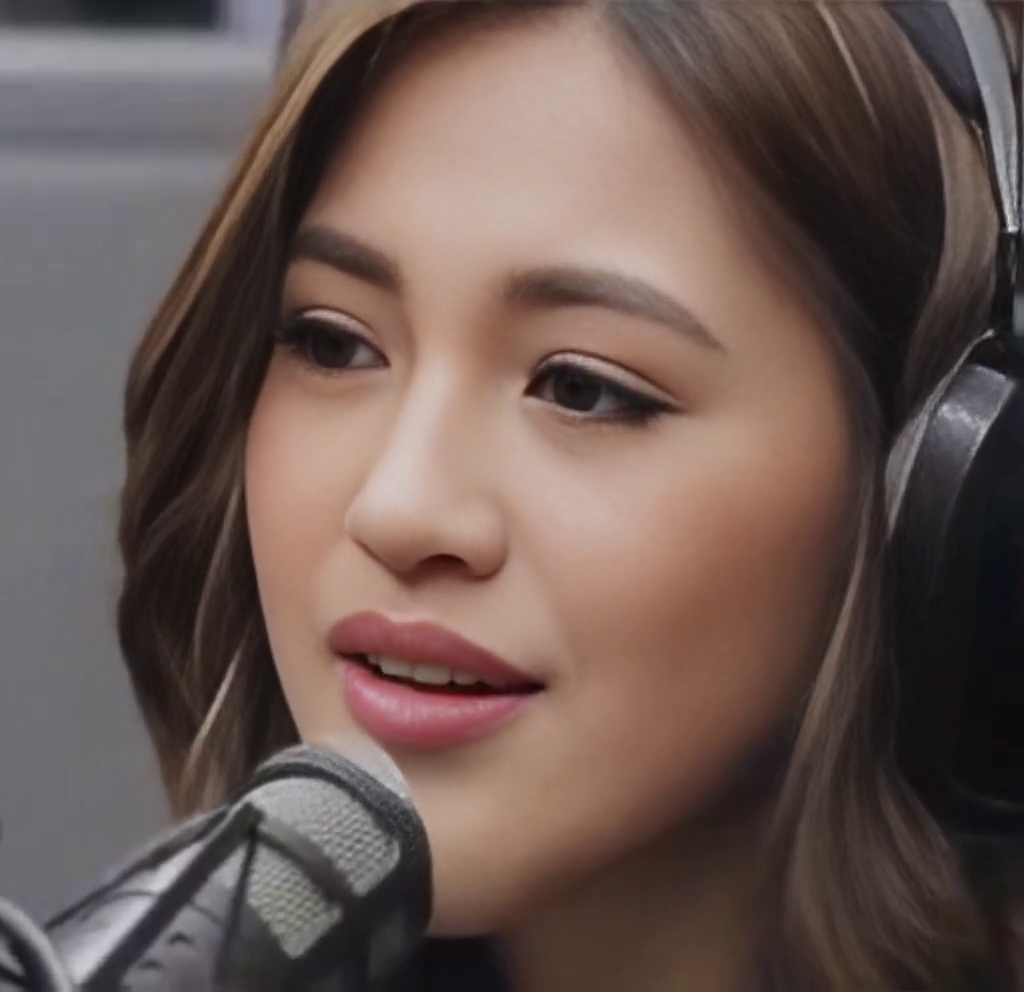}
    \end{subfigure}
    \end{minipage}
      \begin{minipage}{0.02\linewidth}
    \centering
        \rotatebox{90}{MuseTalk}
    \end{minipage}
    \begin{minipage}{0.97\linewidth}
    \begin{subfigure}{0.12\linewidth}
        \includegraphics[width=\linewidth]{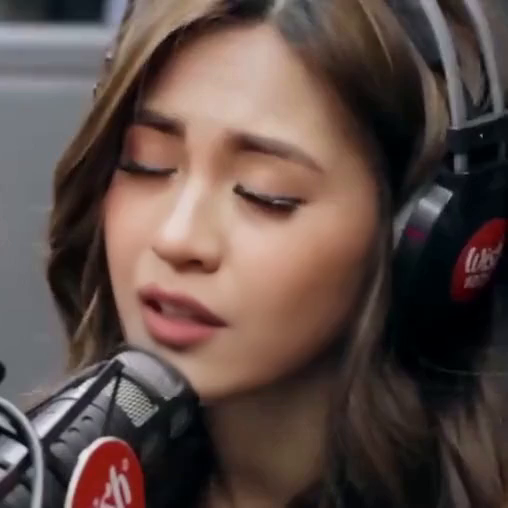}
    \end{subfigure}
    \hspace{-4pt}
        \begin{subfigure}{0.12\linewidth}
        \includegraphics[width=\linewidth]{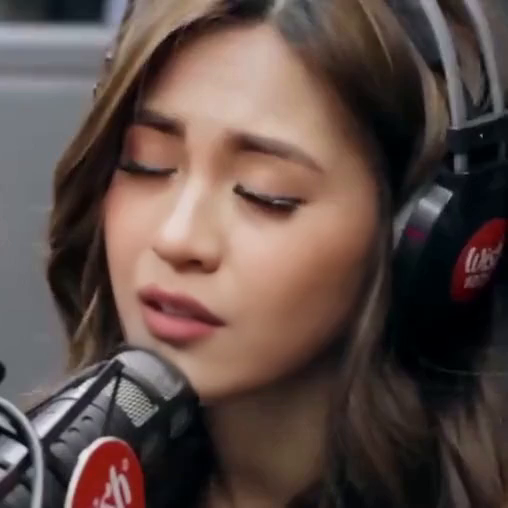}
    \end{subfigure}
     \hspace{-4pt}
        \begin{subfigure}{0.12\linewidth}
        \includegraphics[width=\linewidth]{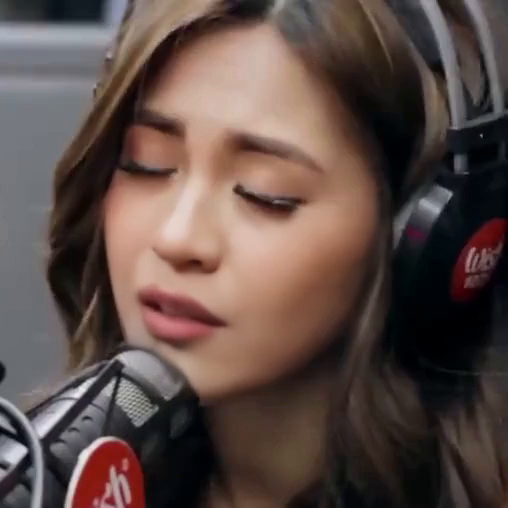}
    \end{subfigure}
     \hspace{-4pt}
        \begin{subfigure}{0.12\linewidth}
        \includegraphics[width=\linewidth]{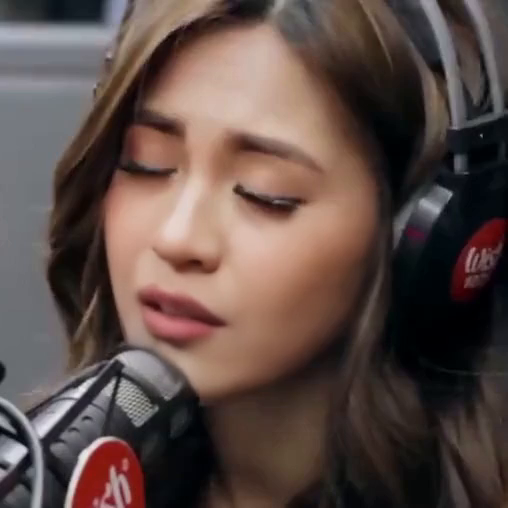}
    \end{subfigure}
     \hspace{-4pt}
        \begin{subfigure}{0.12\linewidth}
        \includegraphics[width=\linewidth]{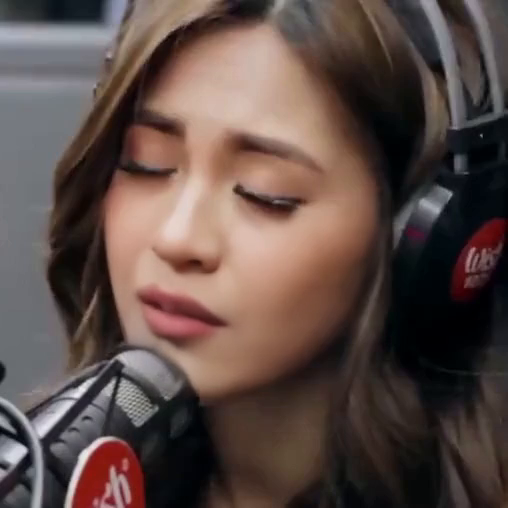}
    \end{subfigure}
     \hspace{-4pt}
        \begin{subfigure}{0.12\linewidth}
        \includegraphics[width=\linewidth]{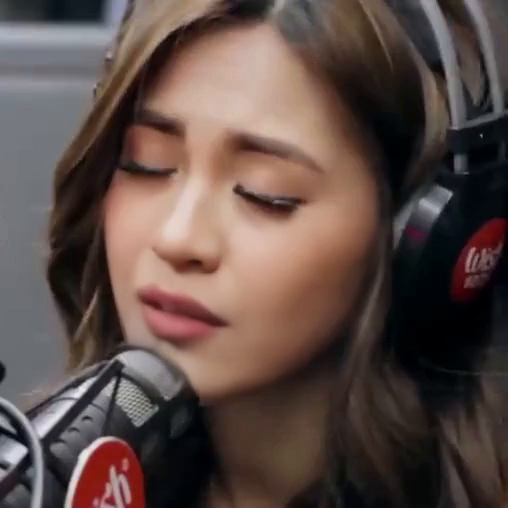}
    \end{subfigure}
    \hspace{-4pt}
        \begin{subfigure}{0.12\linewidth}
        \includegraphics[width=\linewidth]{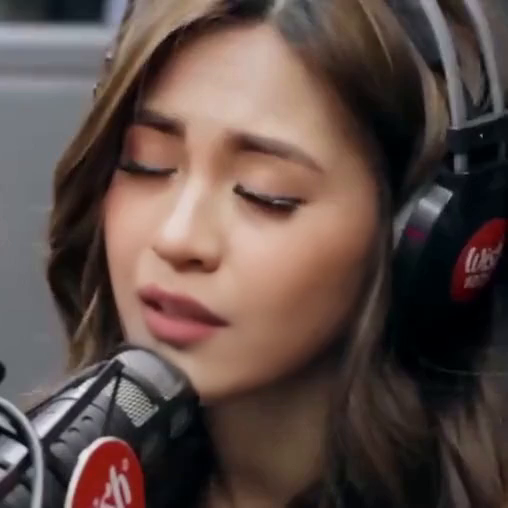}
    \end{subfigure}
    \hspace{-4pt}
        \begin{subfigure}{0.12\linewidth}
        \includegraphics[width=\linewidth]{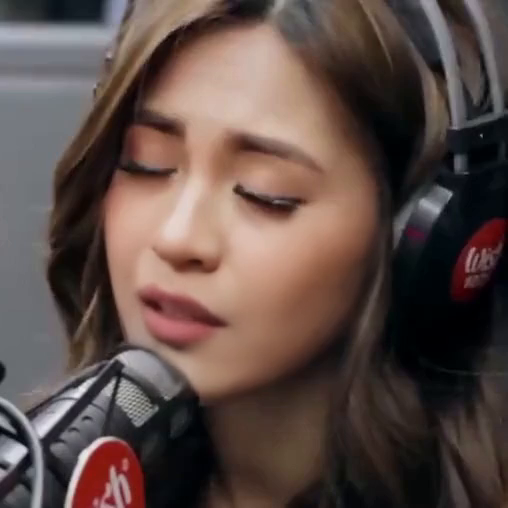}
    \end{subfigure}
    \end{minipage}
      \begin{minipage}{0.02\linewidth}
    \centering
        \rotatebox{90}{AniPortrait}
    \end{minipage}
    \begin{minipage}{0.97\linewidth}
    \begin{subfigure}{0.12\linewidth}
        \includegraphics[width=\linewidth]{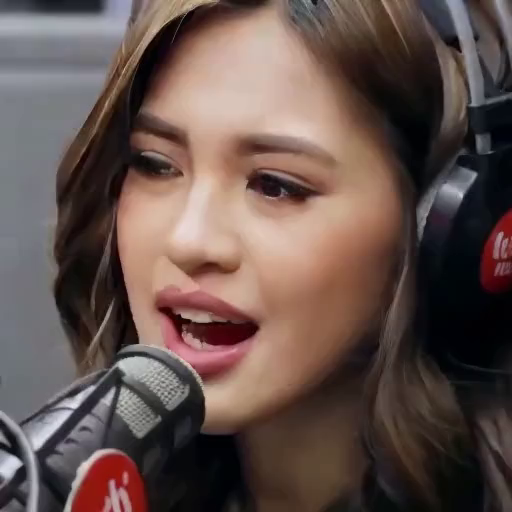}
    \end{subfigure}
    \hspace{-4pt}
        \begin{subfigure}{0.12\linewidth}
        \includegraphics[width=\linewidth]{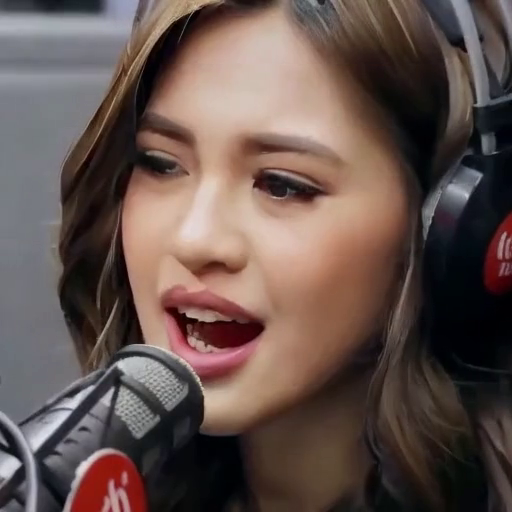}
    \end{subfigure}
     \hspace{-4pt}
        \begin{subfigure}{0.12\linewidth}
        \includegraphics[width=\linewidth]{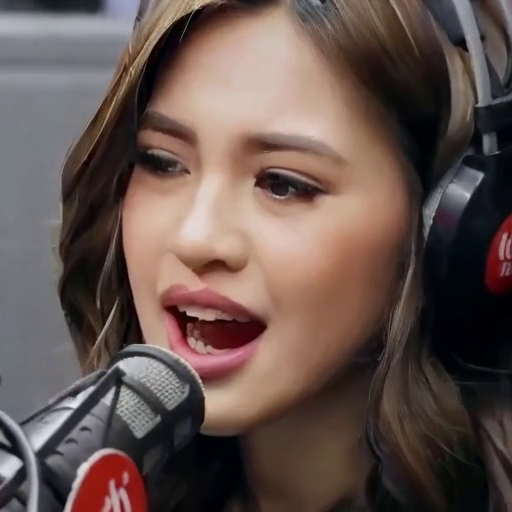}
    \end{subfigure}
     \hspace{-4pt}
        \begin{subfigure}{0.12\linewidth}
        \includegraphics[width=\linewidth]{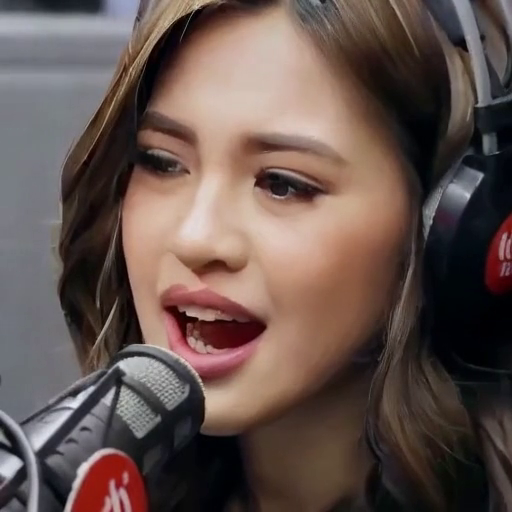}
    \end{subfigure}
     \hspace{-4pt}
        \begin{subfigure}{0.12\linewidth}
        \includegraphics[width=\linewidth]{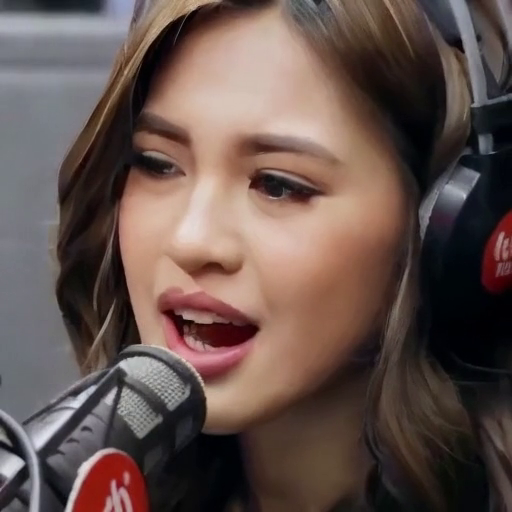}
    \end{subfigure}
     \hspace{-4pt}
        \begin{subfigure}{0.12\linewidth}
        \includegraphics[width=\linewidth]{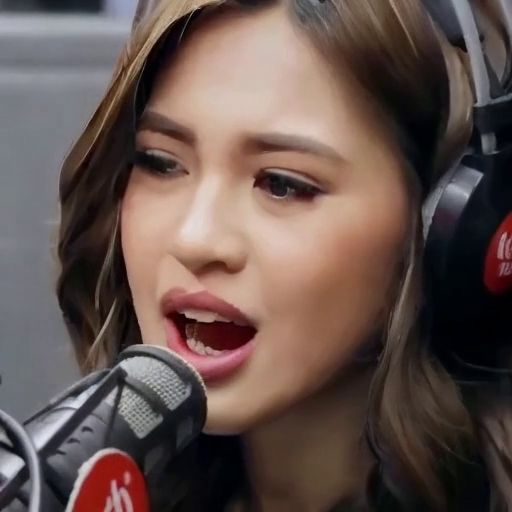}
    \end{subfigure}
    \hspace{-4pt}
        \begin{subfigure}{0.12\linewidth}
        \includegraphics[width=\linewidth]{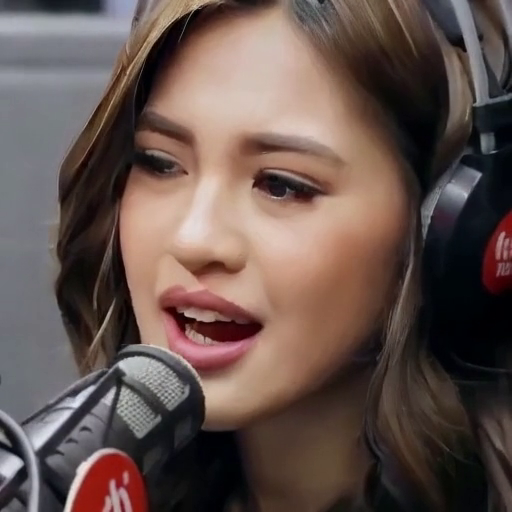}
    \end{subfigure}
    \hspace{-4pt}
        \begin{subfigure}{0.12\linewidth}
        \includegraphics[width=\linewidth]{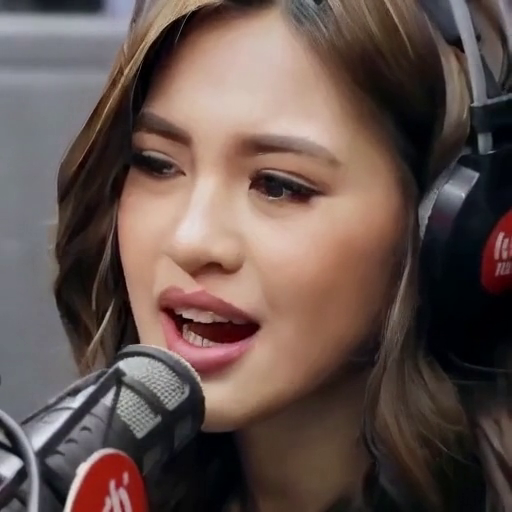}
    \end{subfigure}
    \end{minipage}
      \begin{minipage}{0.02\linewidth}
    \centering
        \rotatebox{90}{Echomimic}
    \end{minipage}
    \begin{minipage}{0.97\linewidth}
    \begin{subfigure}{0.12\linewidth}
        \includegraphics[width=\linewidth]{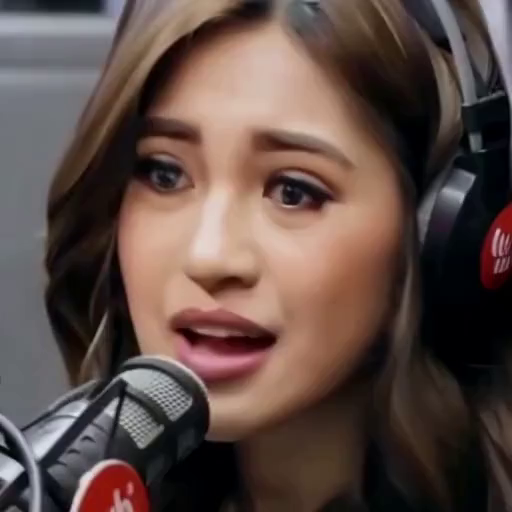}
    \end{subfigure}
    \hspace{-4pt}
        \begin{subfigure}{0.12\linewidth}
        \includegraphics[width=\linewidth]{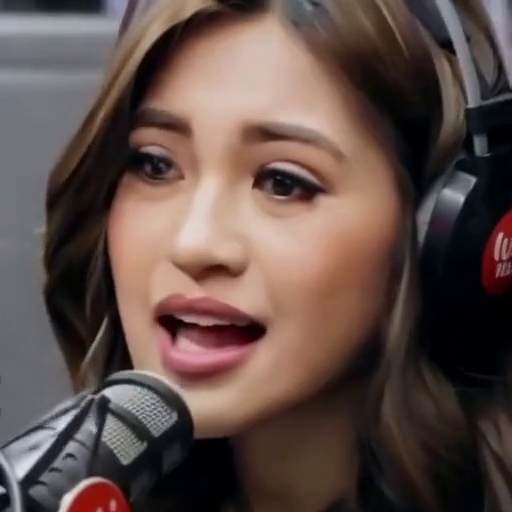}
    \end{subfigure}
     \hspace{-4pt}
        \begin{subfigure}{0.12\linewidth}
        \includegraphics[width=\linewidth]{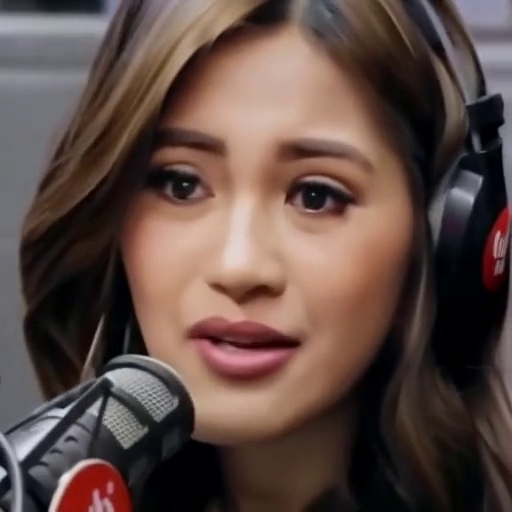}
    \end{subfigure}
     \hspace{-4pt}
        \begin{subfigure}{0.12\linewidth}
        \includegraphics[width=\linewidth]{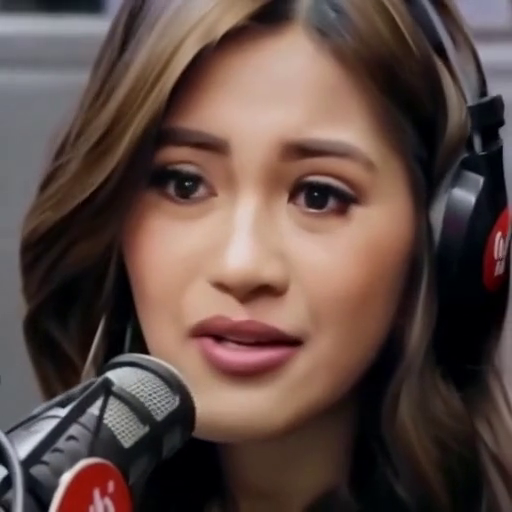}
    \end{subfigure}
     \hspace{-4pt}
        \begin{subfigure}{0.12\linewidth}
        \includegraphics[width=\linewidth]{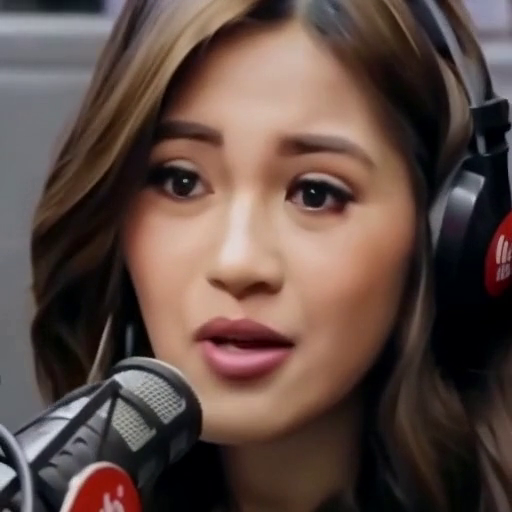}
    \end{subfigure}
     \hspace{-4pt}
        \begin{subfigure}{0.12\linewidth}
        \includegraphics[width=\linewidth]{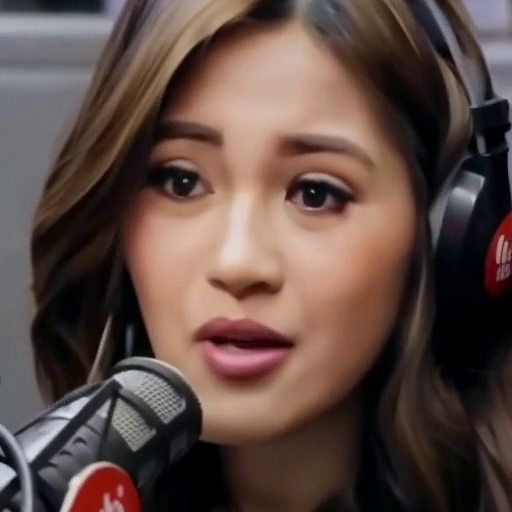}
    \end{subfigure}
    \hspace{-4pt}
        \begin{subfigure}{0.12\linewidth}
        \includegraphics[width=\linewidth]{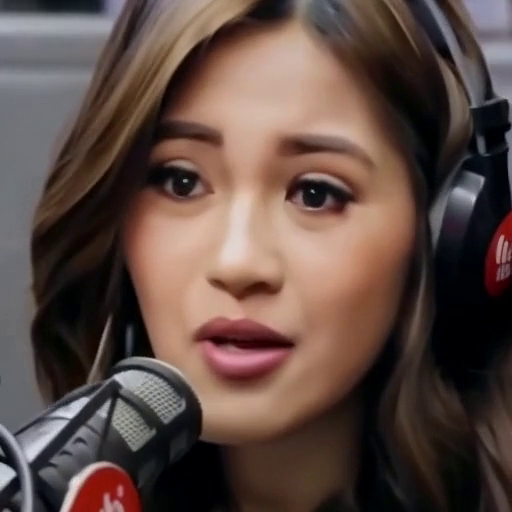}
    \end{subfigure}
    \hspace{-4pt}
        \begin{subfigure}{0.12\linewidth}
        \includegraphics[width=\linewidth]{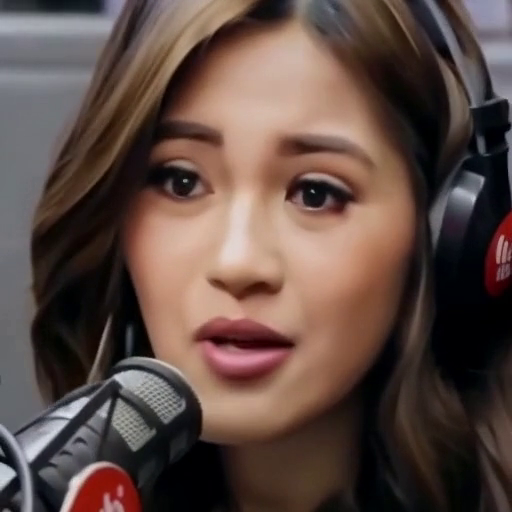}
    \end{subfigure}
    \end{minipage}
      \begin{minipage}{0.02\linewidth}
    \centering
        \rotatebox{90}{Hallo}
    \end{minipage}
    \begin{minipage}{0.97\linewidth}
    \begin{subfigure}{0.12\linewidth}
        \includegraphics[width=\linewidth]{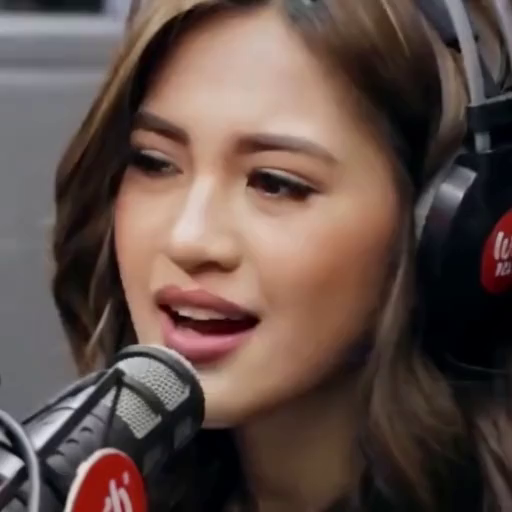}
    \end{subfigure}
    \hspace{-4pt}
        \begin{subfigure}{0.12\linewidth}
        \includegraphics[width=\linewidth]{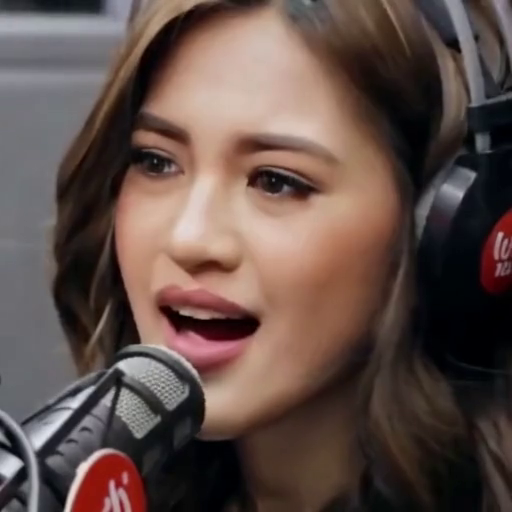}
    \end{subfigure}
     \hspace{-4pt}
        \begin{subfigure}{0.12\linewidth}
        \includegraphics[width=\linewidth]{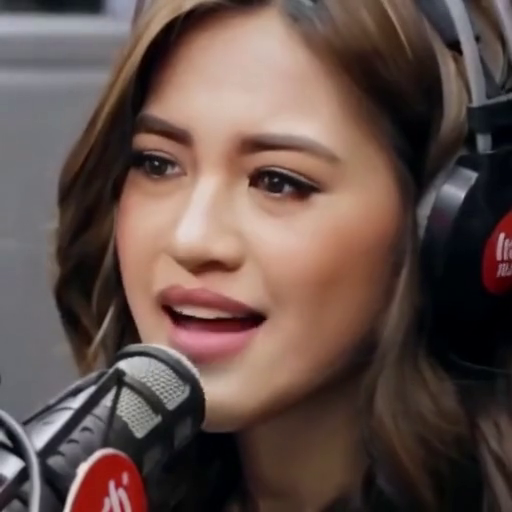}
    \end{subfigure}
     \hspace{-4pt}
        \begin{subfigure}{0.12\linewidth}
        \includegraphics[width=\linewidth]{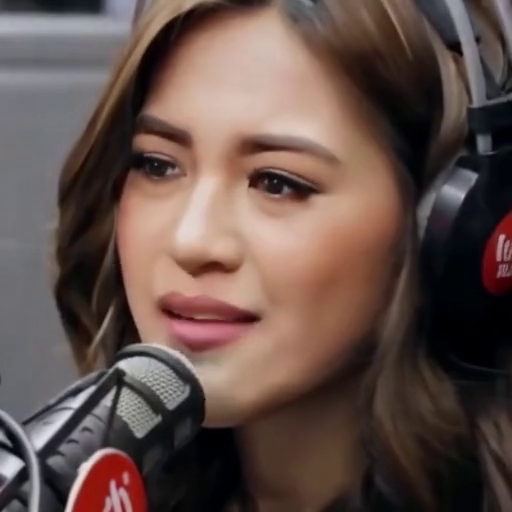}
    \end{subfigure}
     \hspace{-4pt}
        \begin{subfigure}{0.12\linewidth}
        \includegraphics[width=\linewidth]{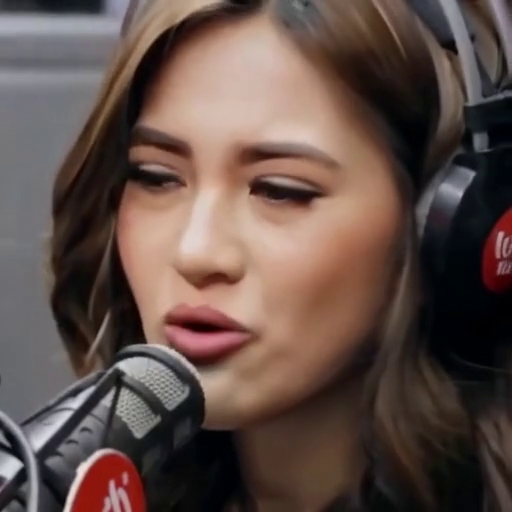}
    \end{subfigure}
     \hspace{-4pt}
        \begin{subfigure}{0.12\linewidth}
        \includegraphics[width=\linewidth]{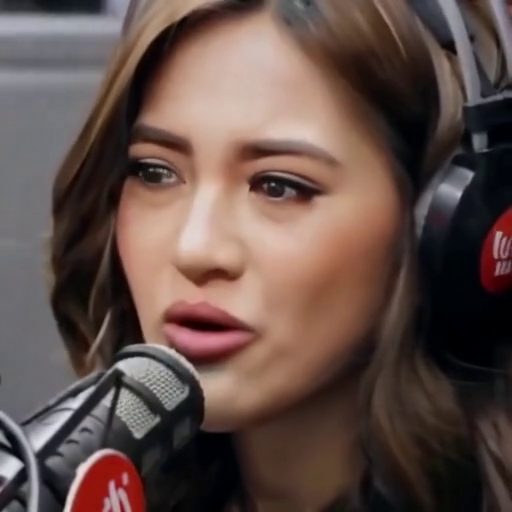}
    \end{subfigure}
    \hspace{-4pt}
        \begin{subfigure}{0.12\linewidth}
        \includegraphics[width=\linewidth]{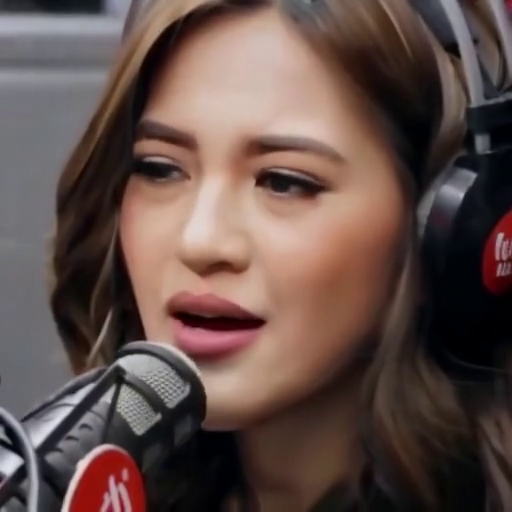}
    \end{subfigure}
    \hspace{-4pt}
        \begin{subfigure}{0.12\linewidth}
        \includegraphics[width=\linewidth]{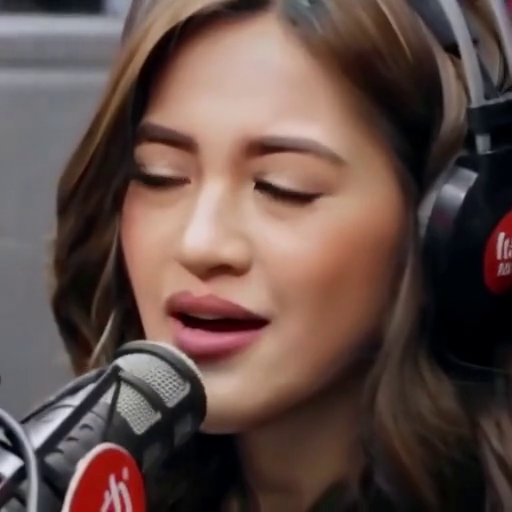}
    \end{subfigure}
    \end{minipage}
      \begin{minipage}{0.02\linewidth}
    \centering
        \rotatebox{90}{Hallo2}
    \end{minipage}
    \begin{minipage}{0.97\linewidth}
    \begin{subfigure}{0.12\linewidth}
        \includegraphics[width=\linewidth]{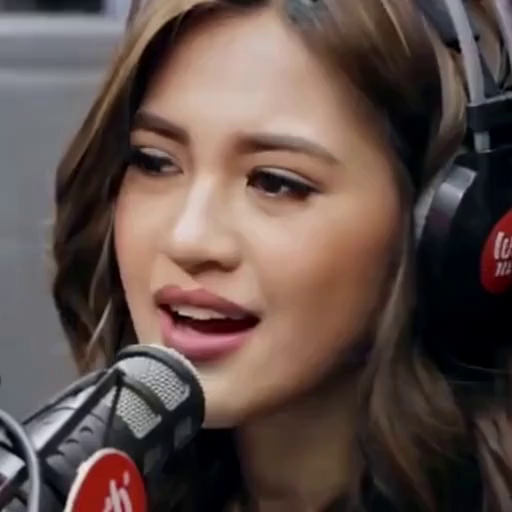}
    \end{subfigure}
    \hspace{-4pt}
        \begin{subfigure}{0.12\linewidth}
        \includegraphics[width=\linewidth]{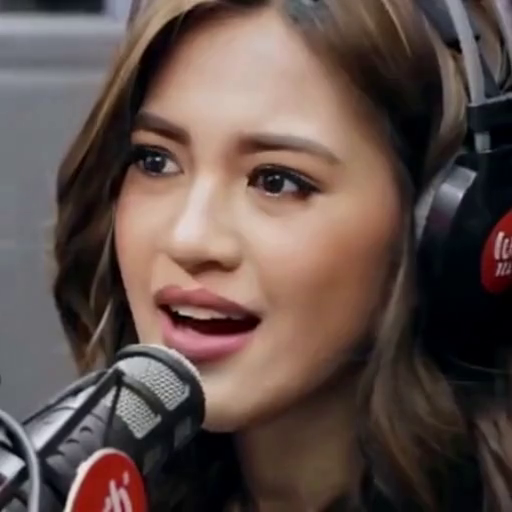}
    \end{subfigure}
     \hspace{-4pt}
        \begin{subfigure}{0.12\linewidth}
        \includegraphics[width=\linewidth]{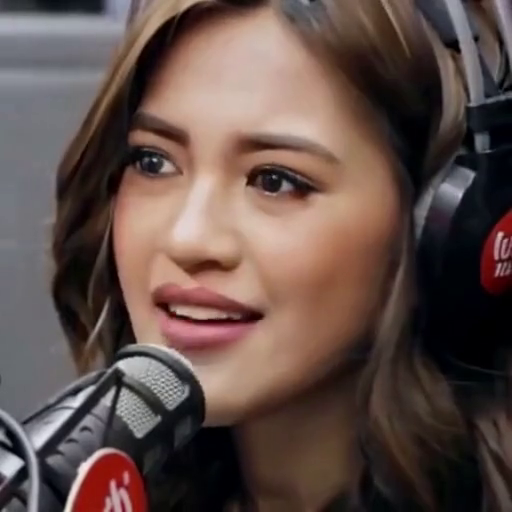}
    \end{subfigure}
     \hspace{-4pt}
        \begin{subfigure}{0.12\linewidth}
        \includegraphics[width=\linewidth]{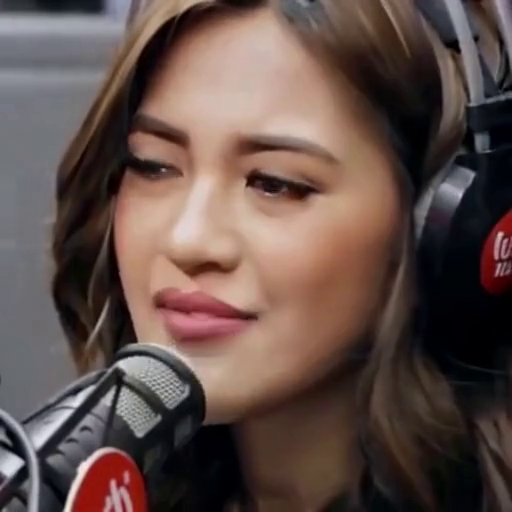}
    \end{subfigure}
     \hspace{-4pt}
        \begin{subfigure}{0.12\linewidth}
        \includegraphics[width=\linewidth]{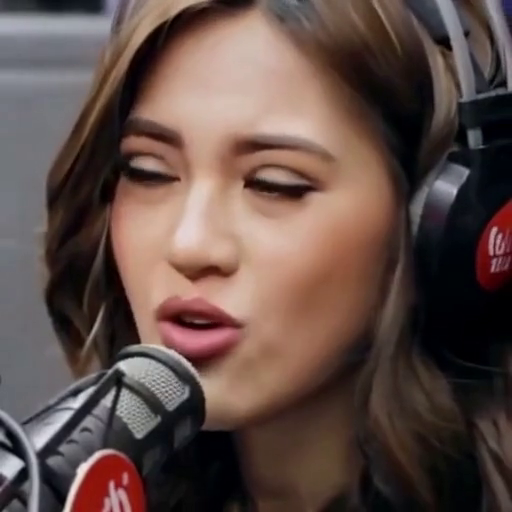}
    \end{subfigure}
     \hspace{-4pt}
        \begin{subfigure}{0.12\linewidth}
        \includegraphics[width=\linewidth]{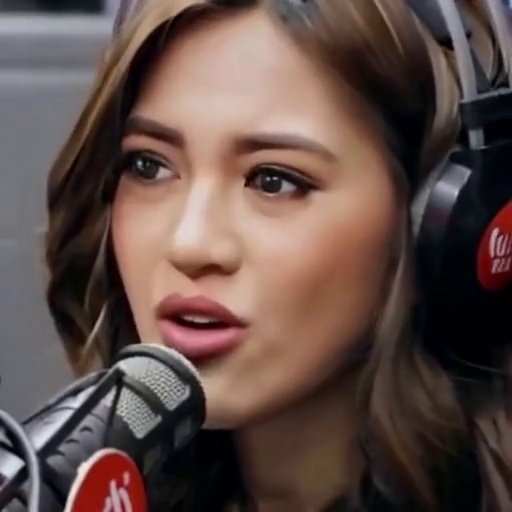}
    \end{subfigure}
    \hspace{-4pt}
        \begin{subfigure}{0.12\linewidth}
        \includegraphics[width=\linewidth]{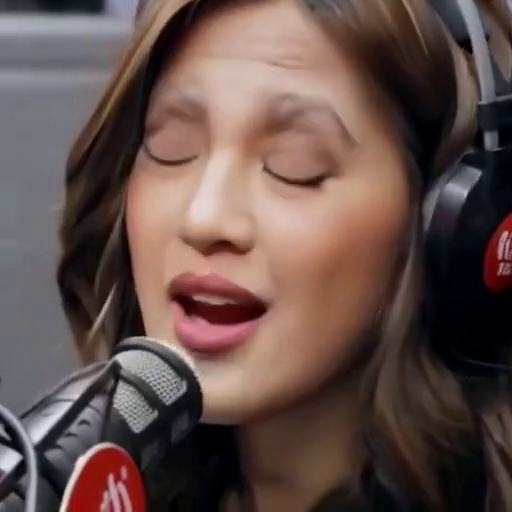}
    \end{subfigure}
    \hspace{-4pt}
        \begin{subfigure}{0.12\linewidth}
        \includegraphics[width=\linewidth]{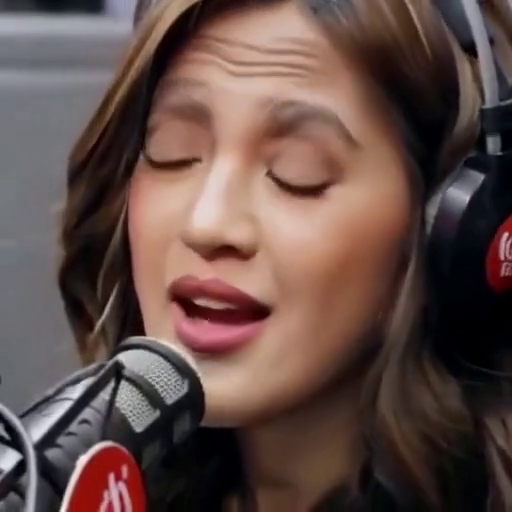}
    \end{subfigure}
    \end{minipage}
  
    \begin{minipage}{0.02\linewidth}
    \centering
        \rotatebox{90}{\model}
    \end{minipage}
    \begin{minipage}{0.97\linewidth}
    \begin{subfigure}{0.12\linewidth}
        \includegraphics[width=\linewidth]{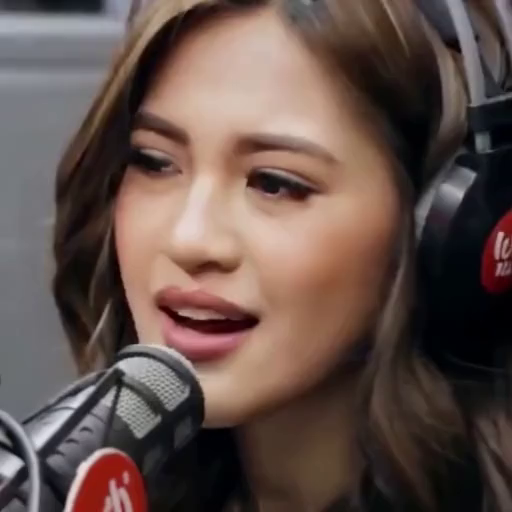}
    \end{subfigure}
    \hspace{-4pt}
        \begin{subfigure}{0.12\linewidth}
        \includegraphics[width=\linewidth]{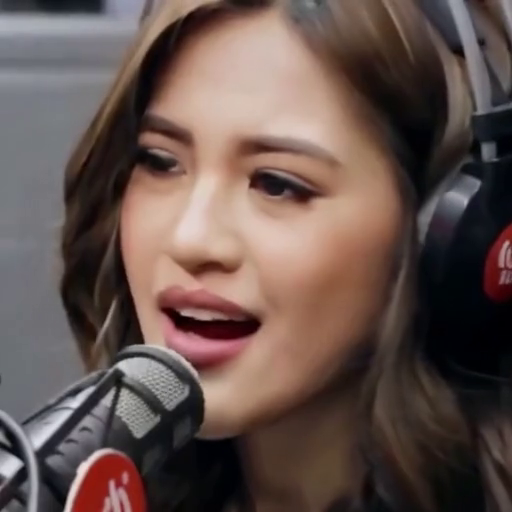}
    \end{subfigure}
     \hspace{-4pt}
        \begin{subfigure}{0.12\linewidth}
        \includegraphics[width=\linewidth]{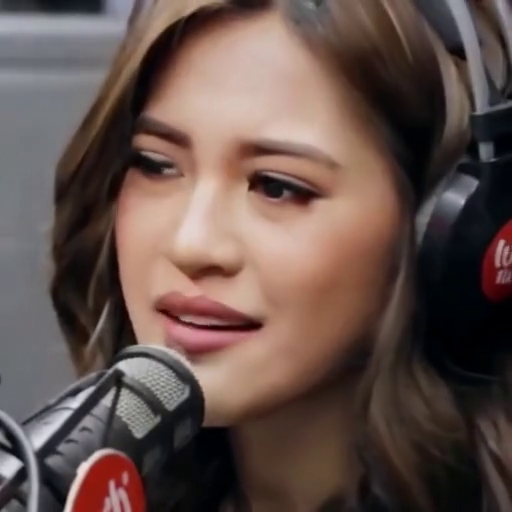}
    \end{subfigure}
     \hspace{-4pt}
        \begin{subfigure}{0.12\linewidth}
        \includegraphics[width=\linewidth]{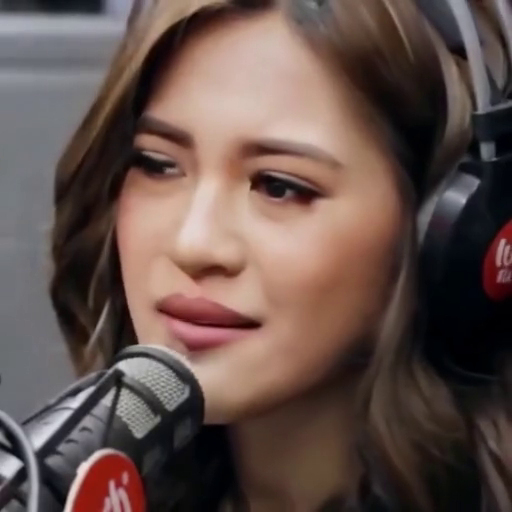}
    \end{subfigure}
     \hspace{-4pt}
        \begin{subfigure}{0.12\linewidth}
        \includegraphics[width=\linewidth]{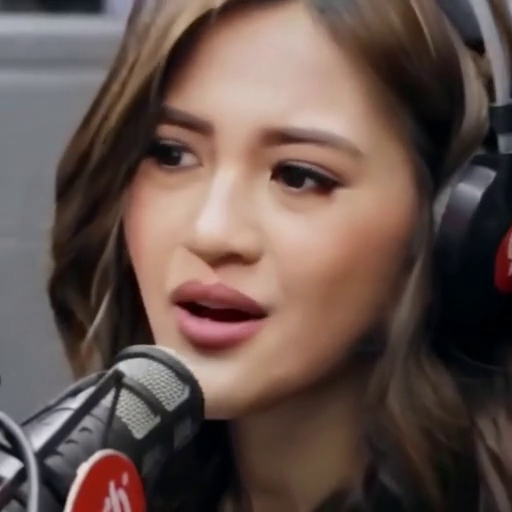}
    \end{subfigure}
     \hspace{-4pt}
        \begin{subfigure}{0.12\linewidth}
        \includegraphics[width=\linewidth]{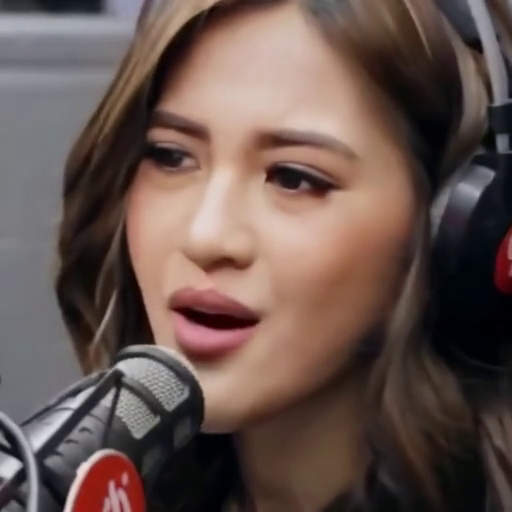}
    \end{subfigure}
    \hspace{-4pt}
        \begin{subfigure}{0.12\linewidth}
        \includegraphics[width=\linewidth]{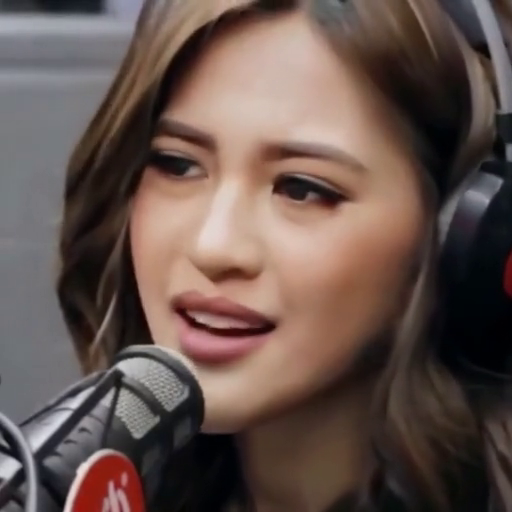}
    \end{subfigure}
    \hspace{-4pt}
        \begin{subfigure}{0.12\linewidth}
        \includegraphics[width=\linewidth]{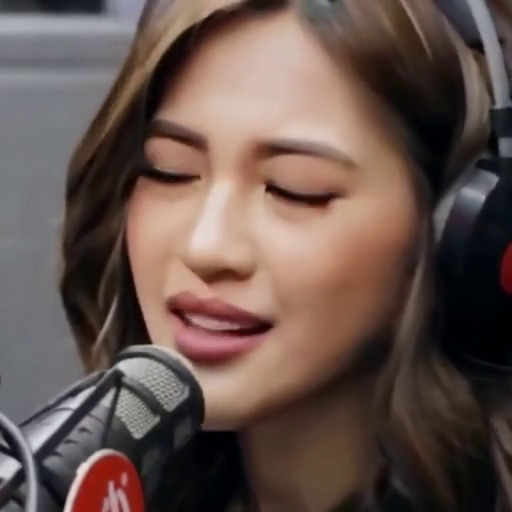}
    \end{subfigure}
    \end{minipage}
    \caption{The visualization of generated singing videos by baseline methods and our \model. }
    \label{fig:full_comprison_1}
\end{figure*}

We present the visualization of generated singing videos by seven baseline methods and \model~in Figures~\ref{fig:full_comprison_app} and~\ref{fig:full_comprison_1}. The results demonstrate the ability of our method to generate more vivid and realistic singing videos compared to the baseline methods, highlighting the effectiveness of our approach in capturing the complex dynamics of singing, including head movements and lip synchronization.

\end{document}